\pdfoutput=1
\documentclass{article}
\PassOptionsToPackage{round,authoryear}{natbib}
\usepackage[preprint]{neurips_2026}
\usepackage[utf8]{inputenc}
\usepackage[T1]{fontenc}
\usepackage{rotating}
\usepackage{microtype}
\usepackage{booktabs}
\usepackage{array}
\usepackage{graphicx}
\usepackage{amsmath,amssymb}
\usepackage{xcolor}
\usepackage{hyperref}
\usepackage{caption}
\usepackage{enumitem}
\usepackage{url}
\usepackage{placeins}
\usepackage{longtable}
\hypersetup{colorlinks,linkcolor=blue!50!black,citecolor=blue!50!black,urlcolor=blue!50!black,pdftitle={CueClash: Frontier Vision-Language Models Judge 3-D Relations Close to the Picture Plane},pdfauthor={Soham Dan}}
\input{numbers}
\newcommand{\bench}{CueClash}
\newcommand{\sgap}{shortcut gap}
\newcommand{\family}[1]{\texttt{#1}}

\title{CueClash: Frontier Vision-Language Models\\Judge 3-D Relations Close to the Picture Plane}
\author{Soham Dan\\Scale AI\\\texttt{soham.dan@scale.com}}
\date{}

\begin{document}
\maketitle

\begin{abstract}
A vision-language model (VLM) can judge 3-D relations from the picture or from the ground layout. In photographs the two usually agree. We separate them with \bench{}, rendered scenes in which the 2-D cue for each of five relations is right on exactly half the statements. The \sgap{} is accuracy when the cue is right minus accuracy when it is wrong. For Claude Opus 5.5, GPT-5.5 and Gemini 3.6 Flash at low reasoning effort it is \n{main.gaprange} percentage points. It is \n{r10.clean.gaprange} without two relations whose cue-wrong items are confounded with borderline cases or occlusion. Given ground coordinates and each relation's definition, they answer \n{main.coordsnoimagendef.accrange}\% correctly. Pixel measurements with camera depths remove most of the gap for objects fitting through openings. The failure lies in the step from the picture to the layout. The maps that the models draw of the scene from above mostly keep the depth order but, like the picture, shorten depth. GPT-5.5's maps stay close to a copy of the picture. Opus 5.5's and Gemini's maps undo about 0.6 of this shortening. Gemini's then shrink all depths by a further uniform factor. Opus 5.5 follows the picture mainly on \family{fits}, and its maps correct depth about as much as human observers of real scenes do. The three models also lose accuracy on CV-Bench photographs where a reliable 2-D cue misleads. In two open models the layout information reaches the language model, but their maps lose much of it.
\end{abstract}

\section{Introduction}
\label{sec:intro}

\begin{figure}[t]
\centering
\includegraphics[width=\textwidth]{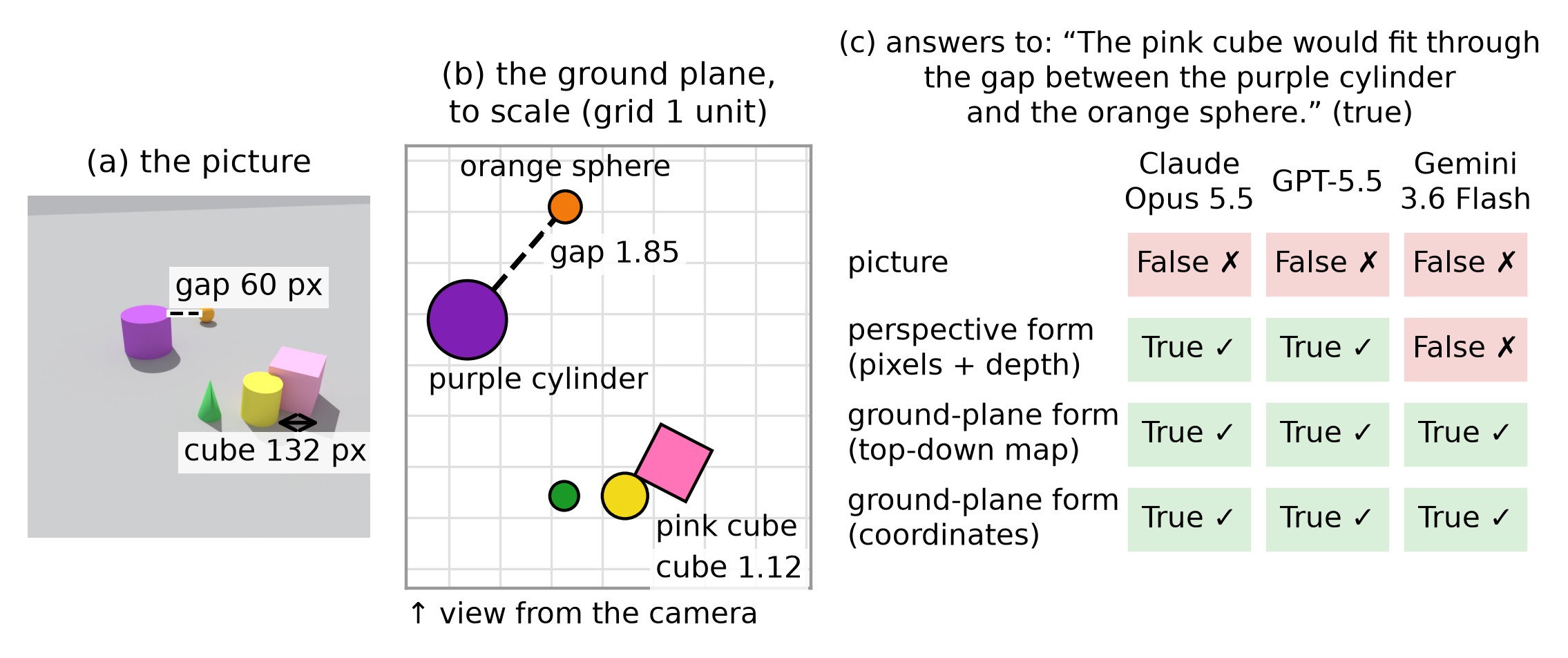}
\caption{The same \family{fits} item read in the picture plane and on the ground plane. (a) In the image the gap spans 60\,px and the nearer cube 132\,px. (b) On the ground plane the gap is 1.85 units and the cube 1.12, so the cube fits. (c) Opus 5.5, GPT-5.5 and Gemini at low reasoning effort answer False from the picture and True from the ground plane. Given pixel measurements and camera depths, which nearly determine the answer, only Gemini still answers False. Only the input changes between the rows. Over the 600 medium items their pooled gaps are \n{Opus.medium.image.gap}, \n{Gpt.medium.image.gap} and \n{Gemini.medium.image.gap} points from the picture and \n{Opus.medium.topdownoracle.gap}, \n{Gpt.medium.topdownoracle.gap} and \n{Gemini.medium.topdownoracle.gap} from the top-down map (Table~\ref{tab:ladder}).}
\label{fig:running}
\end{figure}

Is the pink cube in Figure~\ref{fig:running} narrow enough to pass between the purple cylinder and the orange sphere? In the picture it looks too wide, because the cube spans 132 pixels and the gap only 60. On the ground the gap is 1.85 units and the cube 1.12, so the cube fits with room to spare. The picture is wrong about the widths only because the cube stands nearer the camera than the gap does. Claude Opus 5.5, GPT-5.5 and Gemini 3.6 Flash, which we call the frontier models, all say that it does not fit. Shown a to-scale map of the same scene from above, all three say that it does.

Spatial benchmarks for VLMs ask many questions of this kind \citep{kamath2023whatsup,cheng2024spatialrgpt,tong2024cambrian,yang2025thinking,ma2025threedsrbench}, but their scores cannot tell whether an answer came from the picture or from the scene. In natural photographs the two rarely come apart. The taller silhouette usually belongs to the taller object, just as silhouettes that overlap usually belong to objects that touch. A model that reads relations straight off the image can then score about as well as one that recovers the layout \citep{geirhos2020shortcut}. The usual remedy, giving the model a depth map, coordinates or a bird's-eye view, adds a second confound, because it changes what the model is told and the form in which it is told at the same time.

\begin{figure}[t]
\centering
\includegraphics[width=\textwidth]{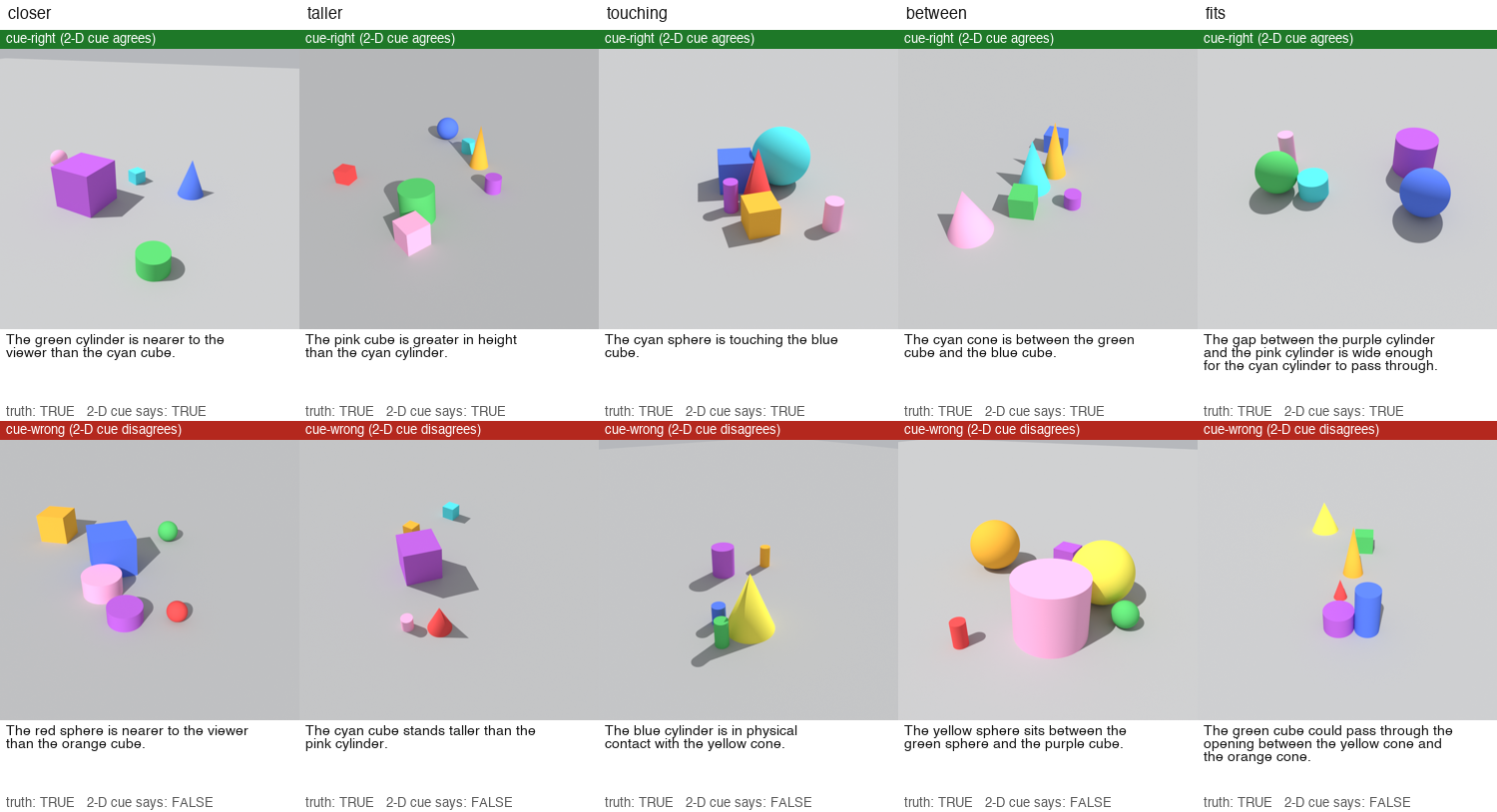}
\caption{One cue-right (top) and one cue-wrong (bottom) true item per relation from the medium tier. In the bottom row the named 2-D cue gives the wrong answer, which is what the \sgap{} tests. The cues are pixel area for \family{closer}, pixel height for \family{taller}, masks within 3\,px for \family{touching}, the angle test on mask centroids for \family{between} and pixel gap against pixel width for \family{fits}.}
\label{fig:teaser}
\end{figure}

We study the question with \bench{}, a benchmark built on a controlled cue conflict. The five relations are \family{closer}, \family{taller}, \family{touching}, \family{between} and \family{fits} (Table~\ref{tab:families}). For each relation we name the 2-D cue that a reader of the picture would use, such as comparing pixel widths for \family{fits}. We balance the statements so that this cue is right on exactly half of them. Items on which the cue gives the right answer are cue-right items, and the others are cue-wrong items (Figure~\ref{fig:teaser}). A model that only reads the cue then scores 50\%, the same as a coin. A model's \emph{\sgap{}} is its accuracy on cue-right items minus its accuracy on cue-wrong items, in percentage points. The gap measures how far the answers follow the picture. The items come in three difficulty tiers, easy, medium and hard.

We then give the models the same scene in other forms. The \emph{perspective form} gives exact pixel measurements, optionally with camera depths, which describe the picture, while the \emph{ground-plane form} gives a top-down map or ground coordinates, which describe the scene. A deterministic solver that reads exactly the text a model received shows how far each form determines the answer, so a gap that remains under a sufficient input can be attributed to the form of that input. In a last task the models draw the scene from above. We compare these drawn maps with the \emph{picture-plane copy}, the layout obtained by reading positions straight off the image.

The design lets us ask five questions in turn. Do the answers follow the picture (RQ1)? If they do, is the information missing from the inputs (RQ2a), or the ability to reason about the relations (RQ2b)? What layout do the models recover from the picture (RQ3)? Does the gap appear in photographs, where the picture rarely misleads (RQ4)? In open models, does the information that the picture carries reach the language model (RQ5)?

The failure lies in the step from the picture to the layout. At low reasoning effort the frontier models follow the picture, with gaps of \n{main.gaprange} points (RQ1, Section~\ref{sec:res-image}). Without \family{between} and \family{touching}, whose cue-wrong items are confounded with borderline cases or occlusion, the gaps are \n{r10.clean.gaprange}. On the cue-wrong \family{fits} items GPT-5.5 falls below chance. High reasoning effort narrows the \family{fits} gap for Opus 5.5 and GPT-5.5 without closing it. Exact pixel measurements with camera depths close most of the \family{fits} gap, so most of that failure lies in getting from the picture to measurements and depths (RQ2a, Section~\ref{sec:res-info}). The \family{between} errors persist, because the models apply the \family{between} rule to image-plane angles and read the word ``between'' loosely. Given ground coordinates and the definition of each relation, the same models answer \n{main.coordsnoimagendef.accrange}\% of items correctly (RQ2b, Section~\ref{sec:res-ground}).

The drawn maps mostly keep the depth order but compress depth relative to width (RQ3, Section~\ref{sec:res-maps}). On a constancy index from 0 for the picture-plane copy to 1 for the scene, GPT-5.5's maps score \n{ti.Gpt.medium} and \n{ti.Gpt.hard} on the medium and hard tiers. Opus 5.5's score \n{ti.Opus.medium} and \n{ti.Opus.hard}, near the partial constancy reported for human observers of real scenes.

On CV-Bench photographs the frontier models show gaps of \n{cv.bottom.gaprange} points where the lower object in the frame is the farther one (RQ4, Section~\ref{sec:res-photo}). In Qwen3-VL-4B and Qwen3-VL-8B a linear probe trained on the true layout reads each object's ground position from the language model's internal states about as well as from exact pixel measurements (RQ5, Section~\ref{sec:res-probe}). The same models' drawn maps keep only \n{probe.own.ratio} and \n{p8.own.ratio} of the true depth-to-width ratio.

This paper makes four contributions.
\begin{itemize}\setlength{\itemsep}{1pt}
\item \bench{}, a benchmark of rendered scenes that balances the 2-D cue against the truth for five relations on three difficulty tiers, with the \sgap{} as its measure (Section~\ref{sec:bench}).
\item A way to locate the failure. The same scene is given in the perspective form and in the ground-plane form, and a solver shows how far each input determines the answer (Sections~\ref{sec:res-info} and~\ref{sec:res-ground}).
\item An account of the layout that the models recover. Drawn maps scored with a constancy index show that the frontier models keep most of the depth order but compress depth, each to a different degree (Section~\ref{sec:res-maps}).
\item Two further checks. The gap appears on CV-Bench photographs where a 2-D cue misleads (Section~\ref{sec:res-photo}). Probes of two open models on the rendered scenes show the picture's information reaching the language model (Section~\ref{sec:res-probe}).
\end{itemize}

\section{Related work}

\paragraph{Cue conflict and balanced evaluation.} Cue-conflict stimuli set two cues against each other to show which one a network follows, as in the texture--shape study of \citet{geirhos2019texture}. Visual question answering has long balanced answers across paired images \citep{zhang2016yinyang,goyal2017vqav2} and shifted answer priors between training and test \citep{agrawal2018vqacp,kervadec2021gqaood}, while minimal pairs isolate one factor at a time \citep{thrush2022winoground,goyal2020rel3d}. SpatialSense collected spatial relations adversarially against a baseline that reads the 2-D configuration \citep{yang2019spatialsense}. Among spatial benchmarks, 3DSRBench avoids questions that the shortcut ``higher in 3-D is higher in 2-D'' would answer \citep{ma2025threedsrbench}. We keep such items and balance them instead, so that the shortcut is right on half of every relation and its use can be measured.

\paragraph{Spatial benchmarks and depth cues.} Benchmarks built from photographs judge relations such as front and behind without metric 3-D truth \citep{liu2023vsr,kamath2023whatsup,fu2024blink}, while newer suites attach 3-D annotations at scale \citep{tong2024cambrian,chen2024spatialvlm,yang2025thinking}. Work on the cues themselves finds monocular depth information in vision encoders \citep{elbanani2024probe3d,danier2025depthcues} and size constancy in VLMs that depends strongly on model scale \citep{sun2025constancy}. VLMs also order depth near chance on real and rendered images \citep{liu2026o3d} and tend to treat higher in the frame as farther \citep{min2026whyfar,zou2026seeingdepth}. Height in the frame is the angular-declination cue of human distance perception, which is valid for objects on a common visible ground \citep{ooi2001distance}.

The closest designs set a 2-D cue against depth order. \citet{min2026whyfar} report an accuracy gap between items consistent with the cue that far objects sit higher in the frame and items that contradict it, which is the depth-order case of our \sgap{}. \citet{zou2026seeingdepth} set apparent size and height in the frame against depth on renders and photographs. They show that depth order stays decodable when the answer is wrong, and they correct some answers by writing the model's own depth readout back into its object representations. Our \family{closer} edits in Section~\ref{sec:res-probe} repeat their result on our scenes. What is new here is the extension of the conflict from depth order to four relations that compare sizes, contacts and alignments across depth. We also add inputs that keep the information fixed while its form changes, each checked by a solver. The drawn maps are scored with the same statements. On photographs we extend the depth-order split of \citet{min2026whyfar} on CV-Bench to the frontier models and to relative distance (Appendix~\ref{app:relwork} covers the remaining benchmarks).

\paragraph{Supplied inputs and maps.} Depth helps VLMs more as text than as an image \citep{vo2026dop}, depth-layered captions also help \citep{roy2025bydeway}, and models given naively fused geometry tokens still lean on 2-D cues \citep{zhang2026geosr}. Coordinates are hard to use without explicit association of objects and names \citep{cheng2024spatialrgpt,yuan2026gr3d}. Bird's-eye-view prompting is an established lever \citep{zhu2025struct2d,qi2026gpt4scene}, although MindCube finds that a ground-truth map helps only when the model reasons over it \citep{wang2026mindcube}. These interventions change the information and its form together. Because our solver measures how far each input determines the answer, a gap that remains under a sufficient input can be attributed to the form in which the information arrives.

\paragraph{Probing.} A linear probe shows what a representation makes available to a simple reader without showing that the model uses it \citep{alain2016probes,belinkov2022probing}. Control tasks test how selective a probe is \citep{hewitt2019control}, and a reference reader of the input tells what the representation adds \citep{hewitt2021conditional}. In language models, linear probes recover truth and world coordinates \citep{gurnee2024space,marks2024geometry}, sometimes beyond what the answers express \citep{orgad2025llms}. In VLMs the language model often fails to use what the vision encoder provides \citep{fu2025hidden,liu2026seeing}. Image information reaches the object and text tokens at identifiable layers \citep{neo2025interpreting,zhang2025crossmodal,kaduri2025whats}. Attention to the wrong regions is associated with some spatial errors \citep{chen2025attention}, and a latent map of the 3-D scene can be extracted and steered \citep{wang2026synspat3d}.

Our probes build on two studies. \citet{min2026whyfar} probe spatial representations in VLMs and \citet{zou2026seeingdepth} decode depth order in models whose answers are wrong, from object features averaged over each object's mask. Our object features follow the construction of \citet{zou2026seeingdepth}. \citet{liao2026decodable} show that probes and steering can overstate what a VLM takes from the image. We probe the full ground-plane layout and compare it with a reader of the 2-D measurements and with the same model's drawn map. We score all three with the same statements. We also compare a probe at the last token of the prompt with an input-only reader of the 2-D measurements.

\section{The \bench{} benchmark}
\label{sec:bench}

\bench{} consists of rendered scenes paired with true or false statements.

\subsection{Scenes}

Each scene holds four to eight convex primitives (cubes, spheres, cylinders and cones) standing on a grey ground plane, lit by one sun lamp and rendered at $768\times768$ with Blender Cycles \citep{blender} from a perspective camera 8 to 14 units away. Every object has its own colour, so that ``the red cube'' names one object. Sizes are drawn independently of shape, which leaves familiar sizes uninformative. Following CLEVR, PTR and Super-CLEVR-3D \citep{johnson2017clevr,hong2021ptr,wang2023superclevr3d}, the truth of every statement is read from the scene graph, while the statements take the true/false format of NLVR \citep{suhr2017nlvr,suhr2019nlvr2}. Instance masks come from an analytic ray caster over the same primitives, which makes the 2-D cues exact (generator settings in Appendix~\ref{app:details}).

\subsection{Relations, truth and cues}

Each of the five relations in Table~\ref{tab:families} has a 3-D truth computed from the scene graph and a 2-D cue computed from the visible masks. Candidates close to the decision threshold fall into a reject band and are discarded, so that every statement is clearly true or false in the scene geometry. Half of the statements are negated and each template has three paraphrases, which leaves the wording uninformative about the label. The default prompt does not state the thresholds.

\begin{table}[t]
\centering\small
\caption{The five relations. Truth is read from the scene graph and the 2-D cue from the rendered view's visible masks, and candidates inside the reject band are discarded. The relation \family{closer} is the control. $s$ is size, $h$ height. In \family{between}, C is the middle object. In \family{fits}, A is the object that passes between B and C. Formal definitions and hand-checked cases are in Appendix~\ref{app:audit}.}
\label{tab:families}
{\setlength{\tabcolsep}{4pt}\begin{tabular}{@{}l>{\raggedright\arraybackslash}p{0.27\textwidth}>{\raggedright\arraybackslash}p{0.27\textwidth}>{\raggedright\arraybackslash}p{0.235\textwidth}@{}}
\toprule
relation & 3-D truth & 2-D cue (balanced) & reject unless \\
\midrule
\family{closer} (control) & camera depth of A's centre $<$ B's & A covers more pixels than B & depth ratio $\ge 1.15$ \\
\family{taller} & height(A) $>$ height(B) & pixel height of A $>$ B & height ratio $\ge 1.25$ \\
\family{touching} & surface distance $d < 0.02$ & visible masks within 3\,px & $d<0.02$ or $d>0.5\min(s_A,h_A,s_B,h_B)$ \\
\family{between} & $\max(\angle CAB, \angle CBA) \le 10^\circ$ on the ground & same test on visible-mask centroids ($20^\circ$) & $\le10^\circ$ or $\ge30^\circ$ \\
\family{fits} & footprint gap(B,C) $\ge$ width(A) & pixel gap(B,C) $\ge$ pixel width(A) & ratio $\ge1.2$ or $\le0.8$ \\
\bottomrule
\end{tabular}}
\end{table}

The relation \family{closer} serves as a control, because its natural trap, that lower in the frame means nearer, is almost always right for objects standing on a ground plane seen from above. It agrees with the truth on all 480 \family{closer} items of the benchmark. We balance \family{closer} on pixel area instead, the size cue that \citet{zou2026seeingdepth} found models follow, and read a zero gap on it only as evidence that pixel area is not used.

\subsection{Balancing}
\label{sec:balancing}

For each relation we draw scenes and enumerate statements until the four cells, true or false crossed with cue-right or cue-wrong, each hold exactly a quarter of the items, with at most two statements per scene. Several cue-wrong cells almost never arise from random placement, so the sampler is steered to build them. It may hide a contact behind a third object, or place a near gap seen side-on next to a wider object farther away. These constructions change which scenes are drawn but never the labels, and they are recorded so that the analysis can control for them. As a result, every cue-wrong \family{touching} item is an occlusion case. Each of three tiers (easy, medium and hard) has 600 items, with a second medium split drawn from an independent seed. The tiers differ in clutter, camera elevation, shadows and occlusion.

\paragraph{Premise checks.} The cue scores exactly 50\% on every split and is right or wrong on exactly the intended items, which a unit test verifies. Under random placement on the medium tier the cues of \family{taller}, \family{touching}, \family{between} and \family{fits} agree with the truth on 82--99\% of candidate statements, so a model that leans on them would be behaving sensibly on natural images \citep{landy1995cue,knill1996perception}. Blender renders of per-object masks from the meshes match the analytic masks with a median per-object intersection over union (IoU) of \n{render.iou}. Relabelling the \n{render.nflip} items whose cue flips under the renders moves no gap by more than \n{r5.render.maxshift} points (Appendix~\ref{app:audit}).

\section{Measurements}
\label{sec:protocol}

\begin{figure}[t]
\centering
\includegraphics[width=\textwidth]{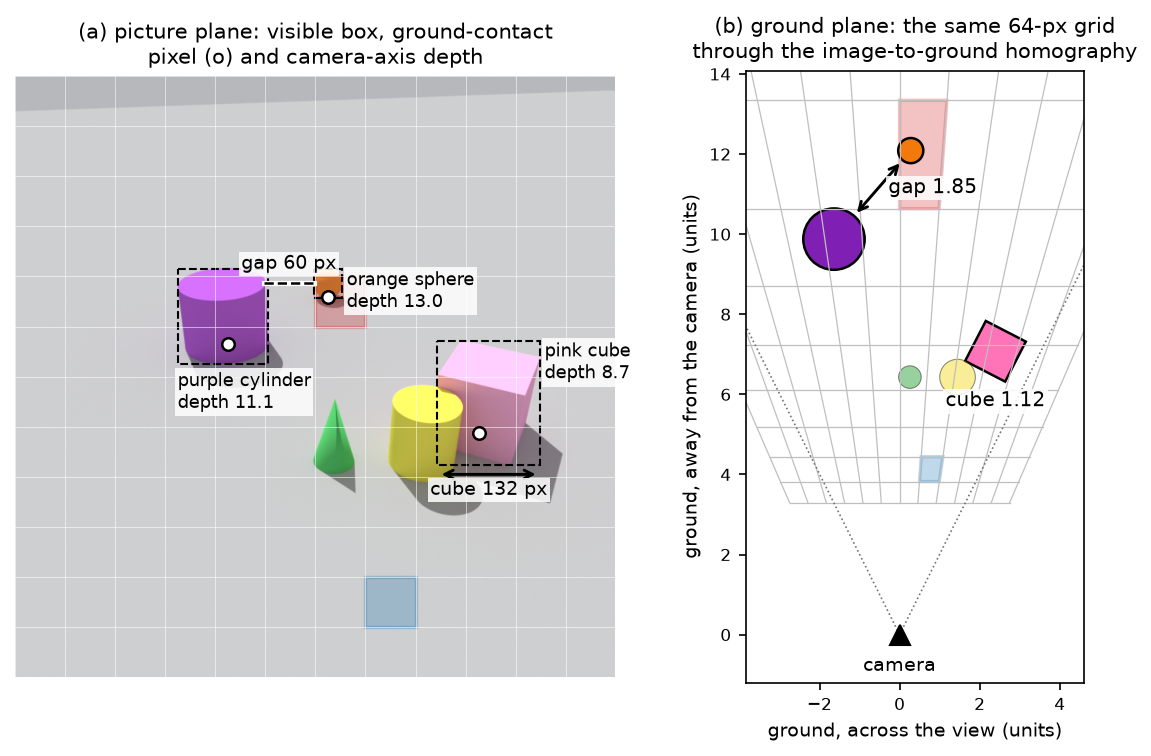}
\caption{The perspective form of the running example. Equal cells in the picture cover very different areas of ground. (a) The picture with a 64-pixel grid and the prompts' visible box, ground-contact pixel and camera depth for the three objects in the statement. (b) The grid carried to the ground plane through the exact image-to-ground homography, the projective map from pixels to ground points. The two shaded cells are equal in the picture, but the far cell covers 2.94 square units of ground and the near cell 0.32. The gap, farther from the camera than the cube and seen at a slant, spans 60\,px against the cube's 132\,px although it is wider on the ground.}
\label{fig:perspective}
\end{figure}

\paragraph{Models and prompts.} Each item pairs one image with one statement, under a prompt that says the statement concerns the 3-D scene and asks for True or False. This default input, the picture and the statement only, is the image condition. Unparseable answers count as wrong. The frontier models, Claude Opus 5.5, GPT-5.5 and Gemini 3.6 Flash, run at low reasoning effort on the image condition of every tier (Table~\ref{tab:main}) and on every medium-tier input. They are the only models run on every input, so the main analyses use them. Grok 4.7, Nova 2 Pro, Pixtral Large, Qwen3-VL-235B, Llama 4 Maverick and Gemma 3 27B run at the provider default on the inputs of Table~\ref{tab:grid}. Grok 4.7 behaves like the frontier models on the image condition, and Figure~\ref{fig:heatmap} shows it beside them.

We do not interpret the gaps of a model whose accuracy stays below 70\% even with ground coordinates, since its failure to reason about the relations is then general. We call this the coordinates rule. All calls use provider-default sampling. Low reasoning effort is the \texttt{low} setting of each provider's reasoning control. Appendix~\ref{app:prompts} prints every prompt and Appendix Table~\ref{tab:manifest} lists the model identifiers.

\paragraph{The \sgap{}.} With truth $Y$, cue $H$ and answer $M$, the gap is $g = P(M{=}Y \mid H{=}Y) - P(M{=}Y \mid H{\ne}Y)$, reported in percentage points (points). Because the cue is right on half of the items, a gap of $g$ means that the answers agree with the cue on $50 + g/2$\% of them. GPT-5.5's medium gap of \n{Gpt.medium.image.gap} points, for example, corresponds to agreement on \n{Gpt.medium.image.cuealign}\%. The gap without \family{between} and \family{touching} is the mean of the \family{closer}, \family{taller} and \family{fits} gaps. Per-relation gaps rest on 120 items per tier, so their 95\% intervals are about $\pm15$ points.

A positive gap does not by itself prove reliance on the cue, since an observer with noisy but faithful geometry would also show one if cue-wrong items were harder for other reasons. We address this by adjusting the gap for item difficulty with a penalised logistic regression \citep{firth1993bias} and, more directly, with the input forms. An observer that corrects the picture only part of the way toward the scene, as human observers do \citep{thouless1931phenomenal,koenderink2001ambiguity}, would also show a positive gap. The gap measures how far the answers follow the picture, and Section~\ref{sec:res-maps} measures how far each model's layout corrects it. Intervals are 95\% bootstrap intervals, resampled over scenes for paired comparisons \citep{efron1979bootstrap}. Appendix~\ref{app:details} lists the covariates and the correction for multiple comparisons.

\paragraph{Input forms.} On the medium tier we present the same scene in further inputs, listed in full in Appendix Table~\ref{tab:conditions}. Pixel measurements, the perspective form, describe the picture exactly. They list as text each object's visible box, centroid, ground-contact pixel and visible area, optionally with its camera depth, the depth along the camera axis (Figure~\ref{fig:perspective}). The perspective form leaves the conversion from pixels and depths to ground distances to the model. One variant adds a depth map as a second image. The ground-plane form describes the scene instead, either as a to-scale top-down map or as ground coordinates.

\paragraph{Other inputs.} Further prompts state the numeric definition of each relation, so that an error cannot come from a looser reading of a word such as ``between''. Other inputs change the prompt wording with stated thresholds or verbal definitions, or ask for chain-of-thought (CoT). High reasoning effort uses a 16{,}384-token reasoning budget. In the tool input Opus 5.5 gets a zoom tool and a Python interpreter over the pixels. A second camera view shows the scene from another angle. Appendix Table~\ref{tab:conditions} lists each input, and Appendix Table~\ref{tab:effort} gives the effort runs.

\paragraph{Reference readers.} For every text input, a deterministic solver reads exactly the text the model saw and applies the benchmark's rule, which shows whether the input determines the answer. The size-constancy reader scales each pixel width and gap by its image distance below the horizon \citep{sedgwick1986space,ooi2001distance} and makes no correction for foreshortening.

The measures that only the maps, the photographs and the open models need are defined where they are used, in Sections~\ref{sec:res-maps}, \ref{sec:res-photo} and~\ref{sec:res-probe}.

\section{Results}
\label{sec:results}

\begin{table}[t]
\centering\small
\caption{Accuracy and \sgap{} (points, 95\% CI) on the image condition, five relations pooled. Each tier has 600 items and one run per model. The frontier models ran at low reasoning effort and the other models at the provider default. Random is a random-answer baseline. The frontier models' answers track the picture plane on every tier. The 2-D cue scores 50 / +100 by construction. The last column of each tier is the mean gap over \family{closer}, \family{taller} and \family{fits}, which leaves out \family{between} and \family{touching}.}
\label{tab:main}
\resizebox{\textwidth}{!}{\setlength{\tabcolsep}{3pt}\begin{tabular}{l rlc rlc rlc}
\toprule
 & \multicolumn{3}{c}{easy} & \multicolumn{3}{c}{medium} & \multicolumn{3}{c}{hard} \\
\cmidrule(lr){2-4}\cmidrule(lr){5-7}\cmidrule(lr){8-10}
model & acc & gap [95\% CI] & \shortstack{w/o\\\family{between},\\\family{touching}} & acc & gap [95\% CI] & \shortstack{w/o\\\family{between},\\\family{touching}} & acc & gap [95\% CI] & \shortstack{w/o\\\family{between},\\\family{touching}} \\
\midrule
2-D cue & 50.0 & +100.0 & -- & 50.0 & +100.0 & -- & 50.0 & +100.0 & -- \\
Random & 52.0 & +3.3 [$-$5, +11] & -- & 51.0 & $-$4.7 [$-$13, +3] & -- & 50.7 & $-$2.7 [$-$10, +5] & -- \\
Claude Opus 5.5 & 84.5 & +19.7 [+14, +25] & +7.2 & 85.7 & +21.3 [+16, +27] & +11.7 & 84.2 & +17.7 [+12, +23] & +8.9 \\
GPT-5.5 & 75.8 & +37.0 [+31, +43] & +29.4 & 78.5 & +35.0 [+29, +41] & +27.2 & 76.0 & +28.7 [+22, +35] & +27.2 \\
Gemini 3.6 Flash & 81.2 & +24.3 [+18, +30] & +16.7 & 81.7 & +24.7 [+19, +30] & +20.6 & 78.5 & +23.7 [+18, +30] & +22.2 \\
Pixtral Large & 60.5 & +11.0 [+3, +20] & -- & 60.5 & +13.7 [+6, +22] & -- & 56.8 & +15.7 [+8, +23] & -- \\
Qwen3-VL-235B & 66.8 & +26.3 [+19, +34] & -- & 69.8 & +23.7 [+16, +31] & -- & 69.0 & +24.7 [+18, +32] & -- \\
Llama 4 Maverick & 53.2 & +15.7 [+8, +23] & -- & 56.5 & +9.0 [+1, +16] & -- & 54.3 & +12.7 [+5, +20] & -- \\
Gemma 3 27B & 50.3 & +10.0 [+2, +18] & -- & 51.8 & +15.0 [+7, +23] & -- & 53.7 & +11.3 [+4, +19] & -- \\
Grok 4.7 & 72.8 & +31.0 [+25, +38] & +25.6 & 76.5 & +33.0 [+27, +39] & +26.1 & 76.3 & +31.3 [+25, +38] & +28.3 \\
Nova 2 Pro & 60.2 & +13.7 [+6, +22] & -- & 62.5 & +26.3 [+19, +34] & -- & 63.8 & +17.0 [+9, +24] & -- \\
\bottomrule
\end{tabular}
}

\vspace{1.4em}
\caption{Accuracy (\%) / \sgap{} (points) under each input on the medium tier, one run on 600 items. The ground-plane form (top-down map, coordinates) brings the frontier models to gaps of +4 to +6 and Grok 4.7 to +5 to +18, while the five others stay at low accuracy. Frontier models used low reasoning effort, others the provider default. Columns marked + add that input to the image. Text only and coordinates only drop it. Camera depth and ground coordinates are text. The second view uses the 465 items with the same truth in both views and its own camera's cue (Appendix Table~\ref{tab:twoviewgap}). Appendix Table~\ref{tab:conditions} gives frontier and Pixtral Large CIs.}
\label{tab:grid}
\resizebox{\textwidth}{!}{\setlength{\tabcolsep}{2.5pt}\begin{tabular}{lcccccccccc}
\toprule
model & image & + CoT & \shortstack{text\\only} & \shortstack{+ pixel\\measure-\\ments} & \shortstack{+ depth\\map} & \shortstack{+ camera\\depth\\(text)} & \shortstack{+ top-\\down\\map} & \shortstack{+ coor-\\dinates} & \shortstack{coor-\\dinates\\only} & \shortstack{second\\view} \\
\midrule
2-D cue & 50 / +100 & -- & -- & -- & -- & -- & -- & -- & -- & -- \\
Claude Opus 5.5 & 86 / +21 & 86 / +17 & 50 / $-$4 & 86 / +15 & 87 / +13 & 89 / +18 & 93 / +4 & 96 / +5 & 96 / +6 & 86 / +3 \\
GPT-5.5 & 78 / +35 & 78 / +33 & 50 / +1 & 80 / +31 & 79 / +30 & 89 / +16 & 93 / +5 & 94 / +4 & 92 / +6 & 83 / +30 \\
Gemini 3.6 Flash & 82 / +25 & 81 / +25 & 46 / +2 & 80 / +24 & 80 / +21 & 84 / +18 & 92 / +5 & 93 / +5 & 93 / +4 & 83 / +9 \\
Pixtral Large & 60 / +14 & 62 / +22 & 50 / +1 & 56 / +23 & 60 / +9 & 64 / +3 & 58 / +9 & 59 / +14 & 63 / +5 & 57 / +16 \\
Qwen3-VL-235B & 70 / +24 & 72 / +18 & 49 / +3 & 64 / +30 & 68 / +22 & 68 / +15 & 66 / +17 & 68 / +15 & 56 / +8 & 68 / +24 \\
Llama 4 Maverick & 56 / +9 & 70 / +27 & 52 / $-$1 & 55 / +20 & 52 / +11 & 58 / +0 & 51 / +2 & 65 / +3 & 55 / +0 & 58 / +1 \\
Gemma 3 27B & 52 / +15 & 58 / +25 & 48 / $-$1 & 52 / +15 & 52 / +17 & 55 / +1 & 55 / +6 & 52 / +9 & 55 / +8 & 53 / +12 \\
Grok 4.7 & 76 / +33 & 74 / +32 & 49 / +1 & 53 / +28 & 68 / +33 & 62 / +24 & 80 / +18 & 91 / +5 & 82 / +10 & 75 / +21 \\
Nova 2 Pro & 62 / +26 & 70 / +25 & 48 / $-$4 & 55 / +22 & 58 / +20 & 62 / +16 & 57 / +12 & 63 / +12 & 57 / +5 & 64 / +18 \\
\bottomrule
\end{tabular}
}
\end{table}

\subsection{Do the answers follow the picture? (RQ1)}
\label{sec:res-image}

The answers of the frontier models follow the picture, most clearly on \family{fits}. On the image condition these models reach \n{main.accrange}\% accuracy with gaps of \n{main.gaprange} points (Table~\ref{tab:main}). The pattern is plainest for GPT-5.5, which on the medium tier is \n{Gpt.medium.image.conc}\% correct on cue-right items and \n{Gpt.medium.image.disc}\% correct on cue-wrong items. The gaps hold on the independently generated split (\n{r6.s11.gaprange}) and move by about a point across three repeated runs (Appendix Tables~\ref{tab:seed} and \ref{tab:repeats}).

Grok 4.7 behaves like the frontier models. The five other models score lower. Nova 2 Pro, Pixtral Large, Llama 4 Maverick and Gemma 3 27B score 51.8--62.5\% on the medium tier and Qwen3-VL-235B scores \n{Qwen.medium.image.acc}\%. None of them exceeds 68\% even with ground coordinates, which points to a general failure to reason about the relations. By the coordinates rule of Section~\ref{sec:protocol} we do not interpret their gaps. No model beats chance on text alone (\n{all.textonly.accrange}\%).

\begin{figure}[t]
\centering
\includegraphics[width=0.62\textwidth]{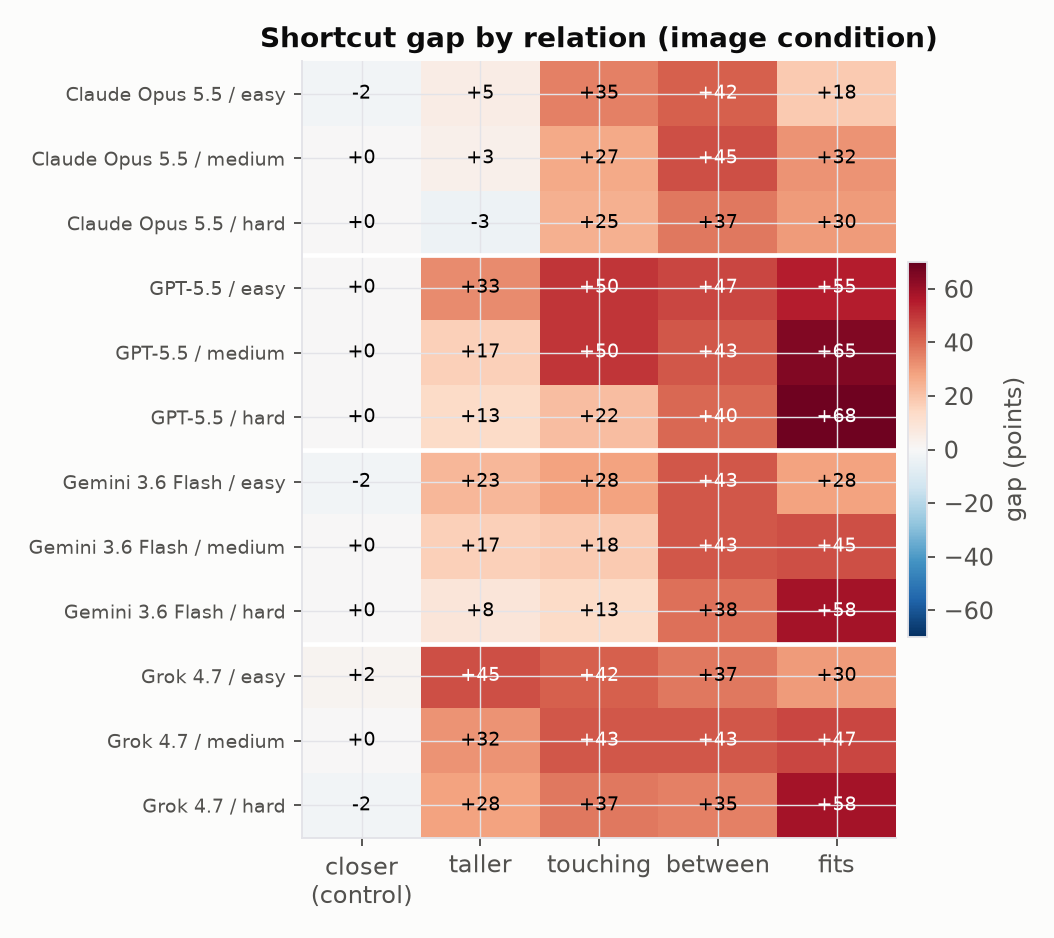}
\caption{The \sgap{} by relation (points, image condition) for the frontier models and Grok 4.7, one row per model and tier. The gap is concentrated in \family{fits}, \family{between} and \family{touching}. Each cell prints its gap, and darker red marks a larger positive gap. Each cell has 120 items and one run, the frontier models at low reasoning effort and Grok 4.7 at the provider default. The relation \family{closer} is the control.}
\label{fig:heatmap}
\end{figure}

The gap is concentrated in the relations that compare sizes across depth or depend on what an occluder hides (Figure~\ref{fig:heatmap}). It reaches its largest values for \family{fits}, at \n{main.fits.gaprange} points. On the medium tier GPT-5.5 answers the cue-wrong half of \family{fits} at \n{Gpt.medium.image.fits.disc}\%, worse than a coin, just as it does on the item of Figure~\ref{fig:running}. The gaps for \family{between} and \family{touching} follow at \n{main.between.gaprange} and \n{main.touching.gaprange}, while the control relation \family{closer} stays at \n{main.closer.gaprange}. The pooled gap therefore leans on these two relations. For Opus 5.5 they supply \n{r10.opus.bt.share}\% of the summed per-relation gaps. Without them the frontier gaps are \n{r10.clean.gaprange} points, and \n{r10.clean.Opus.range} for Opus 5.5, which shows no \family{taller} gap (\n{r10.taller.Opus.range} across the three tiers and the second medium split). For Opus 5.5 the picture-plane errors that we can interpret are concentrated in \family{fits}.

\begin{table}[!t]
\centering
\caption{Accuracy (\%) on the four \family{fits} cells of each tier, split by whether the pixel-width cue is right and whether the object fits (30 items per cell, one run at low reasoning effort). The size-constancy reader of Section~\ref{sec:protocol} scales each pixel width and gap by its image distance below the horizon and makes no correction for foreshortening. It fails where the object fits but looks too wide, because the gap then runs along the line of sight. GPT-5.5 and Gemini fall short of it on the cue-wrong items where the object does not fit, the cell that tests size scaling.}
\label{tab:fitscells}
\small
\begin{tabular}{llcccc}
\toprule
 & & \multicolumn{2}{c}{cue-right} & \multicolumn{2}{c}{cue-wrong} \\
\cmidrule(lr){3-4}\cmidrule(lr){5-6}
tier & reader & fits & does not fit & fits & does not fit \\
\midrule
easy & size-constancy reader & 97 & 100 & 17 & 97 \\
 & Claude Opus 5.5 & 100 & 90 & 73 & 80 \\
 & GPT-5.5 & 93 & 93 & 27 & 50 \\
 & Gemini 3.6 Flash & 97 & 93 & 70 & 63 \\
\midrule
medium & size-constancy reader & 97 & 100 & 3 & 93 \\
 & Claude Opus 5.5 & 100 & 100 & 53 & 83 \\
 & GPT-5.5 & 97 & 100 & 20 & 47 \\
 & Gemini 3.6 Flash & 100 & 90 & 57 & 43 \\
\midrule
hard & size-constancy reader & 100 & 100 & 7 & 87 \\
 & Claude Opus 5.5 & 100 & 97 & 50 & 87 \\
 & GPT-5.5 & 93 & 100 & 10 & 47 \\
 & Gemini 3.6 Flash & 100 & 97 & 47 & 33 \\
\bottomrule
\end{tabular}

\end{table}

The two cue-wrong halves of \family{fits} test different abilities (Table~\ref{tab:fitscells}). Where the object does not fit but looks narrower than the gap, it stands farther from the camera than the gap, and an observer that scales pixel extents by distance answers correctly. The size-constancy reader of Section~\ref{sec:protocol} is right on \n{r10.fitsrd.wf.range}\% of these items. Where the object fits but looks too wide, the gap usually runs along the line of sight and is foreshortened. The same reader is then right on only \n{r10.fitsrd.wt.range}\%, which gives it a gap of \n{r10.fitsrd.gap.range} points. On the first cell Opus 5.5 scores \n{r10.fitscell.Opus.wf.range}\% against the reader's \n{r10.fitsrd.wf.range}\%. On the second it is well above the reader (\n{r10.fitscell.Opus.wt.range}\%). GPT-5.5 is far below the reader on the first cell (\n{r10.fitscell.Gpt.wf.range}\%), as a lack of size scaling would produce. It also scores low on the second (\n{r10.fitscell.Gpt.wt.range}\%), where the reader fails too. Gemini is also below the reader on the first cell (\n{r10.fitscell.Gemini.wf.range}\%).

If cue-wrong items were simply harder, holding difficulty fixed would remove the gap. For \family{fits} it does not. Adjusting for distance from the decision threshold, occlusion, pixel area, clutter, camera elevation and the construction of each scene leaves medium-tier \family{fits} gaps of \n{Opus.medium.image.fits.adjgap} \n{Opus.medium.image.fits.adjgapci} for Opus 5.5, \n{Gpt.medium.image.fits.adjgap} \n{Gpt.medium.image.fits.adjgapci} for GPT-5.5 and \n{Gemini.medium.image.fits.adjgap} \n{Gemini.medium.image.fits.adjgapci} for Gemini. Every interval on the medium and hard tiers excludes zero, while on the easy tier those of Opus 5.5 and Gemini reach it (Appendix Table~\ref{tab:familyadj}). The pooled gaps after adjustment are \n{main.adjgap.range} points with every interval above zero (Appendix Table~\ref{tab:matched}), and the \family{fits} gap survives on the items whose covariate values occur among both cue-right and cue-wrong items (Appendix Table~\ref{tab:overlap}). The relation \family{between} is the exception. In its false half the cue-wrong items are exactly the items near the reject band, so no adjustment can separate the cue from closeness to the threshold. We treat \family{between} as confounded and \family{fits} as the case free of this confound (Appendix~\ref{app:further}).

High reasoning effort narrows the \family{fits} gap without removing it. It moves the pooled gap by \n{main.high16k.deltarange} points (Appendix Table~\ref{tab:effort}) but the \family{fits} gap by \n{r10.eff.fits.drange}. Gemini's change has an interval that reaches zero (\n{r10.eff.Gemini.fits.dci}). Opus 5.5's \family{fits} gap falls from \n{r10.eff.Opus.fits.low} to \n{r10.eff.Opus.fits.high} points, with a 95\% interval of \n{r10.eff.Opus.fits.dci} for the change. GPT-5.5's change has an interval of \n{r10.eff.Gpt.fits.dci}. Its accuracy on the cue-wrong \family{fits} items rises from \n{r10.eff.Gpt.fitsdisc.low}\% to \n{r10.eff.Gpt.fitsdisc.high}\%, so its below-chance score holds only at low effort.

\subsection{Is the information missing? (RQ2a)}
\label{sec:res-info}

\begin{table}[t]
\centering\small
\caption{Solver accuracy (\%) on the 120 medium items of each relation, which shows how far each text input determines the answer. With camera depths, pixel measurements determine \family{closer} and \family{between} exactly and \family{fits} nearly. The solver applies the benchmark's rule to exactly the text in the prompt. Rows with a single entry are inputs given for that relation only (Appendix~\ref{app:details}). Visible-region centres are the mask centroids. The homography maps each pixel to its ground point. Surface distances are given for each pair of objects. $^\dagger$: the input lacks a needed quantity (object sizes, the ground mapping), so the solver uses the best simple estimate with thresholds fitted on the easy tier.}
\label{tab:solvers}
\begin{tabular}{lrrrrr}
\toprule
input (text) & closer & taller & touching & between & fits \\
\midrule
visible-region centres (text) & -- & -- & -- & 72$^\dagger$ & -- \\
ground-contact pixels (text) & -- & -- & -- & 94$^\dagger$ & -- \\
pixel measurements (text) & 100$^\dagger$ & 50$^\dagger$ & 62$^\dagger$ & 94$^\dagger$ & 65$^\dagger$ \\
pixel measurements + camera depth (text) & 100 & 69$^\dagger$ & 96$^\dagger$ & 100 & 97$^\dagger$ \\
ground-contact pixels + homography (text) & -- & -- & -- & 100 & -- \\
pairwise surface distances (text) & -- & -- & 100 & -- & -- \\
ground coordinates (text) & 100 & 100 & 100 & 100 & 100 \\
\bottomrule
\end{tabular}

\end{table}

\begin{table}[t]
\centering\small
\caption{Gap (points) under each input on the medium tier and its paired change against the image condition, 95\% scene-bootstrap CI, one run per model on the 600 items at low reasoning effort. The two rows after the image row change only the prompt wording. Verbal definitions state each rule in words without the reject bands. The metric legend is a numeric colour bar on the depth map. The top-down map without labels drops the text labels on the footprints. The ground-plane form nearly removes the gap. With numeric definitions, ground coordinates give \n{Opus.medium.coordsnoimagendef.acc}, \n{Gpt.medium.coordsnoimagendef.acc} and \n{Gemini.medium.coordsnoimagendef.acc}\% accuracy. Pixel measurements narrow the gap in the point estimates, with several intervals reaching zero.}
\label{tab:ladder}
\resizebox{\textwidth}{!}{\begin{tabular}{lrlrlrl}
\toprule
 & \multicolumn{2}{c}{Claude Opus 5.5} & \multicolumn{2}{c}{GPT-5.5} & \multicolumn{2}{c}{Gemini 3.6 Flash} \\
\cmidrule(lr){2-3}\cmidrule(lr){4-5}\cmidrule(lr){6-7}
input & gap & $\Delta$ vs image [95\% CI] & gap & $\Delta$ vs image [95\% CI] & gap & $\Delta$ vs image [95\% CI] \\
\midrule
image & +21.3 & -- & +35.0 & -- & +24.7 & -- \\
stated thresholds & +11.7 & $-$9.7 [$-$14, $-$5] & +33.0 & $-$2.0 [$-$7, +3] & +22.0 & $-$2.7 [$-$7, +2] \\
verbal definitions (no bands) & +14.7 & $-$6.7 [$-$10, $-$3] & +33.3 & $-$1.7 [$-$7, +4] & +24.3 & $-$0.3 [$-$5, +5] \\
pixel measurements (text) & +15.0 & $-$6.3 [$-$11, $-$2] & +30.7 & $-$4.3 [$-$10, +1] & +24.3 & $-$0.3 [$-$6, +5] \\
pixel measurements, no image & +12.0 & $-$9.3 [$-$14, $-$4] & +19.0 & $-$16.0 [$-$22, $-$10] & +16.0 & $-$8.7 [$-$16, $-$1] \\
pixel measurements + camera depth (text) & +15.7 & $-$5.7 [$-$10, $-$2] & +22.3 & $-$12.7 [$-$19, $-$6] & +22.3 & $-$2.3 [$-$8, +4] \\
depth map & +13.3 & $-$8.0 [$-$12, $-$4] & +30.0 & $-$5.0 [$-$10, $-$0] & +21.0 & $-$3.7 [$-$9, +1] \\
depth map, metric legend & +16.3 & $-$5.0 [$-$9, $-$2] & +26.7 & $-$8.3 [$-$13, $-$4] & +22.7 & $-$2.0 [$-$8, +4] \\
camera depth (text) & +18.0 & $-$3.3 [$-$7, +0] & +15.7 & $-$19.3 [$-$24, $-$14] & +18.0 & $-$6.7 [$-$12, $-$2] \\
top-down map & +4.0 & $-$17.3 [$-$22, $-$12] & +5.3 & $-$29.7 [$-$36, $-$24] & +4.7 & $-$20.0 [$-$26, $-$14] \\
top-down map, no labels & +6.3 & $-$15.0 [$-$20, $-$10] & +7.3 & $-$27.7 [$-$34, $-$22] & +9.0 & $-$15.7 [$-$22, $-$10] \\
ground coordinates (text) & +4.7 & $-$16.7 [$-$21, $-$12] & +4.3 & $-$30.7 [$-$37, $-$25] & +5.0 & $-$19.7 [$-$26, $-$14] \\
ground coordinates, no image & +6.3 & $-$15.0 [$-$20, $-$10] & +6.3 & $-$28.7 [$-$34, $-$23] & +3.7 & $-$21.0 [$-$27, $-$15] \\
ground coordinates, no image, verbal definitions & +3.0 & $-$18.3 [$-$23, $-$13] & +1.7 & $-$33.3 [$-$39, $-$27] & +0.7 & $-$24.0 [$-$30, $-$17] \\
ground coordinates, no image, numeric definitions & +0.0 & $-$21.3 [$-$27, $-$16] & +0.0 & $-$35.0 [$-$40, $-$29] & $-$0.3 & $-$25.0 [$-$30, $-$19] \\
\bottomrule
\end{tabular}
}
\end{table}

\begin{table}[t]
\centering\small
\caption{The \family{fits} and \family{between} gaps (points) under the inputs of Section~\ref{sec:res-info} on the medium tier, 120 items per relation, one run per model at low reasoning effort. Pixel measurements and ground coordinates are given as text. Table~\ref{tab:solvers} gives how far each input determines the answer.}
\label{tab:perspgaps}
\resizebox{\textwidth}{!}{\begin{tabular}{lrrrrrr}
\toprule
 & \multicolumn{2}{c}{Claude Opus 5.5} & \multicolumn{2}{c}{GPT-5.5} & \multicolumn{2}{c}{Gemini 3.6 Flash} \\
\cmidrule(lr){2-3}\cmidrule(lr){4-5}\cmidrule(lr){6-7}
input & \family{fits} & \family{between} & \family{fits} & \family{between} & \family{fits} & \family{between} \\
\midrule
image & +32 & +45 & +65 & +43 & +45 & +43 \\
pixel measurements, with the image & +15 & +33 & +68 & +30 & +42 & +37 \\
pixel measurements, no image & $-$3 & +23 & +42 & +18 & +18 & +15 \\
pixel measurements + camera depths, with the image & +10 & +33 & +17 & +28 & +20 & +37 \\
ground coordinates, no image, no definitions & +0 & +28 & +0 & +23 & $-$2 & +10 \\
ground coordinates, no image, numeric definitions & $-$2 & +0 & +0 & +0 & +0 & +0 \\
\bottomrule
\end{tabular}
}
\end{table}

The information is not missing from the inputs, since exact pixel measurements with camera depths close most of the \family{fits} gap and the \family{between} gap that remains comes from how the models use the numbers. If the picture simply did not carry enough information, an input that does carry it should remove the gap. Pixel measurements test this. Given exact pixel measurements as text and no image, the models score \n{r6.pixnoimage.accrange}\% (Appendix Table~\ref{tab:conditions}), close to their accuracy on the image condition, which shows that they read and use the numbers. A zoom tool and a Python interpreter over the pixels move Opus 5.5's pooled gap by \n{Opus.agent.agentic.dgap} \n{Opus.agent.agentic.dgapci}, although the same model measures visible boxes accurately when that is all it is asked to do (median IoU \n{measure.iou}).

Adding each object's camera depth makes \family{closer} and \family{between} exactly solvable and \family{fits} solvable on \n{solver.pixdepthoracle.fits}\% of items, as the solver confirms (Table~\ref{tab:solvers}). Table~\ref{tab:ladder} gives the pooled gap under each input. Table~\ref{tab:perspgaps} gives the \family{fits} and \family{between} gaps. With pixel measurements and camera depths the \family{fits} gaps of Opus 5.5, GPT-5.5 and Gemini fall from \n{r10.img.fits} points to \n{r10.pd.fits}. The \family{between} gaps stay at \n{r6.between.pixdepth.gaprange}, although this input determines every \family{between} answer (Table~\ref{tab:solvers}). What closes the \family{fits} gap differs by model. For GPT-5.5 it is the camera depths. For Opus 5.5 the pixel measurements already bring most of the drop. For Gemini removing the image does as much as adding depths (Table~\ref{tab:perspgaps}). Most of the failure therefore lies before the arithmetic, in getting from the picture to measurements and depths.

The pixel-measurement inputs come with the image. The same measurements without the image give smaller \family{fits} gaps for every model (Table~\ref{tab:perspgaps}). Part of the residual may be the picture overriding the text. Pixel measurements with camera depths also raise some gaps above the image condition (\n{r10.pd.rises}). A depth map shown as a second image shifts the pooled gap by only \n{main.depthdelta.range} (Table~\ref{tab:ladder}).

The \family{between} residual shows what the models do with such numbers. Given ground-contact pixels and an exact angular rule for \family{between}, they apply the rule to image-plane angles. GPT-5.5's answers agree with that image-plane reading on \n{Gpt.r5.rule10.base.agree}\% of items. On this input they still score \n{r6.bl.base.accrange}\%, because the reject band leaves few items on which image-plane and ground-plane angles disagree. The models compute on the measurements as given and do not carry them onto the ground. Part of the residual is also the looser reading of the word. Ground coordinates without the numeric definition leave \family{between} gaps close to those under pixel measurements with camera depths for Opus 5.5 and GPT-5.5, though not for Gemini (Table~\ref{tab:perspgaps}). Stating the numeric definitions alongside the image moves the gap by \n{r6.fac.ndef.drange} points (Appendix Table~\ref{tab:factorial}). Adding the reject bands lowers Opus 5.5's pooled gap from \n{Opus.fac.image.gap} to \n{Opus.fac.ndefbands.gap} points. The bands add no geometry, although they reveal how the items were sampled.

\subsection{Can the models reason on the ground plane? (RQ2b)}
\label{sec:res-ground}

The models can apply the relations once the layout is given, because describing the scene on the ground nearly removes the gap. The top-down map narrows it to \n{main.topdownoracle.gaprange} points and ground coordinates without the image narrow it to \n{main.coordsnoimage.gaprange} (Table~\ref{tab:ladder}). When the prompt also states each relation's numeric definition (last row of Table~\ref{tab:ladder}), coordinates give \n{main.coordsnoimagendef.accrange}\% accuracy and no gap farther than \n{main.coordsnoimagendef.absgapmax} points from zero. This includes the contacts hidden behind other objects, which Opus 5.5 gets right on only \n{Opus.to.image.hidden}\% of items from the picture. The relations are within the models' reach once the layout is given, which leaves the recovery of the layout as the step that fails.

\subsection{What layout do the models recover? (RQ3)}
\label{sec:res-maps}

\begin{figure}[t]
\centering
\includegraphics[width=0.8\textwidth]{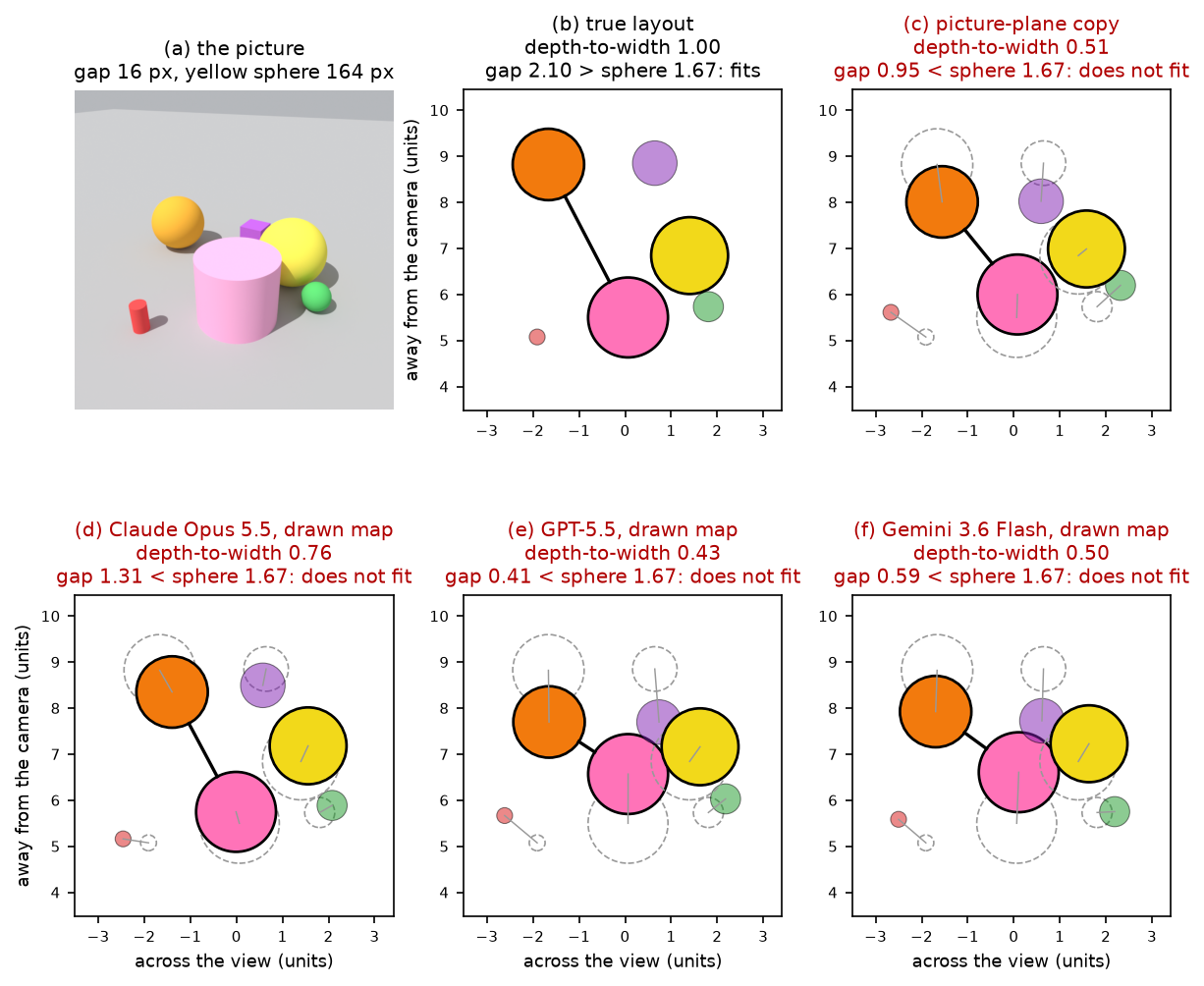}
\caption{One scene drawn from above by each model. The statement ``The yellow sphere would fit through the gap between the orange sphere and the pink cylinder'' is true. (a) In the picture the gap is 16\,px and the sphere 164\,px. (b) On the ground the gap is 2.10 units and the sphere 1.67. (c) The picture-plane copy. (d--f) Each frontier model's drawn map from the image alone. Every layout is aligned to the truth by a similarity transform and drawn with the true footprint sizes, with dashed circles at the true positions. Panel titles give each layout's depth-to-width ratio and gap width in units. Each map compresses depth and narrows the gap below the sphere's width.}
\label{fig:ownmap}
\end{figure}

\begin{table}[t]
\centering\small
\caption{Drawn maps, one run per model (frontier models at low reasoning effort, others at provider default). All frontier maps compress depth (ratio below 1). The constancy index puts GPT-5.5 lowest, Opus 5.5 highest. Section~\ref{sec:res-maps} defines the depth-to-width ratio (1 = undistorted), the constancy index and the share corrected (-- if not computed, Appendix Tables~\ref{tab:constancy} and \ref{tab:constancyfit}). The picture-plane copy's ratio is \n{r5.picturecopy.medium.ratio} (medium) and \n{r5.picturecopy.hard.ratio} (hard). RMSE is the median position error after alignment. Relation columns re-derive the statements from the map with true sizes, scoring those inside a reject band as wrong (accuracy in \%, map gaps in points). Heights are not mapped, so \family{taller} is left out. \% compressed counts scenes with a ratio below 1.}
\label{tab:map}
\resizebox{\textwidth}{!}{\setlength{\tabcolsep}{3pt}\begin{tabular}{lrrllrrrrr rrrr}
\toprule
 & & & & & & & \multicolumn{3}{c}{\shortstack{re-derived from\\the drawn map}} & \multicolumn{4}{c}{map gap by relation} \\
\cmidrule(lr){8-10}\cmidrule(lr){11-14}
model & \shortstack{scenes\\parsed} & \shortstack{RMSE\\(units,\\median)} & \shortstack{depth-to-width\\ratio [95\% CI]} & \shortstack{constancy index\\{}[95\% CI]} & \shortstack{share\\corrected} & \shortstack{\% com-\\pressed} & \shortstack{cue-\\right} & \shortstack{cue-\\wrong} & gap & closer & touching & between & fits \\
\midrule
\multicolumn{14}{l}{\emph{medium tier}} \\
Claude Opus 5.5 & 304/304 & 0.24 & 0.83 [0.80, 0.84] & 0.60 [0.58, 0.63] & 0.64 & 88 & 80 & 60 & +20 & $-$7 & +0 & +47 & +42 \\
GPT-5.5 & 304/304 & 0.46 & 0.61 [0.60, 0.65] & $-$0.01 [$-$0.07, 0.07] & 0.18 & 87 & 75 & 38 & +38 & +0 & +42 & +65 & +43 \\
Gemini 3.6 Flash & 304/304 & 0.44 & 0.63 [0.60, 0.65] & 0.07 [0.01, 0.12] & 0.58 & 96 & 74 & 43 & +30 & +3 & +17 & +55 & +47 \\
Pixtral Large & 303/304 & 0.86 & 0.62 [0.58, 0.65] & -- & -- & 86 & 48 & 33 & +15 & $-$8 & +23 & +33 & +15 \\
Qwen3-VL-235B & 304/304 & 0.78 & 0.38 [0.36, 0.39] & -- & -- & 100 & 54 & 27 & +27 & $-$13 & +40 & +47 & +33 \\
Llama 4 Maverick & 294/304 & 0.90 & 0.56 [0.51, 0.59] & -- & -- & 89 & 53 & 32 & +22 & $-$10 & +32 & +37 & +28 \\
Gemma 3 27B & 304/304 & 0.89 & 0.66 [0.61, 0.69] & -- & -- & 82 & 53 & 40 & +12 & $-$7 & +12 & +17 & +28 \\
Grok 4.7 & 224/304 & 0.39 & 0.75 [0.72, 0.79] & 0.37 [0.30, 0.45] & 0.60 & 83 & 70 & 53 & +18 & $-$3 & +4 & +45 & +23 \\
Nova 2 Pro & 304/304 & 0.86 & 0.40 [0.38, 0.44] & -- & -- & 98 & 52 & 30 & +23 & $-$5 & +25 & +48 & +23 \\
\midrule
\multicolumn{14}{l}{\emph{hard tier}} \\
Claude Opus 5.5 & 302/302 & 0.43 & 0.64 [0.63, 0.66] & 0.63 [0.61, 0.65] & 0.64 & 98 & 73 & 50 & +23 & +2 & +13 & +35 & +42 \\
GPT-5.5 & 302/302 & 0.78 & 0.35 [0.33, 0.37] & 0.12 [0.08, 0.15] & 0.18 & 97 & 57 & 25 & +32 & +8 & +40 & +55 & +23 \\
Gemini 3.6 Flash & 301/302 & 0.71 & 0.47 [0.46, 0.50] & 0.37 [0.35, 0.41] & 0.58 & 99 & 64 & 33 & +30 & +9 & +30 & +51 & +30 \\
Pixtral Large & 302/302 & 1.11 & 0.44 [0.41, 0.46] & -- & -- & 94 & 43 & 23 & +20 & +0 & +30 & +40 & +10 \\
Qwen3-VL-235B & 301/302 & 1.11 & 0.20 [0.19, 0.21] & -- & -- & 100 & 44 & 13 & +31 & +4 & +48 & +60 & +10 \\
Llama 4 Maverick & 283/302 & 1.10 & 0.37 [0.34, 0.40] & -- & -- & 96 & 46 & 21 & +25 & +6 & +32 & +48 & +12 \\
Gemma 3 27B & 301/302 & 1.17 & 0.48 [0.44, 0.50] & -- & -- & 94 & 45 & 28 & +18 & +9 & +25 & +31 & +5 \\
Grok 4.7 & 206/302 & 0.63 & 0.54 [0.52, 0.58] & 0.50 [0.43, 0.55] & 0.60 & 87 & 63 & 42 & +21 & $-$2 & +28 & +37 & +26 \\
Nova 2 Pro & 302/302 & 1.13 & 0.24 [0.22, 0.25] & -- & -- & 100 & 44 & 21 & +23 & +5 & +33 & +43 & +12 \\
\bottomrule
\end{tabular}
}
\end{table}

The frontier models recover a layout that keeps most of the depth order but compresses depth toward the picture, by an amount that differs between models.

\paragraph{Task and measures.} In a separate task the model sees only the image and returns every object's ground position and footprint diameter as JSON, in a top-down frame of its choosing. We call the result its drawn map. We align each drawn map to the truth by a similarity transform, a shift, rotation and uniform scaling, and re-derive the statements from the aligned positions. The picture-plane copy, the ground-contact pixels read as top-down coordinates, is aligned in the same way. Each frontier model's drawn map is also fed back to it as text, either as drawn or with its depth axis stretched by a factor estimated on the other half of the scenes. The gap of the statements re-derived from a map is its map gap.

A map's depth-to-width ratio is the spread of its positions along the camera axis over their spread across it, relative to the truth. A ratio below 1 means that depth is compressed. The constancy index of \citet{thouless1931phenomenal} is the median over scenes of $\ln(m/c)/\ln(1/c)$, with $m$ the map's depth-to-width ratio and $c$ the copy's. It is 0 for the picture-plane copy and 1 for the scene. It is the part of the picture's compression that a typical map undoes, measured on a log scale. A second measure fits $\ln m = a + b \ln c$ over each model's medium and hard scenes. The share corrected is $1-b$, the part of the picture's compression that the map undoes at any level of compression. The factor $e^{a}$ is any further uniform shrinking of depth, which lowers the index but not the share corrected. We also count the scenes in which the aligned map lies closer to the picture-plane copy than to the scene.

The drawn maps put most objects in the right depth order, with median rank correlations along the camera axis of at least \n{Gemini.map.medium.rankdepth} on the medium tier (Appendix~\ref{app:further}). They compress depth relative to width (Table~\ref{tab:map}). Statements about which object is closer, re-derived from the maps, are right on \n{Opus.map.medium.closer.disc}, \n{Gpt.map.medium.closer.disc} and \n{Gemini.map.medium.closer.disc}\% of the cue-wrong items on the medium tier for Opus 5.5, GPT-5.5 and Gemini. On the hard tier they are right on \n{Opus.map.hard.closer.disc}\% for Opus 5.5. For GPT-5.5 and Gemini they are right on \n{Gpt.map.hard.closer.disc}\% and \n{Gemini.map.hard.closer.disc}\% of cue-wrong items, although the depth rank correlations of their maps stay at \n{Gpt.map.hard.rankdepth} and \n{Gemini.map.hard.rankdepth}. On the medium tier GPT-5.5's and Gemini's maps keep \n{Gpt.map.medium.ratio} and \n{Gemini.map.medium.ratio} of the true depth-to-width ratio, close to the \n{r5.picturecopy.medium.ratio} of the picture-plane copy, and Opus 5.5's maps keep \n{Opus.map.medium.ratio}.

Figure~\ref{fig:ownmap} shows a scene in which every map narrows a gap below the width of an object that in fact fits through it. All three models answer False from the image and True from coordinates or the top-down map. Among the 60 medium scenes with a cue-wrong \family{fits} item and parsed maps from all three models, 5 have that item wrong from the image and from every map. The scene of Figure~\ref{fig:ownmap} is the only one of the five whose statement is true.

The constancy index orders the frontier models the same way on both tiers, with GPT-5.5 lowest and Opus 5.5 highest. GPT-5.5's maps score \n{ti.Gpt.medium} \n{ti.Gpt.medium.ci} on the medium tier and \n{ti.Gpt.hard} \n{ti.Gpt.hard.ci} on the hard tier. Gemini's score \n{ti.Gemini.medium} \n{ti.Gemini.medium.ci} and \n{ti.Gemini.hard} \n{ti.Gemini.hard.ci}, and Opus 5.5's \n{ti.Opus.medium} \n{ti.Opus.medium.ci} and \n{ti.Opus.hard} \n{ti.Opus.hard.ci} (Table~\ref{tab:map}). The picture-plane copy keeps half as much depth on the hard tier (copy ratios \n{ti.copy.medium} and \n{ti.copy.hard}).

The fit of $\ln m$ on $\ln c$ separates three patterns (Appendix Table~\ref{tab:constancyfit}). Opus 5.5 corrects a fixed share of about \n{r10.cf.Opus.k} of the compression with no further bias (intercept \n{r10.cf.Opus.a} on the log scale). GPT-5.5 corrects only \n{r10.cf.Gpt.k}. Gemini corrects a share like that of Opus 5.5 (\n{r10.cf.Gemini.k}) but shrinks every map in depth by a further factor of \n{r10.cf.Gemini.factor} (intercept \n{r10.cf.Gemini.a} \n{r10.cf.Gemini.aci} on the log scale), which on the medium tier happens to land on the copy. Distances to the copy and to the scene also separate Opus 5.5 from the other two. After similarity alignment, GPT-5.5's maps lie closer to the picture-plane copy than to the scene on \n{r10.pos.Gpt.medium}\% and \n{r10.pos.Gpt.hard}\% of medium and hard scenes, Gemini's on \n{r10.pos.Gemini.medium}\% and \n{r10.pos.Gemini.hard}\%, and Opus 5.5's on only \n{r10.pos.Opus.medium}\% and \n{r10.pos.Opus.hard}\%.

Human observers show a similar partial correction, known as phenomenal regression \citep{thouless1931phenomenal}. Monocular judgements of the shape of objects on the ground fall roughly midway between the retinal and the true values, an index near 0.5 by our reading of Figure~2 of \citet{loomis1999anisotropy}. Opus 5.5's maps have a constancy index close to this value. GPT-5.5's maps have a far lower index. People also compress depth relative to width in real scenes \citep{loomis1992visual}, and in pictures they recover depth only up to a scaling and shear \citep{koenderink1992surface,koenderink2001ambiguity}.

The maps reproduce the gap in aggregate but not item by item. On the medium tier, statements re-derived from them show gaps of \n{map.medium.gaprange}. They agree with the answers given from the image on \n{mi.agreerange}\% of items (Appendix Table~\ref{tab:mapitems}). On the cue-wrong \family{fits} items, though, the two are only weakly related (correlations of \n{r10.map.phi.range} on 60 items per model). Part of the map gap also comes from scoring as wrong the statements whose re-derived relation falls inside the reject band, which happens on \n{r10.map.band.disc}\% of cue-wrong items and \n{r10.map.band.conc}\% of cue-right ones. Without these items the map gaps are \n{r10.map.exband.range}, and Opus 5.5's \family{between} map gap falls from \n{r10.map.Opus.btw} to \n{r10.map.Opus.btw.ex}. The answers and the maps may be two separate outputs that each lean on the picture.

When a model reads its drawn map back as text, it answers as a solver reading that map would on \n{Opus.om.agree}\% of items. Stretching each map's depth axis by a factor estimated on the other half of the scenes changes the \family{fits} gap by \n{r6.om.fits.drange} points and the \family{between} gap by \n{r6.om.between.drange} (Appendix~\ref{app:further}). This suggests that uniform compression of depth explains part of the \family{fits} errors and none of the \family{between} errors.

Among the other models, Grok 4.7's maps correct a share of \n{r10.cf.Grok.k} of the compression, close to Opus 5.5's \n{r10.cf.Opus.k}, but show a gap of only \n{r10.map.Grok.gap} against \n{r10.grok.image.gap} for its answers. The five other models' maps order depth less reliably, and the Qwen3-VL-235B maps run depth the wrong way (Appendix~\ref{app:further}). The frontier models recover most of the depth order but compress depth. GPT-5.5 stays close to the picture.

\subsection{Do photographs show the same gap? (RQ4)}
\label{sec:res-photo}

\begin{table}[!t]
\centering
\caption{The \sgap{} on the 3-D subset of CV-Bench, in points, under each 2-D cue, on \n{cv.n} photographs per task. The cues are not balanced. The last rows give how often each cue is right and on how many items it is wrong. Where the lower box is the farther one, every frontier model loses accuracy, and the Qwen intervals end within half a point of zero. The size of the box carries no reliable gap, with no interval above zero, as for \family{closer} in the rendered scenes. Where the object nearer the reference in the picture is farther from it in the scene, every model loses accuracy on Distance too. Intervals are 95\% item-bootstrap intervals.}
\label{tab:cvbench}
\small
\resizebox{\textwidth}{!}{\begin{tabular}{lccccc}
\toprule
 & \multicolumn{3}{c}{Depth: gap [95\% CI] under each cue} & \multicolumn{2}{c}{Distance} \\
\cmidrule(lr){2-4}\cmidrule(lr){5-6}
model & acc. & larger box & lower box & acc. & smaller pixel distance \\
\midrule
Claude Opus 5.5 & 98.2 & +0 [$-$2, +3] & +10 [+4, +17] & 95.3 & +19 [+12, +27] \\
GPT-5.5 & 95.0 & +1 [$-$2, +5] & +18 [+10, +27] & 90.3 & +48 [+38, +57] \\
Gemini 3.6 Flash & 95.7 & +3 [$-$0.3, +7] & +22 [+13, +31] & 91.7 & +28 [+20, +37] \\
Qwen3-VL-4B & 95.8 & +3 [$-$0.5, +6] & +6 [+0.1, +13] & 88.7 & +34 [+25, +43] \\
Qwen3-VL-8B & 94.3 & $-$1 [$-$5, +3] & +7 [+0.4, +14] & 90.8 & +44 [+35, +53] \\
\midrule
cue right on & -- & 59.5\% & 85.3\% & -- & 80.2\% \\
cue-wrong items & -- & 243 & 88 & -- & 119 \\
\bottomrule
\end{tabular}
}
\end{table}

The frontier models lose accuracy on the CV-Bench photographs in which a 2-D cue misleads.

\paragraph{Data and cues.} The 3-D subset of CV-Bench \citep{tong2024cambrian} asks which of two marked objects is closer to the camera (Depth) and which of two objects is closer to a third (Distance), on \n{cv.n} photographs per task from indoor, street and synthetic indoor scenes. Its items are not balanced against a cue, so we name the cues a reader of the picture would use and score the \sgap{} against each of them. For Depth these are the larger box and the lower box, whose bottom edge is lower in the frame. For Distance it is the smaller pixel distance from each object's box centre to the centre of the reference box. The frontier models, Qwen3-VL-4B and Qwen3-VL-8B answer every item with one letter. We compare the correlation of the answers with each cue against the correlation of that cue with the truth \citep{brunswik1956perception}. A cue that the answers follow more closely than it tracks the truth is used beyond its validity. As a check on the Depth cue, a small rendered tier puts objects on pedestals, so that height in the frame is wrong about depth on \n{elev.closer.n} of its 32 \family{closer} items (the elevated tier of Appendix~\ref{app:constancy}).

Natural photographs rarely set the picture against the scene, so a photographic benchmark can test the gap only on the few items where a cue misleads. \citet{min2026whyfar} split CV-Bench Depth by whether the farther object sits higher in the frame and report the resulting gap for open models. We extend the test to the frontier models and to the Distance task. CV-Bench Depth has \n{cv.ndisc.bottom} items on which the box lower in the frame belongs to the farther object. These \n{cv.bottom.frac}\% of items hold \n{cv.bottom.share}\% of the frontier models' Depth errors, and the gap under this cue is \n{cv.bottom.gaprange} points with intervals that exclude zero (Table~\ref{tab:cvbench}). Their answers correlate with this cue at \n{r10.lens.Depth.util}, slightly above the cue's own correlation with the truth of \n{r10.lens.Depth.val}, so their errors lean toward the cue \citep{brunswik1956perception}.

The pedestal scenes test height in the frame directly. There the frontier models answer \n{elev.closer.right} of the \n{elev.closer.n} cue-wrong \family{closer} items correctly (Appendix~\ref{app:constancy}). Each of the two perfect scores has a 95\% interval of 79 to 100\%. The Depth errors may therefore come from photographs in which the common support of the two objects is hard to see.

On Distance the models lean on the cue more than it deserves. The object nearer the reference box in the picture is also nearer to it in the scene on \n{cv.cue.dist}\% of items. On the \n{cv.ndisc.dist} items where it is not, the frontier models show gaps of \n{cv.dist.gaprange} points, and these items hold \n{cv.dist.share}\% of their Distance errors. GPT-5.5's answers correlate with this cue at \n{r10.lens.Distance.Gpt.util}, above the cue's own correlation with the truth of \n{r10.lens.Distance.val}, and follow it on \n{r10.lens.Distance.Gpt.follow}\% of the items where it is wrong. Leaving out any one of the three source datasets keeps every frontier point estimate at or above \n{cvs.depth.loo.min} points on Depth and \n{cvs.dist.loo.min} on Distance (Appendix Table~\ref{tab:cvsrc}).

The larger box carries no reliable gap, just as pixel area carries none for \family{closer} in the renders. On Depth the answers are consistent with cue use. On Distance they use the cue more than its validity warrants.

\subsection{Does the information reach the language model? (RQ5)}
\label{sec:res-probe}

\begin{table}[!t]
\centering
\caption{Readers of the ground-plane layout for Qwen3-VL-4B and Qwen3-VL-8B. The hidden states give a layout close in $R^2$ to the reader of the 2-D measurements, with no compression of depth, as for any reader trained on the true positions, while the model's drawn map and answers do much worse. The reader of the 2-D measurements is the same ridge probe applied to exact pixel measurements. The next two rows give it the visible boxes only or use gradient-boosted trees instead. $R^2$ pools the \n{probe.nobj} objects of the three tiers and the second medium split. Maps, aligned as in Section~\ref{sec:res-maps}, and statements are scored on the 600 medium items. Intervals are 95\% scene-bootstrap intervals.}
\label{tab:probe}
\small
\resizebox{\textwidth}{!}{\begin{tabular}{lcccccc}
\toprule
reader & depth $R^2$ [95\% CI] & lateral $R^2$ & depth-to-width ratio & acc. & gap [95\% CI] & \family{fits} gap \\
\midrule
Hidden states, layer 16 & 0.93 [0.92, 0.93] & 0.94 & 1.01 & 80.2 & +8 [+2, +14] & +27 \\
Reader of the 2-D measurements (ridge) & 0.87 [0.86, 0.88] & 0.96 & 1.00 & 87.8 & +6 [+1, +11] & +28 \\
Visible boxes only, ridge & 0.86 [0.84, 0.87] & 0.96 & 0.97 & 79.5 & +20 [+13, +26] & +37 \\
Gradient-boosted trees on 2-D measurements & 0.95 [0.94, 0.95] & 0.98 & -- & -- & -- & -- \\
\midrule
Qwen3-VL-4B's drawn map & -- & -- & 0.25 & 31.5 & +16 & +15 \\
Qwen3-VL-4B's answers from the image & -- & -- & -- & 61.8 & +14 & +2 \\
\midrule
Qwen3-VL-8B hidden states, layer 28 & 0.89 [0.88, 0.90] & 0.91 & 0.99 & 70.8 & +13 [+6, +20] & +23 \\
Qwen3-VL-8B's drawn map & -- & -- & 0.31 & 38.1 & +23 & +25 \\
Qwen3-VL-8B's answers from the image & -- & -- & -- & 63.3 & +14 & $-$10 \\
\bottomrule
\end{tabular}
}
\end{table}

In the two probed models the information that the picture carries reaches the language model, while their drawn maps lose much of it.

\paragraph{Models and probes.} The drawn maps leave open whether the information that the picture carries is lost before it reaches the language model or afterwards. The frontier models are closed, so we follow the information inside Qwen3-VL-4B-Instruct and Qwen3-VL-8B-Instruct, the two probed models, run in bfloat16. A hidden state is the vector of numbers that the language model computes for one token at one layer. Image tokens are the tokens into which the vision encoder turns the picture, one for each $32\times32$-pixel cell. A layer is one of the language model's stacked processing steps. Layer 0 is the output of the vision merger, which the language model receives as input. The merger is the small network that joins neighbouring vision-encoder features into image tokens.

A probe is a linear regression trained to read a known quantity from hidden states. If so simple a reader can recover a quantity, the hidden states make it available, whether or not the model uses it. Our probes are ridge regressions, cross-validated over scenes, that read each visible object's ground depth and lateral position from the mean hidden state of the image tokens that cover its visible mask. The masks come from the generator, so the probes receive a segmentation that the model does not. Qwen3-VL also adds features from three intermediate vision layers to the first three layers of the language model (DeepStack, \citealp{bai2025qwen3vl}), so the language model receives visual information at more than one layer.

The probes cover \n{probe.nobj} objects in \n{probe.nscene} scenes. The same probe applied to each object's exact pixel measurements is the reference, the reader of the 2-D measurements. A probe on the hidden states can read depth better than the reader of the 2-D measurements, because the hidden states encode depth in a form closer to linear than the raw measurements do (Appendix~\ref{app:probe}). Further probes read object positions at the object's name in the statement and the truth of the statement at the answer position. The answer position is the last token of the prompt, whose next-token distribution gives the answer. The two probed models are scored by the log-probabilities of True and False.

\paragraph{Edits.} An edit adds a chosen vector to a hidden state. We then check whether the answer moves. To test whether the answers use the layout we fit encoding directions for ground depth and picture position \citep{haufe2014interpretation}. An encoding direction is the least-squares map from a quantity to the hidden state, the direction along which the hidden states move as the quantity grows. A \family{closer} edit swaps the two objects' values along one of these directions. A \family{between} edit moves the middle object to the midpoint of the other two, on the ground or in the picture. Each edit is made at the size that the swap or move implies and at four times that size, in the image tokens that the objects cover. We read the change in the log-odds of True over False against random edits of the same norm as the combined depth-and-picture edit. Appendix~\ref{app:openfull} gives the full methods and results.

Qwen3-VL-4B answers \n{probe.beh.acc}\% of medium items correctly with a gap of \n{probe.beh.gap} points, and the maps it draws keep \n{probe.own.ratio} of the true depth-to-width ratio (Table~\ref{tab:probe}). A probe of its hidden states recovers far more than its drawn map shows. At layer \n{probe.layer} the probe recovers each object's relative depth with $R^2$ \n{probe.r2depth} \n{probe.r2depth.ci}, against \n{probe.pix.r2depth} \n{probe.pix.r2depth.ci} for the reader of the 2-D measurements. A map decoded from the states is not compressed, as for any reader trained on the true positions. Depth is decodable with $R^2$ \n{probe.r2depth.l0} already at layer 0, the output of the vision merger.

The probe therefore shows that the information in the picture reaches the language model, not that the model forms a better layout than the picture provides. The comparison with the drawn maps is between a probe trained on the true layout and a zero-shot answer. Statements re-derived from the 4B's maps are right on \n{probe.own.acc}\%, below chance, in part because many of its maps put objects at nearly the same depth, inside the reject band of \family{closer}. The compression thus partly reflects a failure at the map task itself. Qwen3-VL-8B, a robustness check within the same family, behaves the same way, with a depth $R^2$ of \n{p8.r2depth} and drawn maps that keep \n{p8.own.ratio} of the ratio. Its lower value at layer 0 ($R^2$ \n{p8.r2depth.l0}) fits the loss of spatial information at the merger that \citet{maheshwari2026catalogue} report for a model of the same size.

The information also reaches the words of the statement and the answer position. At the last token of each object's name a probe reads its depth with $R^2$ \n{ro.ment.r2}. On the cue-wrong statements outside \family{closer}, a probe at the answer position recovers the truth on \n{ro.nc.probe}\% of items against \n{ro.nc.model}\% for the model's answers (difference \n{ro.nc.diff} \n{ro.nc.diff.ci} points). An input-only reader of the 2-D measurements (gradient-boosted trees), which never sees the model, does as well (\n{rc.in.nc}\%, Appendix~\ref{app:openfull-probe}).

Editing the hidden states shows that the answers depend weakly on the image tokens. Swapping the ground depths of the two objects of a \family{closer} statement moves the answers both ways at layers 8 and 12, as in the depth write-back of \citet{zou2026seeingdepth}. The log-odds change by \n{st.dep.rise} on the statements the swap makes true and by \n{st.dep.fall} on those it makes false. At layer 12 the swap at four times the size of the change flips \n{st.flip.dep.l12}\% of the answers, against \n{st.flip.rpc.l12}\% for a random edit (differences of \n{r10.st.4b.l12.fd} \n{r10.st.4b.l12.fd.ci} points in the 4B and \n{r10.st.8b.l12.fd} \n{r10.st.8b.l12.fd.ci} in the 8B). Its effect is gone by layer 20 (Appendix Tables~\ref{tab:steer} and \ref{tab:steergrid}). For \family{between}, picture edits move the answers further than ground edits (rises of \n{st.bp.rise} against \n{st.bg.rise}), although the picture edits are also larger in norm (Appendix Table~\ref{tab:steergrid}). Ground edits flip answers no more often than random edits, so these edits give no evidence that \family{between} is read on the ground. The edits cannot separate ground depth from picture position, because an object's depth and its row in the picture are strongly correlated in these scenes (Appendix~\ref{app:openfull}).

\begin{table}[!t]
\centering
\caption{Behaviour of the two probed models on the 600 medium items, scored by the log-probabilities of True and False. The image condition gives accuracy (\%), the pooled gap and the gap per relation. The coordinate inputs, which have no image, reuse the frontier models' coordinate prompts verbatim. Ground coordinates do not help either model, so by the coordinates rule of Section~\ref{sec:protocol} their gaps are not interpreted.}
\label{tab:qwenfam}
\small
\resizebox{\textwidth}{!}{\begin{tabular}{lcccccccccc}
\toprule
 & \multicolumn{7}{c}{image} & \multicolumn{2}{c}{coordinates only} & coordinates + numeric def. \\
\cmidrule(lr){2-8}\cmidrule(lr){9-10}\cmidrule(lr){11-11}
model & acc. & gap & \family{closer} & \family{taller} & \family{touching} & \family{between} & \family{fits} & acc. & gap & acc. / gap \\
\midrule
Qwen3-VL-4B & 61.8 & +13.7 & +3 & +33 & +18 & +12 & +2 & 53.0 & +3.3 & 53.3 / +0.7 \\
Qwen3-VL-8B & 63.3 & +14.0 & +10 & +22 & +30 & +18 & $-$10 & 55.8 & +3.7 & 54.3 / +5.3 \\
\bottomrule
\end{tabular}
}
\end{table}

The two probed models fail the coordinates rule of Section~\ref{sec:protocol}. Given the frontier models' coordinate prompts without an image, they answer \n{r10.q4b.co.acc}\% and \n{r10.q8b.co.acc}\% of items correctly, and \n{r10.q4b.cond.acc}\% and \n{r10.q8b.cond.acc}\% with the numeric definitions, below their accuracy on the image condition (Table~\ref{tab:qwenfam}). Their largest gaps on the image condition are on \family{taller} and \family{touching} (Table~\ref{tab:qwenfam}).

\FloatBarrier
\section{Discussion}
\label{sec:analysis}

\paragraph{What the evidence locates.} Each manipulation removes one candidate explanation. Balancing the cue rules out a gap that comes from the label distribution, and the difficulty adjustment rules out the measured difficulty covariates as the source of the \family{fits} gap on the medium and hard tiers. The ground-plane form rules out an inability to apply the relations. The perspective form splits the remaining step into reading measurements and depths from the picture and carrying them onto the ground. Getting from the picture to measurements and depths accounts for most of the \family{fits} gap, since text inputs that supply them close most of it. The \family{between} errors persist with exact measurements, where the models apply the rule to image-plane angles and read the word loosely. From the picture, part of the \family{between} gap may arise earlier, in where the models anchor objects (see the paragraph on 2-D cues below). The maps show the layout that this step produces, with depth compressed toward the picture by an amount that differs between models. The maps do not predict the answers item by item (Section~\ref{sec:res-maps}), so the compression they show may accompany the failure without causing it.

For \family{fits} the pattern is consistent with comparing a gap and an object without rescaling them to a common depth, the problem of size constancy \citep{holway1941size,sun2025constancy}. This holds for GPT-5.5 and Gemini in the cell that tests size scaling. The errors of Opus 5.5 lie mainly in the foreshortened cell, where the object fits but looks too wide and the size-constancy reader also fails. In the two probed models the probes show the information reaching the language model and the edits show the answers reading the image tokens weakly.

\paragraph{Which 2-D cues?} Each relation's cue is the simplest 2-D reading, which is not always the cue that human picture perception relies on \citep{cutting1995perceiving,sedgwick1986space}. A reader that uses only each object's lowest visible point solves \n{r10.btwvis.disc}\% of the cue-wrong \family{between} items on the easy, medium and hard tiers, yet the frontier \family{between} gaps on the easy tier are \n{r10.btw.easygap} points. The \family{between} gap may therefore reflect where the models anchor objects in the picture as well as whether they carry the picture onto the ground (Appendix~\ref{app:cues}). Benchmarks that balance a cue should balance the one a competent observer of pictures would use, such as ground contact or height below the horizon, or they risk scoring good picture reading as 3-D understanding.

\FloatBarrier
\section{Limitations}
\label{sec:limitations}

\paragraph{Scope.} For the frontier models the results describe behaviour under different inputs, not the internal computation. A sufficient input locates the failing step in behaviour. A small gap on \bench{} shows that a model's answers do not follow the picture, which is one part of 3-D understanding. The headline numbers are at low reasoning effort. High effort changes the \family{fits} gap by \n{r10.eff.fits.drange} points (Section~\ref{sec:res-image}).

\paragraph{Scenes.} The scenes are rendered primitives on a plane, which is what allows exact balancing. The check on photographs covers depth order and relative distance, on the \n{cv.ndisc.bottom} and \n{cv.ndisc.dist} CV-Bench items where the lower-box and distance cues are wrong. These items are not adjusted for difficulty, so the photographic gaps are an upper bound on the cue's influence. Testing \family{taller}, \family{touching}, \family{between} and \family{fits} on photographs needs metric annotations and items balanced against the cue.

\paragraph{Human reference.} We collected no human judgements of these pictures. The human constancy value we quote comes from real scenes. Appendix~\ref{app:human} gives a protocol for a study with the same pictures.

\paragraph{\family{between} and \family{touching}.} In the false half of \family{between} the cue-wrong items lie near the decision threshold. In every cue-wrong contact of \family{touching} a third object hides the contact point, and we do not show that the picture alone determines these answers. Their gaps may mix word use and occlusion with geometry, which is why we also report the gap without these two relations.

\paragraph{Probed models.} The two probed models are small open models of one family. Neither passes the coordinates rule of Section~\ref{sec:protocol}, so we use them to follow where the picture's information goes and do not interpret their gaps. The probes and the edits are linear. They show what a simple reader can use, not the computation the model performs.

\FloatBarrier
\section{Conclusion}
\label{sec:conclusion}

The failure lies in the step from the picture to the layout. GPT-5.5 recovers a layout close to the picture plane. Gemini 3.6 Flash's layout corrects part of the way and adds a uniform compression of depth. Opus 5.5's maps have a constancy index close to the one reported for human observers of real scenes, although its answers still show a picture-plane gap on \family{fits}. All three read the numbers in inputs that carry the information and can apply the relations once the layout is given. The layouts they draw from the picture keep most of the order of the objects in depth but compress the distances. In the two probed models the hidden states keep the information that the picture carries, while their drawn maps lose much of it.

Accuracy on natural images cannot show this, because there the picture and the scene agree. On the few CV-Bench images where they disagree, the same models lose accuracy. Spatial evaluations can follow \bench{} and balance a 2-D cue against the truth. We suggest the cue that a competent observer of pictures would use, with the resulting gap reported next to accuracy and the reasoning effort. Work on 3-D reasoning in VLMs should target this step. For agents that must judge clearance from a single frame, supplying exact ground coordinates from depth sensing or a map removes most of the failure we measure.

\bibliographystyle{plainnat}
\bibliography{refs}

\appendix

\FloatBarrier
\section{Further analyses}
\label{app:further}

This appendix keeps the full per-model results that the main text summarises as ranges.

\paragraph{Main results.} The frontier models' answers track the picture plane. Opus 5.5, GPT-5.5 and Gemini 3.6 Flash reach \n{main.accrange}\% accuracy with gaps of \n{main.gaprange} points (Table~\ref{tab:main}). On medium, GPT-5.5 is \n{Gpt.medium.image.conc}\% correct on cue-right items and \n{Gpt.medium.image.disc}\% on cue-wrong items. The second medium split, from an independent seed (seed 11), gives gaps of \n{Opus.s11.image.gap}, \n{Gpt.s11.image.gap} and \n{Gemini.s11.image.gap} (Appendix Table~\ref{tab:seed}). Three repeats move the pooled gap by about a point (Appendix Table~\ref{tab:repeats}), and tier changes it little (Table~\ref{tab:main}). Grok 4.7 patterns with the frontier models (\n{Grok.medium.image.acc}\%, \n{Grok.medium.image.gap}) and Qwen3-VL-235B sits between (\n{Qwen.medium.image.acc}\%, \n{Qwen.medium.image.gap}). Nova 2 Pro (60--64\%) and three open-weight models (50.3--60.5\%) score lower and gain little from ground coordinates (Pixtral \n{Pixtral.medium.coordsoracle.acc}\%, Llama \n{Llama.medium.coordsoracle.acc}\%, Gemma \n{Gemma.medium.coordsoracle.acc}\%), so their errors are not specific to the picture plane. By the coordinates rule of Section~\ref{sec:protocol} we do not read their gaps or those of Qwen3-VL-235B. Text-only accuracy is \n{all.textonly.accrange}\% for every model, so the statement wording alone does not give away the label.

\paragraph{Where the gap sits.} The gap is concentrated in the relations that compare quantities across depth or behind occluders (Figure~\ref{fig:heatmap}). Per-relation 95\% CIs are about $\pm15$ points. \family{fits} is \n{main.fits.gaprange}, and GPT-5.5 answers its cue-wrong half at \n{Gpt.medium.image.fits.disc}\%, below chance, as on the item of Figure~\ref{fig:running}. \family{between} is \n{main.between.gaprange} and \family{touching} \n{main.touching.gaprange}. \family{taller} is small for Opus 5.5 and moderate for the other two. The control relation \family{closer} is \n{main.closer.gaprange}.

\paragraph{Harder items or the cue?} If cue-wrong items were only harder, the gap should vanish once difficulty is held fixed. For \family{fits} it does not. The pooled adjusted gaps are \n{main.adjgap.range} points with intervals excluding zero for every frontier model and tier (Appendix Table~\ref{tab:matched}). Appendix Table~\ref{tab:familyadj} gives the adjusted gap of each relation. On the \family{fits} items that share covariate support, the gap is \n{ov.fits.suprange} raw and \n{ov.fits.ipwrange} after inverse-propensity weighting (Appendix Table~\ref{tab:overlap}). Within tertiles of the margin (the distance from the decision threshold), the pooled gap is \n{main.T1.gaprange} on the smallest margins and \n{main.T3.gaprange} on the largest. This grading is consistent with a cue-combination observer \citep{landy1995cue}. \family{between} cannot be settled this way. Its adjusted gap is \n{Opus.r5.btw.shared}, \n{Gpt.r5.btw.shared} and \n{Gemini.r5.btw.shared} under the shared-slope model, \n{Opus.r5.btw.ipw}, \n{Gpt.r5.btw.ipw} and \n{Gemini.r5.btw.ipw} with inverse-propensity weighting, and \n{Opus.r5.btw.trim}, \n{Gpt.r5.btw.trim} and \n{Gemini.r5.btw.trim} in the within-relation model with a binary construct flag (Appendix Table~\ref{tab:overlap}). In its false half, cue status and nearness to the reject band coincide almost exactly, so no regression can separate the cue from the margin. We report \family{between} as confounded and take \family{fits} as the case free of the margin confound.

\paragraph{Information or form?} Ground-plane form removes the gap. Perspective form that carries nearly the same information removes only part of it (Tables~\ref{tab:solvers} and \ref{tab:ladder}). Exact pixel measurements as text leave accuracy close to the image condition (\n{Opus.medium.pixnoimage.acc}, \n{Gpt.medium.pixnoimage.acc} and \n{Gemini.medium.pixnoimage.acc}\% without the image), so the models read the text. Adding camera depth makes \family{closer} and \family{between} exactly solvable and \family{fits} solvable to \n{solver.pixdepthoracle.fits}\% by the solver. The models then cut the \family{fits} gap from \n{Opus.medium.image.fits.gap}, \n{Gpt.medium.image.fits.gap} and \n{Gemini.medium.image.fits.gap} to \n{Opus.medium.pixdepthoracle.fits.gap}, \n{Gpt.medium.pixdepthoracle.fits.gap} and \n{Gemini.medium.pixdepthoracle.fits.gap}, while \family{between} stays at \n{Opus.medium.pixdepthoracle.between.gap}, \n{Gpt.medium.pixdepthoracle.between.gap} and \n{Gemini.medium.pixdepthoracle.between.gap}. A depth map as a second image changes the pooled gap by \n{main.depthdelta.range}. The top-down map, a second image with labelled footprints, narrows the pooled gap to \n{main.topdownoracle.gaprange}, as coordinates do without the image (\n{main.coordsnoimage.gaprange}, and \n{Opus.s11.coordsnoimage.gap}, \n{Gpt.s11.coordsnoimage.gap} and \n{Gemini.s11.coordsnoimage.gap} on the second medium split). With coordinates and the numeric definitions, accuracy is \n{main.coordsnoimagendef.accrange}\% and no gap is more than \n{main.coordsnoimagendef.absgapmax} points from zero. Table~\ref{tab:ladder} gives the gap under each input with its interval, and Table~\ref{tab:perspgaps} the \family{fits} and \family{between} gaps. The models reason correctly on the ground plane and convert a perspective description into it only partly.

\paragraph{The \family{between} rule.} Given an exact rule, the models apply it to whatever points they receive. Given pixel points, they apply it to image-plane angles. In Table~\ref{tab:betweenladder} we fix the numeric definition and change only the position input. Visible-region centres give \n{Opus.bl.cent.acc}, \n{Gpt.bl.cent.acc} and \n{Gemini.bl.cent.acc}\%, ground-contact pixels \n{Opus.bl.base.acc}, \n{Gpt.bl.base.acc} and \n{Gemini.bl.base.acc}\%, and the homography or coordinates up to 100\%. Applying the $10^\circ$ rule to the image-plane angles of the same points scores \n{r5.rule10.cent.acc} and \n{r5.rule10.base.acc}\%. GPT-5.5's answers agree with that image-plane rule on \n{Gpt.r5.rule10.base.agree}\% of items. Image-plane angles decide most items only because the reject band leaves nothing between $10^\circ$ and $30^\circ$. Without the numeric rule, ground coordinates still leave \family{between} gaps of \n{Opus.medium.coordsnoimage.between.gap}, \n{Gpt.medium.coordsnoimage.between.gap} and \n{Gemini.medium.coordsnoimage.between.gap}, so the models' own ``between'' is looser than ours. The \family{between} gap combines a rule looser than ours with the composition of its cue-wrong cell near the reject band (Appendix Table~\ref{tab:betweenbands}).

\begin{table}[t]
\centering\small
\caption{Accuracy (\%) / gap (points) [95\% CI] of \family{between} under one rule (numeric definitions) and five position inputs (the image and four text inputs), medium tier (120 items). Text inputs have no image, and the solver column reads the same text. The models match the solver on each input. $^\dagger$ marks an approximate solver (Table~\ref{tab:solvers}).}
\label{tab:betweenladder}
\resizebox{\textwidth}{!}{\begin{tabular}{llllr}
\toprule
input (all with numeric definitions) & Claude Opus 5.5 & GPT-5.5 & Gemini 3.6 Flash & solver \\
\midrule
image & 87.5 / +18.3 [+7, +30] & 78.3 / +36.7 [+23, +50] & 82.5 / +28.3 [+16, +41] & -- \\
visible-region centres (text) & 61.7 / +50.0 [+35, +65] & 61.7 / +53.3 [+39, +68] & 60.0 / +53.3 [+39, +68] & 72$^\dagger$ \\
ground-contact pixels (text) & 94.2 / +5.0 [$-$3, +14] & 93.3 / +6.7 [$-$2, +16] & 92.5 / +5.0 [$-$4, +15] & 94$^\dagger$ \\
ground-contact pixels + homography (text) & 99.2 / +1.7 [+0, +6] & 100.0 / +0.0 [+0, +0] & 91.7 / $-$3.3 [$-$14, +6] & 100 \\
ground coordinates (text) & 100.0 / +0.0 [+0, +0] & 100.0 / +0.0 [+0, +0] & 100.0 / +0.0 [+0, +0] & 100 \\
\bottomrule
\end{tabular}
}
\end{table}

\paragraph{Contacts.} The two best models' \family{touching} gap sits on contacts hidden behind a third object (Table~\ref{tab:touchoracles}). Opus 5.5 answers apart pairs perfectly, whether or not they overlap in projection. It answers hidden contacts at \n{Opus.to.image.hidden}\%. With coordinates and no rule, hidden contacts are \n{Opus.to.coordsnoimage.hidden}, \n{Gpt.to.coordsnoimage.hidden} and \n{Gemini.to.coordsnoimage.hidden}\% correct. With the numeric contact rule the same coordinates give \n{Opus.to.coordsnoimagendef.hidden}, \n{Gpt.to.coordsnoimagendef.hidden} and \n{Gemini.to.coordsnoimagendef.hidden}\%. Exact surface distances give 100\%. On the ground plane, contact is a computation the models perform once the rule is stated. From the picture, the contact point is hidden in \n{cv.hidden} of \n{cv.ncontact} true contacts, a split that is close to forced for two convex solids that meet. A third object covers the region between the silhouettes in every cue-wrong contact, so whether these contacts can be resolved from the picture at all is left open (Section~\ref{sec:limitations}).

\begin{table}[t]
\centering\small
\caption{\family{touching} on the medium tier, accuracy (\%) by proposition truth (the truth of the relation before any negation) and cue status, and pooled accuracy / gap. Hidden contacts are in contact, with the visible masks more than 3\,px apart because a third object covers the contact. Overlapping pairs are apart, with masks that overlap in projection. Each cell has 30 items. Coordinates with the numeric rule solve every cell.}
\label{tab:touchoracles}
\resizebox{\textwidth}{!}{\begin{tabular}{ll rrrr r}
\toprule
 & & \multicolumn{2}{c}{in contact} & \multicolumn{2}{c}{apart} & \\
\cmidrule(lr){3-4}\cmidrule(lr){5-6}
model & input & masks within 3 px & hidden & masks apart & overlapping & acc / gap \\
\midrule
Claude Opus 5.5 & image & 63 & 10 & 100 & 100 & 68 / +27 \\
Claude Opus 5.5 & image + numeric definitions & 70 & 20 & 100 & 97 & 72 / +27 \\
Claude Opus 5.5 & coordinates only & 90 & 83 & 100 & 100 & 93 / +3 \\
Claude Opus 5.5 & coordinates only + verbal definitions & 90 & 77 & 100 & 100 & 92 / +7 \\
Claude Opus 5.5 & coordinates only + numeric definitions & 100 & 100 & 100 & 97 & 99 / +2 \\
Claude Opus 5.5 & image + surface distances + numeric definitions & 100 & 100 & 100 & 100 & 100 / +0 \\
\addlinespace
GPT-5.5 & image & 80 & 23 & 97 & 53 & 63 / +50 \\
GPT-5.5 & image + numeric definitions & 57 & 27 & 97 & 50 & 57 / +38 \\
GPT-5.5 & coordinates only & 63 & 47 & 100 & 100 & 78 / +8 \\
GPT-5.5 & coordinates only + verbal definitions & 63 & 50 & 100 & 100 & 78 / +7 \\
GPT-5.5 & coordinates only + numeric definitions & 100 & 100 & 100 & 100 & 100 / +0 \\
GPT-5.5 & image + surface distances + numeric definitions & 100 & 100 & 100 & 100 & 100 / +0 \\
\addlinespace
Gemini 3.6 Flash & image & 57 & 33 & 100 & 87 & 69 / +18 \\
Gemini 3.6 Flash & image + numeric definitions & 67 & 37 & 100 & 93 & 74 / +18 \\
Gemini 3.6 Flash & coordinates only & 67 & 47 & 100 & 100 & 78 / +10 \\
Gemini 3.6 Flash & coordinates only + verbal definitions & 60 & 47 & 100 & 100 & 77 / +7 \\
Gemini 3.6 Flash & coordinates only + numeric definitions & 93 & 97 & 100 & 100 & 98 / $-$2 \\
Gemini 3.6 Flash & image + surface distances + numeric definitions & 100 & 100 & 100 & 100 & 100 / +0 \\
\bottomrule
\end{tabular}
}
\end{table}

\paragraph{Definitions and bands.} The stated-threshold prompt mixes each relation's numeric rule with the reject bands, which tell the model which near-miss cases are absent. We separate the two in Table~\ref{tab:factorial}. For Opus 5.5, numeric definitions change the gap by \n{Opus.fac.ndef.dgap} \n{Opus.fac.ndef.dgapci}, bands by \n{Opus.fac.bands.dgap} \n{Opus.fac.bands.dgapci} and both by \n{Opus.fac.ndefbands.dgap} \n{Opus.fac.ndefbands.dgapci}. The two do not add. Their interaction is \n{Opus.r5.fac.inter} \n{Opus.r5.fac.interci} for Opus 5.5 and \n{Gemini.r5.fac.inter} \n{Gemini.r5.fac.interci} for Gemini. The bands carry sampling information rather than geometry. With bands alone, Opus 5.5 answers hidden contacts at \n{Opus.r5.hidden.bands}\% against \n{Opus.r5.hidden.image}\% with the default prompt, because the bands imply that pairs that look close are in contact. For GPT-5.5 every paired interval includes zero. Headline numbers use the default prompt.

\begin{table}[t]
\centering\small
\caption{Prompt factorial on the medium image condition, with the stated-threshold prompt split into numeric definitions and reject bands. $\Delta$gap is the paired change against the default prompt (scene bootstrap), and the last three columns are per-relation gaps. No prefix closes the gap.}
\label{tab:factorial}
\resizebox{\textwidth}{!}{\begin{tabular}{ll l l rrr}
\toprule
model & prompt prefix & acc / gap [95\% CI] & $\Delta$gap vs.\ default [CI] & touching & between & fits \\
\midrule
Claude Opus 5.5 & neither (default prompt) & 85.7 / +21.3 [+16, +27] & -- & +27 & +45 & +32 \\
Claude Opus 5.5 & numeric definitions only & 89.0 / +14.0 [+9, +19] & $-$7.3 [$-$13, $-$2] & +27 & +18 & +23 \\
Claude Opus 5.5 & reject bands only & 92.3 / +12.0 [+8, +16] & $-$9.3 [$-$14, $-$4] & +7 & +27 & +22 \\
Claude Opus 5.5 & numeric definitions + bands & 94.2 / +9.0 [+5, +13] & $-$12.3 [$-$18, $-$7] & +17 & +8 & +18 \\
Claude Opus 5.5 & stated thresholds (original wording) & 93.2 / +11.7 [+8, +16] & $-$9.7 [$-$14, $-$5] & +15 & +15 & +20 \\
Claude Opus 5.5 & verbal definitions & 88.0 / +14.7 [+10, +19] & $-$6.7 [$-$10, $-$3] & +20 & +28 & +20 \\
\addlinespace
GPT-5.5 & neither (default prompt) & 78.5 / +35.0 [+29, +41] & -- & +50 & +43 & +65 \\
GPT-5.5 & numeric definitions only & 78.7 / +32.0 [+25, +38] & $-$3.0 [$-$9, +3] & +38 & +37 & +68 \\
GPT-5.5 & reject bands only & 80.5 / +34.3 [+28, +40] & $-$0.7 [$-$5, +4] & +53 & +43 & +63 \\
GPT-5.5 & numeric definitions + bands & 77.7 / +38.0 [+32, +44] & +3.0 [$-$2, +8] & +53 & +43 & +75 \\
GPT-5.5 & stated thresholds (original wording) & 80.8 / +33.0 [+27, +39] & $-$2.0 [$-$7, +3] & +53 & +38 & +58 \\
GPT-5.5 & verbal definitions & 77.0 / +33.3 [+27, +39] & $-$1.7 [$-$7, +4] & +42 & +38 & +67 \\
\addlinespace
Gemini 3.6 Flash & neither (default prompt) & 81.7 / +24.7 [+19, +30] & -- & +18 & +43 & +45 \\
Gemini 3.6 Flash & numeric definitions only & 85.2 / +19.7 [+15, +25] & $-$5.0 [$-$10, $-$0] & +18 & +28 & +47 \\
Gemini 3.6 Flash & reject bands only & 84.5 / +21.7 [+16, +27] & $-$3.0 [$-$8, +2] & +23 & +42 & +33 \\
Gemini 3.6 Flash & numeric definitions + bands & 83.8 / +23.7 [+18, +30] & $-$1.0 [$-$6, +4] & +25 & +38 & +42 \\
Gemini 3.6 Flash & stated thresholds (original wording) & 84.0 / +22.0 [+17, +27] & $-$2.7 [$-$7, +2] & +22 & +35 & +40 \\
Gemini 3.6 Flash & verbal definitions & 81.5 / +24.3 [+19, +30] & $-$0.3 [$-$5, +5] & +17 & +38 & +52 \\
\bottomrule
\end{tabular}
}
\end{table}

\paragraph{Effort, tools and views.} More reasoning and exact measurement narrow the pooled gap by a few points and leave it in place, although high effort changes the \family{fits} gap by \n{r10.eff.fits.drange} points (Section~\ref{sec:res-image}). High effort with a 16{,}384-token budget changes the gap by \n{main.high16k.deltarange} points (Appendix Table~\ref{tab:effort}). A zoom tool and a sandboxed Python interpreter over the pixel array change Opus 5.5's gap by \n{Opus.agent.agentic.dgap} \n{Opus.agent.agentic.dgapci}, with tools used on \n{Opus.agent.agentic.compliance}\% of items (Appendix Table~\ref{tab:agentic}). The measurements are accurate. Asked only to measure, Opus 5.5 returns visible-pixel boxes with a median IoU of \n{measure.iou}. Its lowest row is within 2\,px on \n{measure.bottom2}\% of \n{measure.nobj} objects. Accurate measurement in the picture plane does not move the answer onto the ground plane. A checkerboard ground, a dense perspective texture, leaves the gap unchanged within noise (Appendix Table~\ref{tab:checker}). On a second camera, scored against that camera's own cue, the gaps are \n{Opus.tv.v2own.gap} \n{Opus.tv.v2own.gapci}, \n{Gpt.tv.v2own.gap} \n{Gpt.tv.v2own.gapci} and \n{Gemini.tv.v2own.gap} \n{Gemini.tv.v2own.gapci}, on only \n{tv.v2disc} cue-wrong items (Appendix Table~\ref{tab:twoviewgap}). GPT-5.5's gap follows the cue to the new camera. Agreement between the two cameras is in Appendix Table~\ref{tab:twoview}.

\paragraph{The models' drawn maps.} Drawn from above, the frontier models' maps sit at or between the picture-plane copy and the scene (Table~\ref{tab:map}). The picture-plane copy compresses depth to \n{r5.picturecopy.medium.ratio} of the lateral spread on medium and \n{r5.picturecopy.hard.ratio} on the hard tier. GPT-5.5's and Gemini's maps are compressed to \n{Gpt.map.medium.ratio} and \n{Gemini.map.medium.ratio} on medium, about as much as the copy, while Opus 5.5's reach \n{Opus.map.medium.ratio}. Depth order is mostly right in all three (median rank correlation along the camera axis \n{Opus.map.medium.rankdepth}, \n{Gpt.map.medium.rankdepth} and \n{Gemini.map.medium.rankdepth}). This is the pattern of phenomenal regression, a percept that moves only part of the way from the retinal image toward the real object \citep{thouless1931phenomenal}. It resembles the exocentric compression of human observers \citep{loomis1992visual}. Statements re-derived from the aligned maps show gaps of \n{map.medium.gaprange}. They match the direct answers on \n{mi.agreerange}\% of items (Mantel--Haenszel odds ratio \n{mi.mhorrange} within relation $\times$ cue strata, Appendix Table~\ref{tab:mapitems}). Fed their drawn maps as text, the models read them exactly as a solver does (\n{Opus.om.agree}\%). Stretching the depth axis by a factor estimated on the other half of the scenes changes the \family{fits} gap by \n{Opus.r5.om.fits.dgap} \n{Opus.r5.om.fits.dgapci}, \n{Gpt.r5.om.fits.dgap} \n{Gpt.r5.om.fits.dgapci} and \n{Gemini.r5.om.fits.dgap} \n{Gemini.r5.om.fits.dgapci}, and the \family{between} gap by \n{Opus.r5.om.between.dgap}, \n{Gpt.r5.om.between.dgap} and \n{Gemini.r5.om.between.dgap} (Appendix Table~\ref{tab:ownmap}). Uniform compression explains part of \family{fits} and none of \family{between}. Figure~\ref{fig:ownmap} shows one \family{fits} item on which every map compresses the gap below the object's width. Grok 4.7 behaves like the frontier models but returns parseable maps for 224 of 304 scenes. The five other models' maps order depth less reliably, and the Qwen3-VL-235B maps run depth the wrong way (median rank correlation \n{Qwen.map.medium.rankdepth} on medium).

\FloatBarrier
\section{Generator and statistics details}
\label{app:details}

\paragraph{Scenes.} Each scene has four to eight convex primitives (cube, sphere, cylinder, cone) resting on a matte grey ground plane, lit by one sun lamp with a random azimuth and 30--60$^\circ$ elevation, and viewed by a perspective camera with a 35--50\,mm equivalent lens at 8--14 units from a jittered target near the scene centroid. Every object has a distinct colour from a palette of eight, so ``the red cube'' names exactly one object. Sizes are drawn log-uniformly in $[0.3, 2.0]$ world units independently of shape, so familiar-size priors are useless. Perspective, ground contact, occlusion and shadows are the only depth and scale cues. Objects rest on the ground, do not interpenetrate, and lie inside the frame with at least 0.15\% of the pixels and 35\% of their silhouette visible. Images are rendered at $768\times768$ with Blender 5.0.1 Cycles \citep{blender}. Depth maps and instance masks come from an analytic ray caster over the same primitives, so the 2-D statistics that define the cues are exact.

\paragraph{Relations.} Table~\ref{tab:families} defines the five relations. Each has a 3-D truth read from the scene graph, a 2-D cue computed from the visible instance masks, and a reject band around the decision threshold inside which candidates are discarded, so that the truth is unambiguous in the geometry. Surface distances are exact minimum distances between convex solids. In \family{fits}, the gap between the two objects is the minimum distance between their footprints, compared with the moving object's narrowest width (cube side or diameter), and the gap must be clear of any third object by 0.25 units. The item of Figure~\ref{fig:running} has a ratio of 1.64 on the ground, computed from the unrounded sizes, and a pixel ratio of 0.45. Half of all statements are negated and every template has three paraphrases, so no wording predicts the label. The default prompt does not state the numeric thresholds.

\paragraph{The control relation.} The natural cue for \family{closer}, ``lower in the frame means closer'', is close to veridical for objects on a ground plane seen from above. It agrees with the truth on all 480 \family{closer} items of the benchmark and on \n{cue.medium.vertical}\% of candidate pairs under random placement (Appendix Table~\ref{tab:cue}). This is an empirical property of our scene distribution and the 15\% depth margin, not a general theorem. We balance \family{closer} on pixel area instead, the size-to-depth confound that \citet{zou2026seeingdepth} found models follow, and treat it as a control whose zero gap says only that pixel area is not used.

\paragraph{Balancing.} For each relation we sample scenes and enumerate candidate statements until each of the four cells $\{\text{true},\text{false}\}\times\{\text{cue agrees},\text{disagrees}\}$ holds exactly $N/4$ items, half negated, with at most two statements per scene and never two about the same objects. Several cue-wrong cells almost never occur under random placement. The balancer therefore passes hints to the sampler, which constructs tangent pairs, occluders over a contact line or a base, near-collinear triples, and a near gap seen side-on with a wider far object. The constructs change the sampling distribution, not the labels. They are recorded per scene. Constructed objects are 80--97\% of every cell, so construct membership enters the regression as a covariate. Appendix Table~\ref{tab:natcon} gives the gap on constructed and natural items separately. Two side effects shape the interpretation. The cue-wrong \family{touching} cells are both occlusion configurations, a contact hidden behind a third object or an overlap in projection. The cue-wrong false \family{between} cell lies near the reject band, since all 30 of its triples deviate from the line by 30--60$^\circ$, against 4 of the 30 cue-right false triples (Appendix Table~\ref{tab:betweenbands}). Generation is deterministic given a seed. Three tiers vary clutter, elevation, shadows and occlusion, each with 600 items (Appendix Table~\ref{tab:splits}). A second medium split comes from an independent seed (Appendix Table~\ref{tab:seed}). View-invariant relations are also rendered from a second camera.

\paragraph{Premise checks.} Three checks run before any model. The 2-D cue scores exactly 50.0\% overall and 100 / 0\% on cue-right / cue-wrong items on every split, verified by a test. The named cues are informative under random placement. They agree with the truth on 82--99\% of medium candidates. Because \family{touching} and \family{between} are rare at random, we also report their balanced accuracies (\n{cue.medium.touching.bal} and \n{cue.medium.between.bal}\%, Appendix Table~\ref{tab:cue}). A learner that weights these cues is not irrational \citep{landy1995cue,knill1996perception}. The analytic masks agree with Blender renders of per-object masks from the tessellated meshes to a median per-object IoU of \n{render.iou}. Relabelling the \n{render.nflip} items whose cue flips on the rendered masks moves no gap by more than \n{r5.render.maxshift} points (Appendix~\ref{app:audit}). A datasheet is in Appendix~\ref{app:datasheet}.

\paragraph{Prompts and models.} Each item is one image and one statement. The instruction says the statement is about the 3-D scene, not the picture, and asks for one word, \emph{True} or \emph{False}, parsed leniently, with unparseable answers counted as wrong (Appendix Table~\ref{tab:parse}). Every prompt is printed in Appendix~\ref{app:prompts}, with filled examples in Appendix~\ref{app:filled}. Claude Opus 5.5, GPT-5.5 and Gemini 3.6 Flash run on every tier for the image condition and on every medium-tier input, at low reasoning effort (we did not switch internal reasoning off), three times on the medium image condition and once at high effort with a 16{,}384-token budget. Grok 4.7, Nova 2 Pro, Pixtral Large, Qwen3-VL-235B, Llama 4 Maverick and Gemma 3 27B run on the inputs of Table~\ref{tab:grid}. All calls go through one LiteLLM proxy with provider-default sampling and lossless PNG images. Appendix Table~\ref{tab:manifest} lists requested and returned model identifiers, routes and dates.

\paragraph{Input forms.} We also run the medium tier with the same scene in other forms. In perspective form we add, as text, every object's visible bounding box, centroid, ground-contact pixel, visible area and hidden fraction, with or without the image, then the same measurements plus each object's camera depth and the field of view. Figure~\ref{fig:perspective} shows these measurements for the running example and what one pixel covers on the ground. A depth map is added as a second image, without and with a metric legend. In ground-plane form we add a to-scale top-down footprint map as a second image, with and without labels, or exact ground coordinates as text, with or without the image. The second camera view changes the picture but not the scene. For the frontier models we add a second set of inputs that isolates single steps. It splits the stated-threshold prompt into \emph{numeric definitions} (each relation's exact rule) and \emph{bands} (the reject bands only). For \family{between} it supplies visible-region centres, ground-contact pixels, or ground-contact pixels with the exact image-to-ground homography, each without the image. For \family{touching} it supplies every pairwise surface distance. These all carry the numeric definitions, so they differ only in the input.

\paragraph{Solvers.} For every text input a deterministic solver reads exactly the serialised text the model received, verified identical to the prompts sent to the model, and applies the benchmark's rule (Table~\ref{tab:solvers}). Where the text lacks a needed quantity, such as object sizes in perspective form, the solver uses the best simple estimate with thresholds fitted on the easy tier. Such cells are marked approximate.

\paragraph{Map task.} A separate task asks the model, from the image alone, for every object's ground position and footprint diameter as JSON in a top-down frame of its choosing, with the scale set by naming one object's diameter. We score maps by the root-mean-square error (RMSE) of their positions after similarity alignment to the scene graph, by the ratio of their depth spread to their lateral spread relative to the truth, and by re-deriving the split's statements from the aligned positions with the true sizes. The alignment uses the truth, so the re-derived scores isolate position error. We align the picture-plane copy, the contact pixels read as top-down coordinates, the same way and compute its depth-to-width ratio. We also feed each frontier model's drawn map back as text, raw or with its depth axis stretched by a factor estimated on the other half of the scenes.

\paragraph{Metrics.} With truth $Y$, cue $H$ and answer $M$, the \sgap{} is $g = P(M{=}Y \mid H{=}Y) - P(M{=}Y \mid H{\ne}Y)$. On a split with equal halves $P(M{=}H) = \tfrac12 + \tfrac{g}{2}$, so GPT-5.5's medium gap of \n{Gpt.medium.image.gap} means its answers agree with the cue on \n{Gpt.medium.image.cuealign}\% of items. The gap describes alignment with the cue. A geometric observer with noisy evidence also shows a positive gap if cue-wrong items are harder for other reasons, which is why the adjustment, the overlap checks and the input forms are needed before reading it as reliance. We resample items for pooled intervals, which gives the same 95\% intervals as resampling scenes because a scene holds at most two items (Opus 5.5: \n{Opus.r5.boot.item} against \n{Opus.r5.boot.scene}). Every paired comparison resamples scenes 2{,}000 times \citep{efron1979bootstrap}. The difficulty-adjusted gap is the average marginal effect of the cue being right in a Firth-penalised logit \citep{firth1993bias} with margin, depth ratio, smallest pixel area, largest occlusion, object count, elevation, polarity (whether the statement is negated), construct membership, relation, and interactions of cue status with relation and margin. We call this pooled fit the shared-slope model, because each covariate has one slope for all relations. The within-relation model fits the same logit to each relation separately. The primary endpoints are the three frontier models' pooled medium gaps, with Bonferroni-corrected 98.3\% intervals of \n{Opus.primary.ci}, \n{Gpt.primary.ci} and \n{Gemini.primary.ci}. Relation-level contrasts are secondary. Of \n{mult.ntests} per-relation image gaps, \n{mult.nholm} survive a Holm correction (\n{mult.nraw} unadjusted, \n{mult.expected} expected under a global null).

\FloatBarrier
\section{Further related work}
\label{app:relwork}

\paragraph{Cue conflict and balanced evaluation.} Cue-conflict stimuli set two cues against each other to reveal which one a network uses, as in the texture--shape contrast of \citet{geirhos2019texture}. Balanced-pair VQA asks one question of two scenes with opposite answers \citep{zhang2016yinyang,goyal2017vqav2}. Prior-shift sets change the answer distribution \citep{agrawal2018vqacp} or score rare pairs separately \citep{kervadec2021gqaood}. Minimal pairs isolate one factor \citep{thrush2022winoground,hendricks2021probing}. Rel3D builds minimally contrastive pairs with 3-D truth \citep{goyal2020rel3d}. NLVR supplies our statement format \citep{suhr2017nlvr,suhr2019nlvr2}, and CLEVR and PTR the rendered-primitive lineage \citep{johnson2017clevr,hong2021ptr}. 3DSRBench balances with complementary image pairs and filters questions that the shortcut ``higher in 3-D is higher in 2-D'' answers \citep{ma2025threedsrbench}. These designs balance labels, remove shortcut items or conflict two cues. \bench{} keeps the shortcut items and balances a named 2-D cue against the 3-D truth, so the picture plane and the ground plane disagree on half of every relation.

\paragraph{Spatial benchmarks and depth cues.} VSR, What's Up and GQA collect spatial relations judged by people from photographs, including front and behind, without metric 3-D truth \citep{liu2023vsr,kamath2023whatsup,hudson2019gqa}. The spatial subsets of BLINK and the 2-D subset of CV-Bench are similar, while CV-Bench's 3-D subset asks depth order and relative distance on Omni3D images \citep{fu2024blink,tong2024cambrian}. SpatialVLM, SpatialRGPT-Bench, SpatialBot, Spatial457, EmbSpatial, SPACE and VSI-Bench scale 3-D-annotated questions \citep{chen2024spatialvlm,cheng2024spatialrgpt,cai2024spatialbot,wang2025spatial457,du2024embspatial,ramakrishnan2025space,yang2025thinking}. SpatialEval compares text-only and visual inputs \citep{wang2024spatialeval}. COMFORT scores consistency across frames of reference \citep{zhang2025comfort} and SpinBench under viewpoint change \citep{yu2025spinbench}. Low-level failures of VLM vision are documented by \citet{rahmanzadehgervi2024blind}, MMVP \citep{tong2024mmvp} and GeoMeter \citep{azad2025geometer}, occlusion failures by CAPTURe \citep{pothiraj2025capture}, O-Bench \citep{obench2025} and \citet{maheshwari2026catalogue}, who localise the loss in one open model to the vision merger. DepthCues probes monocular depth cues in vision encoders \citep{danier2025depthcues}, perceptual-constancy probes test size constancy in VLMs \citep{sun2025constancy}, and O3-D varies pictorial depth cues one at a time \citep{liu2026o3d}. Their vertical-position bias matches the angular-declination cue of human distance perception \citep{ooi2001distance} and the behaviour of self-supervised depth networks \citep{vandijk2019depth}. \citet{min2026whyfar} and \citet{zou2026seeingdepth} are the closest designs. Each puts a cue in conflict with depth order, on renders and photographs. \citet{zou2026seeingdepth} also show that depth remains decodable when the answer is wrong. \bench{} extends the conflict to four further relations and adds input forms that hold the information fixed.

\paragraph{Supplied inputs, maps and text.} Depth helps a VLM as text and can hurt as an image \citep{vo2026dop}. Layered depth captions also help \citep{roy2025bydeway}. Coordinates are hard to use without explicit object association \citep{cheng2024spatialrgpt,yuan2026gr3d}. VLMs also over-trust text that conflicts with the image \citep{deng2025blindfaith}. Bird's-eye-view (BEV) prompting is an established lever. Struct2D and GPT4Scene prompt with BEV images and object marks \citep{zhu2025struct2d,qi2026gpt4scene}, VST shows a teacher model BEV renderings to generate part of its reasoning data \citep{yang2025vst}, VSI-Bench's ground-truth cognitive map lifts relative-distance accuracy from 46\% to 66--78\% \citep{yang2025thinking}, and 3ViewSense argues for a view-consistent representation \citep{zhan2026threeviewsense}. MindCube finds a ground-truth map lowers accuracy unless the model reasons over it \citep{wang2026mindcube}. TopViewRS finds VLMs weak on realistic top-view maps \citep{li2024topviewrs}. Our top-down input is at the easy end of that spectrum. Our homography input supplies the image-to-ground mapping of inverse perspective mapping \citep{mallot1991inverse}. These studies vary the form and the information together. Our inputs hold the information fixed, as a solver checks, and vary only the form.

\FloatBarrier
\section{Open models in full}
\label{app:openfull}

\paragraph{Probes.} Section~\ref{sec:res-probe} introduces the probes, and Appendix~\ref{app:probe} gives their settings. Qwen3-VL-4B-Instruct is of the same family as Qwen3-VL-235B, and every probe is repeated on Qwen3-VL-8B-Instruct. Object features are read at every fourth decoder layer. Image tokens precede the text, so under the causal mask these states do not depend on the statement. Besides the reader of the 2-D measurements, the reference includes gradient-boosted trees that see the measurements of every object in the scene. Decoded maps are scored with the code that scores the drawn maps. The same model also answers every medium item and draws a map under the map prompt. The probe at the answer position uses the \n{ro.nitems} items of both medium seeds. It is scored on the cue-wrong items and compared with a probe trained on labels permuted within each relation and cue cell. Any supervised reader of the input can beat a model, so we also train the probe on cue-right items only and compare it with an input-only reader, gradient-boosted trees on the 2-D measurements of the statement's objects with the relation, the polarity and the cue's verdict, cross-validated over the same scenes.

\paragraph{Steering.} To test whether the answers use the layout we edit the hidden states of both open models and read the change in the log-odds of True over False. At each layer we fit encoding directions, the least-squares map from an object's ground depth, lateral position, picture position and size to its hidden state \citep{haufe2014interpretation}, and add them to the states as in work on steering language models \citep{li2023iti,zou2023repe}. The directions are fitted on the states of all scenes, the steered ones included, which matters little with \n{probe.nobj} objects. A \family{closer} edit swaps the two objects' values along one group of directions, either ground depth, picture position (column, row and size), both together, which keeps the scene physically consistent, or lateral position as a control. A \family{between} edit moves the middle object to the midpoint of the other two, either on the ground or in the picture. Each edit is added to the image tokens the object covers or to the tokens of its name in the statement, at layers 8, 12, 16, 20 and 24 and at one and four times the size that the swap or move implies, on the \n{st.n} \family{closer} and \n{st.n} \family{between} items of the two medium seeds. Random edits with the norm of the combined edit, isotropic or within the top 64 principal components of the states, control for disruption. We report the change separately on the items whose log-odds the edit should raise and those it should lower. A random edit does not move both groups the same way. It pulls the log-odds toward zero, so the split partly follows the sign of the base answer, and we also compare each edit with the random edit within strata of the base log-odds (below). We also report the share of answers that cross to the side the edit implies. Appendix Table~\ref{tab:steergrid} gives every layer and edit.

\subsection{Probes}
\label{app:openfull-probe}

\begin{figure}[h]
\centering
\includegraphics[width=0.62\textwidth]{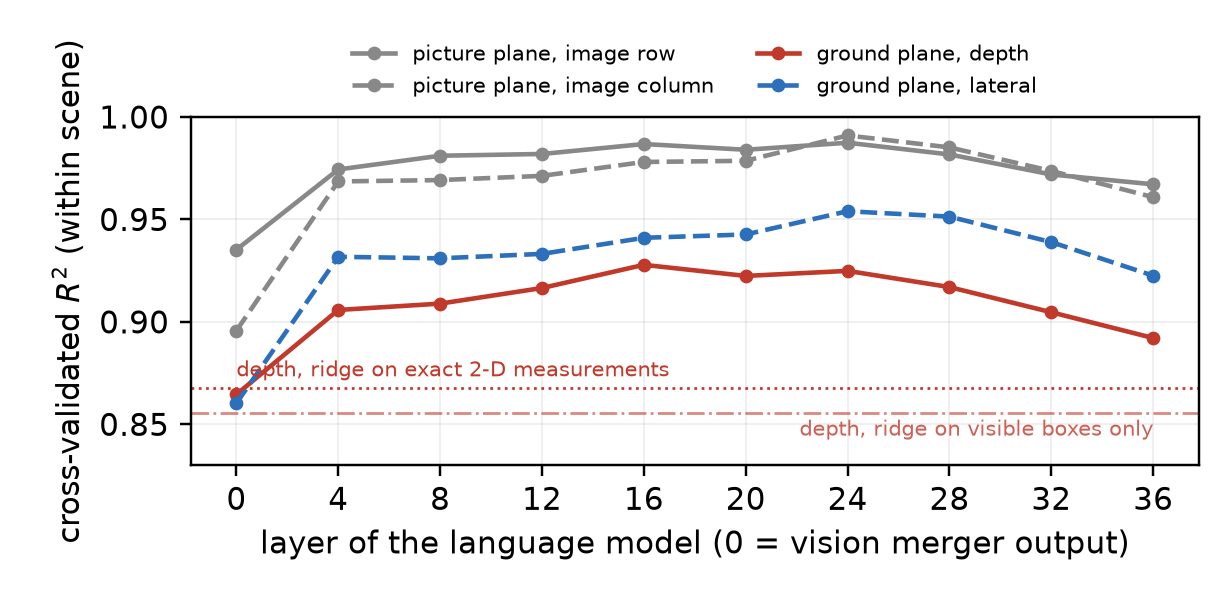}
\caption{Linear probes on Qwen3-VL-4B's object representations, by decoder layer. Ground-plane depth (red) and lateral position (blue) are decodable at every layer and peak in the middle of the network. Picture-plane position (grey) is decoded better still. Horizontal lines give the same ridge readout applied to the exact 2-D measurements, and to the visible boxes alone. $R^2$ is within scene and cross-validated over \n{probe.nscene} scenes.}
\label{fig:probe}
\end{figure}

Qwen3-VL-4B is weak on the task, answering \n{probe.beh.acc}\% of medium items correctly with a gap of \n{probe.beh.gap} points, and the maps it draws keep \n{probe.own.ratio} of the true depth-to-width ratio, with \n{probe.own.reversed}\% of them running depth the wrong way (Table~\ref{tab:probe}). At layer \n{probe.layer} a linear probe recovers each object's relative depth with $R^2$ \n{probe.r2depth}, compared with \n{probe.pix.r2depth} for the same probe on the exact 2-D measurements and \n{probe.gbt.r2depth} for gradient-boosted trees given the measurements of every object in the scene (Figure~\ref{fig:probe}). The difference is \n{r10.rc.4b.paired} \n{r10.rc.4b.paired.ci} in a paired scene bootstrap, and \n{r10.rc.8b.paired} \n{r10.rc.8b.paired.ci} for the 8B. Probes trained on the scenes of the easy split and both medium splits and tested on the hard tier reach $R^2$ \n{r10.rc.4b.ct.probe} for the 4B and \n{r10.rc.8b.ct.probe} for the 8B, against \n{r10.rc.4b.ct.pix} for the reader of the 2-D measurements, when the penalty is chosen by cross-validation grouped by scene. A map decoded from the hidden states is not compressed, with a depth-to-width ratio of \n{probe.map.ratio}, and statements read from it show a gap of \n{probe.map.gap} points. Any reader trained on the true positions is uncompressed in this sense. The ridge reader of the 2-D measurements gives a ratio of \n{probe.pixmap.ratio}.

The hidden states keep about as much of the layout as an exact reading of the picture does, while the model's drawn map and answers keep far less. The probe adds little to the input itself, and depth is decodable with $R^2$ \n{probe.r2depth.l0} already at layer 0, the output of the vision merger (Section~\ref{sec:res-probe}). For depth order, \citet{zou2026seeingdepth} found the analogous dissociation between a decodable readout and the answer, most strongly on rendered scenes. Three cautions bound the claim. Decodability shows that the information is available to a linear reader, not that the model computes with it \citep{belinkov2022probing}, and the probed model is far weaker than the frontier models, whose internals we cannot inspect. Its answers also lack the frontier models' \family{fits} gap (\n{probe.beh.fitsgap} points for the 4B and \n{p8.beh.fitsgap} for the 8B), so the probes follow the information path and do not reach the \family{fits} failure. The decoded map also keeps a \family{fits} gap of \n{probe.map.fitsgap} points, about as large as the exact 2-D measurements give, so the hidden states do not hold a sharper layout than the picture itself provides.

The layout also reaches the words of the statement. At the last token of each object's name the probe reads the object's depth with $R^2$ \n{ro.ment.r2} at layer \n{ro.mlayer}, close to the \n{ro.img.r2} of the image tokens and below the \n{rc.mpix.c} of a polynomial reader of the exact 2-D measurements centred within the scene. At layer 0 the same probe gives \n{ro.ment.r2.l0}, as it should, since the name has not yet attended to the image. The truth of the statement is partly present at the answer position as well. On the items whose cue is wrong, a probe at layer \n{ro.layer} recovers the truth on \n{ro.truth.disc}\% of items \n{ro.truth.disc.ci}, while the model's own answers are right on \n{ro.model.disc}\% and probes trained on labels permuted within each relation and cue cell score \n{r10.rc.4b.perm.mean}\% over 20 draws (SD \n{r10.rc.4b.perm.sd}, at most \n{r10.rc.4b.perm.max}). Part of this advantage comes from \family{closer}, whose cue-wrong items the probe gets \n{ro.truth.closer}\% right. Without \family{closer} the probe is right on \n{ro.nc.probe}\% of the \n{ro.nc.n} cue-wrong items and the model on \n{ro.nc.model}\%, a difference of \n{ro.nc.diff} points \n{ro.nc.diff.ci}. The difference is positive at every layer from 20 to 36, between \n{ro.nc.diffmin} and \n{ro.nc.diffmax} points, and at most \n{ro.nc.early} before layer 20. The advantage does not show that the model forms an answer and then ignores it (Appendix Table~\ref{tab:readoutctl}). The input-only reader plays for the answer probe the role that an image ablation plays for depth probes \citep{zou2026seeingdepth}. It never sees the model and is right on \n{rc.in.nc}\% of the same items. Trained only on the \n{rc.cr.n} cue-right items of each fold, the probe at the answer position beats the answers by \n{rc.cr.diff} points \n{rc.cr.diff.ci}, against \n{r10.rc.4b.rn.mean} for random subsets of the same size (SD \n{r10.rc.4b.rn.sd} over 30 draws), so leaving out the cue-wrong items changes it by \n{r10.rc.4b.crmr} points \n{r10.rc.4b.crmr.ci} at equal training size, an interval that reaches zero. With growing training sets its advantage rises through \n{rc.lc} points (\n{rc.lc.n} items) without levelling off. Choosing the penalty within each training fold leaves it at \n{rc.nest.diff}, and at \n{r10.rc.4b.nc} with the grid of the inverse penalty $C$ extended to $C = 10$, since the search often picks the edge of the original grid. Choosing the layer within each fold as well gives \n{r10.rc.4b.nl} \n{r10.rc.4b.nl.ci}, so the advantage does not depend on picking layer 24 after the fact. A reader trained on the labels of cue-wrong items learns the relation from the state at the answer position, as one can from the measurements alone. The state carries the information. The probe does not show what the model computes from it.

Qwen3-VL-8B shares the 4B's family, items and analysis, with the answer and name layers chosen on the 4B and the image-probe layer chosen again on the 8B, so it is a robustness check rather than independent evidence. It gives the same overall picture with some differences (Table~\ref{tab:probe}). It answers \n{p8.beh.acc}\% of medium items correctly with a gap of \n{p8.beh.gap} points. A probe at layer \n{p8.layer} reads depth with $R^2$ \n{p8.r2depth} \n{p8.r2depth.ci}, above the \n{probe.pix.r2depth} of the reader of the 2-D measurements. The decoded map keeps a depth-to-width ratio of \n{p8.map.ratio}. The maps the 8B model draws keep \n{p8.own.ratio} \n{p8.own.ratio.ci} of the true ratio and run depth the wrong way in \n{p8.own.reversed}\% of the \n{p8.own.nrank} maps with spread on both axes, with a median depth rank correlation of \n{p8.own.rank}. The order in depth is nearly perfect apart from its sign, which looks like copying the rows of the picture into the map, since nearer objects sit lower in the frame. The statements re-derived from the maps align them with reflection allowed, so the sign does not affect them. Unlike the 4B, the 8B decodes depth less well at layer 0 ($R^2$ \n{p8.r2depth.l0}), and its answers show no \family{fits} gap. Its name tokens carry depth with $R^2$ \n{p8.ment.r2} at layer \n{ro.mlayer}. On the cue-wrong items outside \family{closer} its probe at the answer position is right on \n{p8.nc.probe}\% against \n{p8.nc.model}\% for its answers, a difference of \n{p8.nc.diff} points \n{p8.nc.diff.ci} that stays between \n{p8.nc.diffmin} and \n{p8.nc.diffmax} from layer 20 to 36. The input-only reader matches it here too (\n{rcc.in.diff} points over the answers), and trained on cue-right items only the probe gains \n{rcc.cr.diff} \n{rcc.cr.diff.ci}, a change of \n{r10.rc.8b.crmr} \n{r10.rc.8b.crmr.ci} against random subsets of the same size. Choosing its layer within each fold gives \n{r10.rc.8b.nl} \n{r10.rc.8b.nl.ci}.

\subsection{Steering}
\label{sec:res-steer}

\begin{table}[h]
\centering
\caption{Steering Qwen3-VL-4B and 8B at layer 12 with edits four times the size of the change. Entries are mean changes in the log-odds of True over False, on the items whose log-odds the edit should raise (rise) and lower (fall). Bold rises have 95\% item-bootstrap intervals above zero, uncorrected for the number of cells. A ground-depth swap in the image tokens moves \family{closer} answers both ways. The \family{between} edits in the image tokens move its answers both ways by smaller amounts, while the other edits mostly lower every answer, as a disruption does. Appendix Table~\ref{tab:steergrid} gives every layer, the swap of both depth and picture position, an isotropic random edit and the share of flipped answers.}
\label{tab:steer}
\small
\resizebox{\textwidth}{!}{\begin{tabular}{llcccccccc}
\toprule
 & & \multicolumn{4}{c}{image tokens} & \multicolumn{4}{c}{name tokens} \\
\cmidrule(lr){3-6}\cmidrule(lr){7-10}
 & & \multicolumn{2}{c}{4B} & \multicolumn{2}{c}{8B} & \multicolumn{2}{c}{4B} & \multicolumn{2}{c}{8B} \\
relation & edit & rise & fall & rise & fall & rise & fall & rise & fall \\
\midrule
\family{closer} & ground depth & \textbf{+1.23} & $-$2.00 & \textbf{+1.48} & $-$2.09 & $-$0.39 & $-$4.95 & $-$0.32 & $-$5.59 \\
 & picture position & $-$0.13 & $-$1.40 & +0.08 & $-$1.09 & $-$2.12 & $-$9.30 & $-$2.00 & $-$8.68 \\
 & lateral position & $-$0.02 & $-$0.09 & $-$0.06 & $-$0.06 & $-$0.27 & $-$0.63 & $-$0.38 & $-$0.61 \\
 & random (top PCs) & +0.06 & $-$0.90 & +0.01 & $-$0.54 & $-$0.44 & $-$1.60 & $-$0.56 & $-$1.36 \\
\midrule
\family{between} & ground position & \textbf{+0.15} & $-$0.15 & \textbf{+0.16} & $-$0.17 & $-$0.51 & $-$0.26 & $-$0.52 & $-$0.22 \\
 & picture position & \textbf{+0.20} & $-$0.20 & \textbf{+0.34} & $-$0.27 & $-$0.95 & $-$0.07 & $-$0.60 & $-$0.16 \\
 & random (top PCs) & $-$0.00 & $-$0.08 & $-$0.04 & $-$0.02 & $-$0.06 & $-$0.05 & +0.01 & +0.01 \\
\bottomrule
\end{tabular}
}
\end{table}

A decodable layout need not be the one the answer reads, and editing it tests this directly (Table~\ref{tab:steer}). In the image tokens of both models, swapping the ground depths of the two objects in a \family{closer} statement moves the answer the way the swap should. At layers 8 and 12 the log-odds change by \n{st.dep.rise} on the statements the swap makes true and by \n{st.dep.fall} on those it makes false. Swapping the objects' picture positions instead changes them by \n{st.pic.fall} on the second group and by only \n{st.pic.rise} on the first, which is what a random edit of comparable norm also does. Per unit of edit norm the depth swap changes the signed log-odds by \n{st.dep.pn}, against \n{st.pic.pn} for the picture swap and \n{st.rpc.pn} for a random edit. Swapping lateral positions changes nothing (at most \n{st.lat.max}). The effect fades in later layers, with rises of \n{st.l16.dep.rise} at layer 16. At layer 20 none of the depth, picture or ground edits to the image tokens moves either group by more than \n{st.l20.max}, so by then the answer no longer reads the image tokens, at about the depth where image information is reported to stop flowing into the text tokens \citep{zhang2025crossmodal,kaduri2025whats}. In the taxonomy of \citet{liao2026decodable}, the \family{closer} edits fall in the regime in which a decodable variable is also used with the right sign. The effect is real but small. The depth swap at four times the size of the change moves \n{st.flip.dep}\% of the answers to the side it implies at layers 8 and 12, against \n{st.flip.rpc}\% for the random edit. This repeats on our scenes the depth write-back of \citet{zou2026seeingdepth}.

\paragraph{Strata of the base answer.}
\label{app:openfull-steer-strata}
Because random edits pull the log-odds toward zero, we also compare the depth swap at four times the size of the change with the random top-PC edit within strata of the base answer. In the 4B at layer 12 the swap moves the log-odds toward the answer it implies by \n{r10.st.4b.l12.dep.neg_base} on items first answered False and \n{r10.st.4b.l12.dep.pos_base} on items first answered True, against \n{r10.st.4b.l12.rpc.neg_base} and \n{r10.st.4b.l12.rpc.pos_base} for the random edit. In the 8B the values are \n{r10.st.8b.l12.dep.neg_base} and \n{r10.st.8b.l12.dep.pos_base} against \n{r10.st.8b.l12.rpc.neg_base} and \n{r10.st.8b.l12.rpc.pos_base}. The share of answers flipped the intended way exceeds the random edit's by \n{r10.st.4b.l8.fd} \n{r10.st.4b.l8.fd.ci} and \n{r10.st.4b.l12.fd} \n{r10.st.4b.l12.fd.ci} points in the 4B at layers 8 and 12, and by \n{r10.st.8b.l8.fd} \n{r10.st.8b.l8.fd.ci} and \n{r10.st.8b.l12.fd} \n{r10.st.8b.l12.fd.ci} in the 8B, so three of the four intervals exclude zero.

The edits do not separate ground depth from picture position. In these scenes an object's depth correlates with its picture row at $r = \n{rc.col.r}$. Its lateral position, column, row and size explain $R^2$ \n{rc.col.r2} of its depth, so the depth direction carries only the part of depth that the picture does not predict. Swapping it alone gives a state that no real scene produces. The swap of both depth and picture position, which keeps the scene consistent, moves the answers about as far at layer 8 (rises of \n{st.both.rise.l8}) and less at layer 12 (\n{st.both.rise.l12}). For \family{closer}, the control relation, the answer reads the objects' depth from the image tokens, without telling whether it reads it on the ground or in the picture.

For \family{between}, a relation with a gap in the 8B and a smaller, uncertain one in the 4B (Table~\ref{tab:qwenfam}), the ground edit has no advantage over the picture edit. Moving the middle object to the midpoint of the other two raises the log-odds by \n{st.bg.rise} when the move is made on the ground and by \n{st.bp.rise} when it is made in the picture, and lowers them by \n{st.bg.fall} and \n{st.bp.fall} on the statements the move should make false, with effects per unit of norm of \n{st.bg.pn} and \n{st.bp.pn}. Comparisons per unit of norm are rough, because the directions are fitted to variables in different units. Each edit moves the answers both ways, with every interval excluding zero at the 95\% level taken one interval at a time, so the answer reads the image tokens for \family{between} as well. The effects are small. The picture edit moves the answers about twice as far, and the ground edit flips \n{st.flip.bg.l12}\% of answers at layer 12 against \n{st.flip.bgrpc.l12}\% for a random edit. An edit to one object may miss whatever the model compares. Edits to the name tokens move no answer the intended way in either model. They lower the log-odds whatever the edit should do. A depth swap changes them by \n{st.men.dep.fall} on the statements it should make false and by \n{st.men.dep.rise} on those it should make true. The position carried by the name tokens is decodable, but these edits give no evidence that the answer reads it there.

\FloatBarrier
\section{Picture cues that human observers use}
\label{app:cues}

Each relation's cue is the simplest 2-D reading, which is not always the cue that human picture perception relies on \citep{cutting1995perceiving,sedgwick1986space}. People place objects on the ground by their points of contact \citep{gibson1950perception}. For \family{between} our cue uses the centroids of the visible masks. The same 20\textdegree{} rule applied to the projected footprint centres, which lie hidden under the objects and so are not visible to the models, is right on \n{cs.betc.medium.acc}\% of medium items and \n{cs.betc.medium.disc}\% of the cue-wrong ones, because collinearity of ground points survives projection. A reader that sees only visible points, the horizontal centre of each object's visible box at its lowest visible row, is right on \n{r10.btwvis.acc}\% of easy, medium and hard items and \n{r10.btwvis.disc}\% of the cue-wrong ones. The models weight centroid collinearity and give contact collinearity almost no weight (Appendix Table~\ref{tab:cuebank}), which suggests that their \family{between} errors come from placing objects by their image regions rather than by their contacts. For \family{taller} the human cue is the horizon ratio, the object's image height divided by the image distance from its base to the horizon, which gives its height in units of the camera's height when the picture plane is vertical \citep{sedgwick1986space,bertamini1998relative}. Our cameras are pitched down, the visible top of a cube or cylinder seen from above is its far top edge, and occlusion can raise the lowest visible row, so our reader is approximate and its accuracy a lower bound for an exact reader. On the easy and medium tiers it also uses the true horizon row, which lies outside the frame. The horizon lies inside the frame in \n{cs.hz.easy}, \n{cs.hz.medium} and \n{cs.hz.hard}\% of easy, medium and hard scenes, and a reader of the horizon ratio solves \n{cs.tallhz.easy.disc}, \n{cs.tallhz.medium.disc} and \n{cs.tallhz.hard.disc}\% of the cue-wrong \family{taller} items on those tiers. GPT-5.5's and Gemini's \family{taller} gaps are smallest on the hard tier (\n{Gpt.hard.image.taller.gap} and \n{Gemini.hard.image.taller.gap}), where the horizon is visible. Shadows, a strong human cue to contact \citep{mamassian1998cast}, are rendered on the easy and medium tiers only, so the tiers vary shadows, elevation and occlusion together.

\FloatBarrier
\section{Probe details}
\label{app:probe}

\paragraph{Model and features.} We run Qwen3-VL-4B-Instruct from Hugging Face in bfloat16 on Apple-silicon MPS for features, behaviour and the drawn map, and on an NVIDIA L40S for steering, and Qwen3-VL-8B-Instruct in bfloat16 on an L40S throughout, with identical code and settings. The image processor is fixed at $768\times768$, so that each merged visual token covers a $32\times32$-pixel cell and the image yields $24\times24$ tokens. Each scene gets one forward pass with the fixed prompt ``Describe the scene.'' Because the image tokens precede the prompt, their states are the same for any statement. For each object we weight the image-token states by the fraction of each cell that its visible instance mask covers and average them, at decoder layers 0, 4, \ldots, 36, where layer 0 is the output of the vision merger. Qwen3-VL adds the features of vision layers 5, 11 and 17, each through its own merger, to the outputs of the first three decoder layers (DeepStack, \citealp{bai2025qwen3vl}), so the image-token states from layer 4 on also contain intermediate vision features. Objects with no visible pixel have no representation and are dropped, along with scenes that keep fewer than three objects, which leaves \n{probe.nobj} objects in \n{probe.nscene} scenes from the easy, hard and both medium splits.

\paragraph{Targets and readers.} Targets are each object's ground position in the camera's horizontal frame, split into depth along the camera's forward direction and lateral position across it, both centred within the scene. Features are centred within the scene in the same way. Ridge regressions with the penalty chosen by internal cross-validation from $10^{-1}$ to $10^{5}$ are fitted on standardised features, with 5-fold cross-validation grouped by scene. The readers of the 2-D measurements see each object's projected footprint centre, visible box, centroid, square root of visible area, width and height, expanded to second-order polynomial terms for the ridge reader. The visible-only reader drops the footprint centre, which can lie behind an occluder. The gradient-boosted reader (scikit-learn \texttt{HistGradientBoostingRegressor}, 600 iterations, learning rate 0.05) also receives the scene mean and standard deviation of every measurement, so that it can infer the camera from the arrangement of the picture. $R^2$ intervals resample scenes 1{,}000 times.

\paragraph{Picture-plane targets and the residual.} The same probes read the picture-plane copy, the footprint centre's image column and row, with $R^2$ of \n{probe.r2ppu} and \n{probe.r2ppv} at layer \n{probe.layer}, above the ground-plane values at every layer (Figure~\ref{fig:probe}). The hidden states also predict \n{probe.resid.r2} \n{probe.resid.r2.ci} of the depth variance that the reader of the 2-D measurements leaves unexplained. We do not read this as information beyond the picture plane, because the gradient-boosted reader of the same measurements reaches a depth $R^2$ of \n{probe.gbt.r2depth} \n{probe.gbt.r2depth.ci}, above the hidden-state probe. The residual shows only that the hidden states encode depth in a form closer to linear than the raw measurements do.

\paragraph{Behaviour and the drawn map.} For behaviour we feed the benchmark's image-condition prompt for each of the 600 medium items and compare the next-token log-probabilities of ``True'' and ``False''. An exact tie counts as False, which happens on 15 of the 600 items for the 4B and on 12 for the 8B. For the map we use the map prompt of Appendix~\ref{app:prompts} verbatim with greedy decoding and a 700-token budget, and parse and score the responses with the same code as for the other models. Of the \n{probe.own.n} maps, \n{probe.own.parsed} parse. In \n{probe.own.degen} of them every object shares one $x$ or one $y$ coordinate, so the depth-to-width ratio is taken over the other \n{probe.own.nratio}, while the statement scores use all parsed maps. The median depth rank correlation is \n{probe.own.rank} (median absolute value \n{probe.own.absrank} with ties broken by order, and 0.46 with tied coordinates given their average rank, since most of these maps repeat a $y$ value), and \n{probe.own.reversed}\% of the \n{probe.own.nrank} maps with spread on both axes place farther objects at smaller $y$ although the prompt asks for $y$ to increase away from the camera. For the 8B the median absolute value is \n{p8.own.absrank}. The statement scores align each map with reflection allowed, so a reversed axis does not lower them. Statements re-derived from the drawn maps are \n{probe.own.acc}\% correct, which is below chance. Their gap is \n{probe.own.gap} points.

\paragraph{Probes on the statement.} For the \n{ro.nitems} items of both medium seeds we run the image-condition prompt once and keep the hidden states at decoder layers 0, 2, \ldots, 36 at two kinds of position. The answer position is the last token of the prompt, whose next-token distribution gives the model's answer. A mention position is the last token of an object's name, at its last occurrence in the statement. The mention probe is the ridge reader above, fitted on the \n{ro.ment.n} mentions of objects with a visible pixel and compared with the same reader on those objects' image-token states and with the polynomial reader of their 2-D measurements. The probes at the answer position are logistic regressions on standardised states, with the inverse penalty chosen once at layer 18 by 3-fold grouped cross-validation from $10^{-4}$ to $10^{-1}$ and then fixed for every layer, and with 5-fold cross-validation grouped by scene. The permuted control shuffles the truth labels within each relation and cue cell and is fitted in the same way. Intervals resample scenes 1{,}000 times.

\paragraph{Controls for the probe at the answer position.} Table~\ref{tab:readoutctl} scores every reader on the cue-wrong items outside \family{closer}. The input-only reader is scikit-learn's \texttt{HistGradientBoostingClassifier} (300 iterations, learning rate 0.05) on the 11 measurements of each of the statement's objects and their pairwise differences, with the relation, the polarity, the cue's verdict and the number of objects, cross-validated over the same five scene folds as the probe. The cue-right probe is fitted on the cue-right items of each training fold. The comparison probe is fitted on a random subset of each training fold of the same size. The nested probe chooses its penalty by 3-fold grouped cross-validation inside each training fold.

\begin{table}[h]
\centering
\caption{Readers of the statement's truth on the cue-wrong items outside \family{closer} (\n{ro.nc.n} items), with the difference from the model's answers and 95\% scene-bootstrap intervals. The input-only reader, which never sees the model, does as well as the probe at the answer position. Trained without cue-wrong items, the probe scores a few points lower at equal training size, with intervals that reach zero. Its advantage grows with training size.}
\label{tab:readoutctl}
\small
\resizebox{\textwidth}{!}{\begin{tabular}{lcccc}
\toprule
 & \multicolumn{2}{c}{Qwen3-VL-4B} & \multicolumn{2}{c}{Qwen3-VL-8B} \\
\cmidrule(lr){2-3}\cmidrule(lr){4-5}
reader & acc. & minus answers [95\% CI] & acc. & minus answers [95\% CI] \\
\midrule
model's answers & 51.0 & -- & 51.7 & -- \\
answer-position probe, layer 24 (as in the text) & 65.8 & +14.8 [+9.5, +20.0] & 64.8 & +13.1 [+7.5, +18.6] \\
same probe, $C$ chosen within each training fold & 65.8 & +14.8 [+9.5, +20.0] & 65.4 & +13.7 [+8.1, +19.4] \\
same probe, trained on cue-right items only & 54.6 & +3.5 [$-$1.6, +8.5] & 54.4 & +2.7 [$-$2.6, +7.7] \\
same probe, trained on a random subset of that size (mean $\pm$ SD of 30 draws) & 58.1 & +7.0 $\pm$ 2.0 & 57.3 & +5.7 $\pm$ 1.3 \\
cue-right only minus random subset, paired & -- & $-$3.5 [$-$6.9, +0.1] & -- & $-$2.9 [$-$6.5, +0.4] \\
same probe, layer chosen within each training fold & 65.4 & +14.4 [+8.5, +19.9] & 64.2 & +12.5 [+7.0, +17.8] \\
input-only reader (no model) & 65.2 & +14.2 [+8.4, +20.2] & 65.2 & +13.5 [+7.9, +19.9] \\
\bottomrule
\end{tabular}
}
\end{table}

\paragraph{Steering grid.} Table~\ref{tab:steergrid} gives every image-token edit at four times the size of the change. Edits at the size of the change show the same pattern at about a third of the size. For the image tokens the directions are fitted at each layer on the object states of the probe, by least squares from the scene-centred ground depth, lateral position, centroid column and row and square root of visible area to the state, and the edit is added to every image token that the object's visible mask covers by at least 30\%. For the name tokens the directions are fitted on the state at the last token of each name and added to every token of the name, which may explain part of the disruption these edits cause. The two random edits take the norm of the combined edit of depth and picture position.

\begin{table}[h]
\centering
\caption{Image-token steering of Qwen3-VL-4B and 8B at four times the size of the change. Each cell gives the mean change in the log-odds of True over False on the items whose log-odds the edit should raise and on those it should lower. Bold cells have 95\% item-bootstrap intervals that exclude zero in the intended direction for both the rise and the fall. The last column gives the percentage of answers at layer 12 that cross to the side the edit implies and away from it.}
\label{tab:steergrid}
\small
\resizebox{\textwidth}{!}{\begin{tabular}{lllcccccc}
\toprule
model & relation & edit & layer 8 & layer 12 & layer 16 & layer 20 & layer 24 & flips at 12 (\%) \\
\midrule
4B & \family{closer} & ground depth & \textbf{+1.37 / $-$2.01} & \textbf{+1.23 / $-$2.00} & \textbf{+0.30 / $-$0.77} & +0.04 / $-$0.08 & +0.01 / $-$0.04 & 8.8 / 1.7 \\
 &  & picture position & +0.24 / $-$1.87 & $-$0.13 / $-$1.40 & $-$0.00 / $-$0.28 & $-$0.00 / $-$0.14 & +0.02 / $-$0.07 & 3.8 / 3.8 \\
 &  & both & \textbf{+1.47 / $-$3.36} & +0.44 / $-$2.19 & \textbf{+0.54 / $-$0.93} & $-$0.04 / $-$0.14 & +0.02 / $-$0.04 & 7.1 / 2.9 \\
 &  & lateral position & $-$0.02 / $-$0.09 & $-$0.02 / $-$0.09 & $-$0.11 / $-$0.04 & +0.03 / $-$0.05 & $-$0.01 / $-$0.03 & 0.4 / 0.4 \\
 &  & random (isotropic) & +0.20 / $-$0.58 & +0.02 / $-$0.59 & $-$0.05 / $-$0.27 & +0.00 / $-$0.07 & +0.02 / $-$0.05 & 1.2 / 2.5 \\
 &  & random (top PCs) & +0.11 / $-$1.19 & +0.06 / $-$0.90 & $-$0.03 / $-$0.31 & +0.03 / $-$0.07 & +0.01 / $-$0.03 & 2.9 / 2.1 \\
\midrule
 & \family{between} & ground position & \textbf{+0.16 / $-$0.11} & \textbf{+0.15 / $-$0.15} & \textbf{+0.12 / $-$0.10} & $-$0.01 / $-$0.02 & $-$0.01 / $-$0.03 & 3.3 / 0.4 \\
 &  & picture position & \textbf{+0.20 / $-$0.25} & \textbf{+0.20 / $-$0.20} & +0.06 / $-$0.10 & $-$0.01 / $-$0.02 & $-$0.03 / $-$0.04 & 5.0 / 0.4 \\
 &  & both & \textbf{+0.43 / $-$0.25} & \textbf{+0.33 / $-$0.22} & +0.06 / $-$0.11 & $-$0.00 / $-$0.03 & $-$0.03 / $-$0.04 & 4.6 / 1.2 \\
 &  & random (isotropic) & $-$0.03 / $-$0.03 & $-$0.01 / $-$0.02 & +0.01 / $-$0.07 & $-$0.00 / $-$0.01 & $-$0.03 / $-$0.02 & 1.7 / 0.8 \\
 &  & random (top PCs) & $-$0.06 / $-$0.10 & $-$0.00 / $-$0.08 & $-$0.05 / $-$0.05 & $-$0.03 / $-$0.05 & $-$0.02 / $-$0.03 & 4.2 / 0.8 \\
\midrule
8B & \family{closer} & ground depth & \textbf{+1.66 / $-$2.36} & \textbf{+1.48 / $-$2.09} & \textbf{+0.56 / $-$0.73} & +0.02 / $-$0.05 & $-$0.01 / $-$0.00 & 6.7 / 0.0 \\
 &  & picture position & +0.07 / $-$1.78 & +0.08 / $-$1.09 & +0.16 / $-$0.62 & $-$0.04 / $-$0.20 & $-$0.03 / $-$0.01 & 2.1 / 2.5 \\
 &  & both & \textbf{+0.85 / $-$3.09} & +0.32 / $-$1.52 & \textbf{+0.26 / $-$0.83} & $-$0.09 / $-$0.18 & $-$0.01 / $-$0.01 & 4.6 / 2.5 \\
 &  & lateral position & $-$0.04 / $-$0.02 & $-$0.06 / $-$0.06 & $-$0.04 / $-$0.00 & $-$0.05 / $-$0.03 & $-$0.03 / +0.01 & 1.2 / 0.0 \\
 &  & random (isotropic) & $-$0.14 / $-$0.31 & $-$0.05 / $-$0.53 & $-$0.01 / $-$0.44 & $-$0.03 / $-$0.04 & $-$0.01 / +0.01 & 0.0 / 2.1 \\
 &  & random (top PCs) & +0.09 / $-$0.63 & +0.01 / $-$0.54 & +0.03 / $-$0.46 & $-$0.05 / $-$0.10 & $-$0.03 / $-$0.00 & 2.1 / 2.1 \\
\midrule
 & \family{between} & ground position & \textbf{+0.10 / $-$0.14} & \textbf{+0.16 / $-$0.17} & \textbf{+0.11 / $-$0.11} & +0.00 / $-$0.02 & $-$0.01 / $-$0.02 & 0.8 / 2.1 \\
 &  & picture position & \textbf{+0.44 / $-$0.37} & \textbf{+0.34 / $-$0.27} & +0.08 / $-$0.13 & $-$0.04 / $-$0.00 & $-$0.01 / +0.00 & 2.5 / 2.5 \\
 &  & both & \textbf{+0.54 / $-$0.27} & \textbf{+0.33 / $-$0.21} & \textbf{+0.10 / $-$0.14} & $-$0.02 / $-$0.02 & $-$0.01 / +0.01 & 2.5 / 2.5 \\
 &  & random (isotropic) & $-$0.01 / $-$0.01 & $-$0.01 / $-$0.04 & $-$0.02 / $-$0.01 & $-$0.02 / $-$0.01 & $-$0.01 / $-$0.00 & 0.4 / 2.5 \\
 &  & random (top PCs) & $-$0.05 / $-$0.02 & $-$0.04 / $-$0.02 & $-$0.07 / +0.01 & $-$0.01 / $-$0.01 & $-$0.02 / $-$0.02 & 0.0 / 2.5 \\
\bottomrule
\end{tabular}
}
\end{table}

The probe and CV-Bench code will accompany the code release.

\FloatBarrier
\section{Constancy of the maps and the photographic items}
\label{app:constancy}

\paragraph{Constancy index.} For each scene we compute the depth-to-width ratio $m$ of a model's map and the ratio $c$ of the picture-plane copy, both as in Table~\ref{tab:map}, and the index $\ln(m/c)/\ln(1/c)$ of \citet{thouless1931phenomenal}, which is 0 for a map that copies the picture and 1 for an undistorted map. Scenes whose copy is nearly undistorted ($c \geq 0.95$) would be dropped, but no scene on either tier reaches that value. The index has no upper bound, and 12 to 13\% of Opus 5.5's and GPT-5.5's medium-tier maps exceed 1, so we summarise it by the median. Grok 4.7's maps cannot be scored on 80 of 304 medium and 96 of 302 hard scenes, and its index uses the rest. Table~\ref{tab:constancy} gives the median over scenes with a 2{,}000-draw scene-bootstrap interval, and the median ratio within each tertile of camera elevation. The copy's ratio rises with elevation, roughly as the sine of the angle at which the camera sees the ground. Opus 5.5's index stays near 0.6 on both tiers and in every tertile, while GPT-5.5's stays near the copy on the medium tier.

\begin{table}[h]
\centering
\caption{Constancy of the drawn maps. The index is the median over scenes of $\ln(m/c)/\ln(1/c)$, with $m$ the map's depth-to-width ratio and $c$ the picture-plane copy's, so that 0 is a copy of the picture and 1 the scene. Opus 5.5's index is about 0.6 on both tiers, while GPT-5.5's medium-tier maps are copies of the picture.}
\label{tab:constancy}
\small
\resizebox{\textwidth}{!}{\begin{tabular}{llcccccc}
\toprule
 & & & & \multicolumn{3}{c}{depth/width ratio by camera-elevation tertile} & \\
\cmidrule(lr){5-7}
tier & map & scenes & constancy index [95\% CI] & low & middle & high & median ratio \\
\midrule
medium & picture-plane copy & -- & 0 & 0.51 & 0.60 & 0.69 & 0.61 \\
 & Claude Opus 5.5 & 304 & 0.60 [0.58, 0.63] & 0.76 & 0.83 & 0.88 & 0.83 \\
 & GPT-5.5 & 304 & $-$0.01 [$-$0.07, 0.07] & 0.53 & 0.61 & 0.72 & 0.61 \\
 & Gemini 3.6 Flash & 304 & 0.07 [0.01, 0.12] & 0.57 & 0.61 & 0.70 & 0.63 \\
 & Grok 4.7 & 224 & 0.37 [0.30, 0.45] & 0.65 & 0.75 & 0.81 & 0.75 \\
 & \multicolumn{3}{l}{elevation tertiles: below 32$^\circ$, 32--39$^\circ$, above 39$^\circ$} & & & & \\
\midrule
hard & picture-plane copy & -- & 0 & 0.24 & 0.30 & 0.36 & 0.30 \\
 & Claude Opus 5.5 & 302 & 0.63 [0.61, 0.65] & 0.62 & 0.63 & 0.68 & 0.64 \\
 & GPT-5.5 & 302 & 0.12 [0.08, 0.15] & 0.29 & 0.37 & 0.40 & 0.35 \\
 & Gemini 3.6 Flash & 301 & 0.37 [0.35, 0.41] & 0.46 & 0.46 & 0.49 & 0.47 \\
 & Grok 4.7 & 206 & 0.50 [0.43, 0.55] & 0.50 & 0.55 & 0.57 & 0.54 \\
 & \multicolumn{3}{l}{elevation tertiles: below 13$^\circ$, 13--17$^\circ$, above 17$^\circ$} & & & & \\
\bottomrule
\end{tabular}
}
\end{table}

\paragraph{A two-parameter fit.} The index assumes that a map corrects a fixed share of the picture's compression with no further bias. We test this by regressing $\ln m$ on $\ln c$ over the medium and hard scenes of each model, $\ln m = a + b \ln c$, where $1-b$ is the share corrected and $a$ a uniform bias (Table~\ref{tab:constancyfit}). Opus 5.5 and Grok 4.7 have intercepts near zero and correct \n{r10.cf.Opus.k} and \n{r10.cf.Grok.k} of the compression. GPT-5.5 corrects \n{r10.cf.Gpt.k}. Gemini corrects \n{r10.cf.Gemini.k} but has an intercept of \n{r10.cf.Gemini.a} \n{r10.cf.Gemini.aci}, a further uniform compression of depth by a factor of \n{r10.cf.Gemini.factor}.

\begin{table}[h]
\centering
\caption{Constancy of the drawn maps as a fit of $\ln m = a + b \ln c$ over the medium and hard scenes, with 2{,}000-draw scene-bootstrap intervals. A copy of the picture has $a = 0$ and $b = 1$, and an undistorted map has $a = 0$ and $b = 0$.}
\label{tab:constancyfit}
\small
\begin{tabular}{lcccc}
\toprule
map & scenes & intercept $a$ [95\% CI] & slope $b$ [95\% CI] & share corrected $1-b$ \\
\midrule
Claude Opus 5.5 & 606 & $-$0.01 [$-$0.04, +0.02] & 0.36 [0.32, 0.39] & 0.64 \\
GPT-5.5 & 606 & $-$0.01 [$-$0.07, +0.05] & 0.82 [0.76, 0.89] & 0.18 \\
Gemini 3.6 Flash & 605 & $-$0.25 [$-$0.30, $-$0.19] & 0.42 [0.36, 0.48] & 0.58 \\
Grok 4.7 & 430 & $-$0.06 [$-$0.14, +0.02] & 0.40 [0.29, 0.49] & 0.60 \\
\bottomrule
\end{tabular}

\end{table}

\paragraph{Photographic items by source.} Table~\ref{tab:cvsrc} splits the CV-Bench gaps of Table~\ref{tab:cvbench} by source dataset and recomputes them with each source left out. The nuScenes street scenes give the largest Depth gaps for every model except GPT-5.5, on only 12 cue-wrong items, and without them the frontier Depth gaps are \n{cvs.depth.nonus} points for Opus 5.5, GPT-5.5 and Gemini.

\begin{table}[h]
\centering
\caption{The CV-Bench \sgap{} (points, 95\% item-bootstrap intervals) within each source dataset and with each source left out. The first row of each block gives the number of items and, in parentheses, of cue-wrong items per source.}
\label{tab:cvsrc}
\small
\resizebox{\textwidth}{!}{\begin{tabular}{llcccccc}
\toprule
 & & \multicolumn{3}{c}{one source only} & \multicolumn{3}{c}{leaving one source out} \\
\cmidrule(lr){3-5}\cmidrule(lr){6-8}
task (cue) & model & SUN RGB-D & nuScenes & Hypersim & without SUN RGB-D & without nuScenes & without Hypersim \\
\midrule
Depth (lower box) & items (cue-wrong) & 200 (36) & 200 (12) & 200 (40) & & & \\
 & Claude Opus 5.5 & +8 [+0, +19] & +32 [+7, +58] & +5 [+0, +12] & +11 [+3, +21] & +7 [+1, +13] & +14 [+5, +25] \\
 & GPT-5.5 & +33 [+17, +50] & +29 [+4, +57] & +3 [$-$4, +11] & +8 [+0, +17] & +17 [+9, +26] & +31 [+17, +44] \\
 & Gemini 3.6 Flash & +22 [+8, +36] & +57 [+30, +82] & +11 [+1, +22] & +21 [+11, +33] & +16 [+8, +25] & +30 [+18, +44] \\
 & Qwen3-VL-4B & +2 [$-$1, +8] & +35 [+9, +61] & +3 [$-$3, +11] & +9 [+1, +19] & +3 [$-$1, +7] & +9 [+0, +19] \\
 & Qwen3-VL-8B & +6 [+0, +14] & +39 [+12, +66] & +3 [$-$3, +11] & +8 [$-$1, +20] & +4 [+0, +10] & +11 [+1, +22] \\
\midrule
Distance (pixel distance) & items (cue-wrong) & 200 (14) & 200 (65) & 200 (40) & & & \\
 & Claude Opus 5.5 & +14 [+0, +36] & +16 [+7, +26] & +24 [+11, +38] & +20 [+12, +27] & +22 [+11, +33] & +17 [+9, +26] \\
 & GPT-5.5 & +14 [+0, +36] & +65 [+54, +77] & +30 [+18, +45] & +52 [+43, +61] & +26 [+15, +39] & +57 [+46, +68] \\
 & Gemini 3.6 Flash & +7 [+0, +21] & +36 [+23, +49] & +19 [+6, +32] & +30 [+21, +40] & +17 [+7, +28] & +33 [+23, +44] \\
 & Qwen3-VL-4B & +7 [+0, +21] & +51 [+38, +64] & +9 [$-$2, +21] & +35 [+26, +45] & +10 [+2, +20] & +47 [+36, +58] \\
 & Qwen3-VL-8B & +0 [+0, +0] & +71 [+59, +81] & +15 [+5, +28] & +50 [+40, +59] & +11 [+4, +20] & +59 [+48, +70] \\
\bottomrule
\end{tabular}
}
\end{table}

\paragraph{Photographs in more detail.} The cue-wrong items hold 9 of Opus 5.5's 11 Depth errors, 18 of GPT-5.5's 30 and 20 of Gemini's 26, and 24 of 28, 57 of 58 and 37 of 50 of their Distance errors. nuScenes supplies 65 of the 119 cue-wrong Distance items. CV-Bench releases no 3-D depths or distances, so the gaps cannot be stratified by the 3-D ratio without the Omni3D annotations, which we did not use. Accuracy on CV-Bench Depth is \n{cv.Opus.depth.acc}\% for Opus 5.5 and at least \n{cv.Gpt.depth.acc}\% for the other frontier models. The lower-box gap is in line with the entanglement of height in the frame with distance that \citet{min2026whyfar} report. The larger box, which is right on only \n{cv.cue.area}\% of items, carries a gap of at most +3 points, with no interval above zero. The two probed models show gaps of \n{cv.Qwenall.dist.gaprange} points on Distance. The two Depth cues do not add up, which is inconsistent with an observer that adds them and points to effects of item type. Accuracy is lowest where only the lower box misleads (\n{cvc.onlybottom}\% for Opus 5.5, GPT-5.5 and Gemini on \n{cvc.onlybottom.n} items) and higher where the larger box misleads as well (\n{cvc.bothwrong}\% on \n{cvc.bothwrong.n} items). An observer that weights a cue right on \n{cv.cue.bottom}\% of photographs will err where it fails \citep{brunswik1956perception}, so these errors show the cue in use without showing that its use is irrational. In the rendered \family{closer} items height in the frame cannot be tested, because every object rests on the ground. The elevated tier, a small pilot split with objects on pedestals and stacks, balances \family{closer} on height in the frame, and there the frontier models answer \n{elev.closer.right} of its \n{elev.closer.n} cue-wrong \family{closer} items correctly (Appendix Table~\ref{tab:smokes}). Where the support surface is visible, they do not follow height in the frame on these \n{elev.closer.n} items. Their Depth answers correlate with the lower-box cue at \n{r10.lens.Depth.util}, slightly above the cue's own correlation with the truth of \n{r10.lens.Depth.val}, so their errors lean toward the cue. On Distance, Opus 5.5, GPT-5.5 and Gemini correlate with the pixel-distance cue at \n{r10.lens.Distance.Opus.util}, \n{r10.lens.Distance.Gpt.util} and \n{r10.lens.Distance.Gemini.util} against the cue's own correlation with the truth of \n{r10.lens.Distance.val}, and follow it on \n{r10.lens.Distance.Opus.follow}, \n{r10.lens.Distance.Gpt.follow} and \n{r10.lens.Distance.Gemini.follow}\% of the items where it is wrong.

\FloatBarrier
\section{Prompts}
\label{app:prompts}
\small
Placeholders in braces are filled per item, and Appendix~\ref{app:filled} shows filled examples. The second-view input uses the image prompt on the other render.

\paragraph{image (default).} \texttt{Look at the image and decide whether the statement is true or false. The statement is about the actual 3D scene shown in the image (real sizes, distances and positions in the scene), not about the 2D picture. Statement: \{statement\} Answer with exactly one word: True or False.}

\paragraph{chain of thought (suffix replacing the one-word instruction).} \texttt{Reason step by step about the 3D geometry first. Then give your final answer on its own last line, in exactly the form 'Answer: True' or 'Answer: False'.}

\paragraph{stated thresholds: numeric definitions and reject bands (prepended to the image prompt).} \texttt{Definitions used by the statements: 'closer to the camera/viewer' compares the distance from the camera to each object's centre (the statements only use pairs whose distances differ by at least 15\%). 'Taller' compares the objects' real heights (at least 25\% apart). 'Touching/in contact' means the surfaces are within 0.02 units; 'not touching' items are at least half the smaller object's size apart. 'Between' means the object's ground position is within 10 degrees of the straight ground line joining the other two, as seen from either endpoint; 'not between' items are at least 30 degrees off that line. 'Fits/can pass through the gap' means the free gap between the two objects' footprints on the ground is at least 1.2 times the moving object's footprint width; 'does not fit' means it is at most 0.8 times. }

\paragraph{numeric definitions only, no bands (prepended, used in the prompt factorial of Table~\ref{tab:factorial} and in the single-step inputs of Appendix~\ref{app:details}).} \texttt{Definitions used by the statements: 'closer to the camera/viewer' compares the depth of each object's centre along the camera's viewing direction. 'Taller' compares the objects' real heights. 'Touching/in contact' means the two objects' surfaces are within 0.02 units of each other. 'Between' means that, on the ground, the angle at each of the two end objects between the line to the other end object and the line to the middle object is at most 10 degrees. 'Fits/can pass through the gap' means the free gap between the two objects' footprints on the ground is wider than the moving object's footprint width. }

\paragraph{reject bands only (prepended, used in the prompt factorial of Table~\ref{tab:factorial}).} \texttt{About the statements: they only use clear cases. Objects compared for closeness differ in depth by at least 15\%, and objects compared for height differ by at least 25\%. Objects described as not touching are at least half the smaller object's size apart. Objects described as not between two others are at least 30 degrees off the line joining them. In statements about fitting through a gap, the gap is either at least 1.2 times or at most 0.8 times the moving object's width. }

\paragraph{verbal definitions without numbers or bands (prepended).} \texttt{Definitions used by the statements: 'closer to the camera/viewer' compares how far each object's centre is from the camera. 'Taller' compares the objects' real heights. 'Touching/in contact' means the two objects' surfaces actually meet; objects that merely look close or overlap in the picture but have space between them are not touching. 'Between' means the object lies on (or very near) the straight line joining the other two; an object clearly off that line is not between them. 'Fits/can pass through the gap' means the free space between the two objects is wider than the moving object; if the object is wider than the space it does not fit. }

\paragraph{text only.} \texttt{Decide whether the following statement about a scene is true or false. No image is provided, so answer as best you can from the statement alone. Statement: \{statement\} Answer with exactly one word: True or False.}

\paragraph{pixel measurements.} \texttt{You are shown a rendered 3D scene and, below, exact 2D measurements of every object in the picture, in pixels (image is \{w\}x\{h\}; x grows to the right, y grows downward). For each object: the bounding box of its visible pixels, the centroid of its visible pixels, the image position of the point where it touches the ground, the number of visible pixels, and the fraction of it hidden by other objects; then the minimum pixel distance between every pair of visible silhouettes (0 = overlapping). Use the image and these measurements to decide whether the statement is true or false. The statement is about the actual 3D scene (real sizes, distances and positions), not about the 2D picture. 2D measurements: \{geometry\} Statement: \{statement\} Answer with exactly one word: True or False.}

\paragraph{pixel measurements, no image.} \texttt{Below are exact 2D measurements of every object in a picture of a 3D scene, in pixels (image is \{w\}x\{h\}; x grows to the right, y grows downward; the camera looks slightly down at objects resting on a flat ground). For each object: the bounding box of its visible pixels, the centroid of its visible pixels, the image position of the point where it touches the ground, the number of visible pixels, and the fraction of it hidden by other objects; then the minimum pixel distance between every pair of visible silhouettes (0 = overlapping). No image is provided. Decide whether the statement is true or false. The statement is about the actual 3D scene (real sizes, distances and positions), not about the 2D picture. 2D measurements: \{geometry\} Statement: \{statement\} Answer with exactly one word: True or False.}

\paragraph{pixel measurements + camera depth.} \texttt{You are shown a rendered 3D scene and, below, exact measurements of every object: its 2D measurements in the picture, in pixels (image is \{w\}x\{h\}; x grows to the right, y grows downward): the bounding box of its visible pixels, the centroid of its visible pixels, the image position of the point where it touches the ground, the number of visible pixels and the fraction hidden by other objects; and its exact distance from the camera in scene units (the same units as the objects' sizes), for its centre and for its ground-contact point. The camera's horizontal field of view is \{hfov:.1f\} degrees. Use the image and these measurements to decide whether the statement is true or false. The statement is about the actual 3D scene (real sizes, distances and positions), not about the 2D picture. Measurements: \{geometry\} Statement: \{statement\} Answer with exactly one word: True or False.}

\paragraph{depth map (no legend).} \texttt{You are shown two images. The first is a rendered 3D scene. The second is the exact ground-truth depth map of the same view, colour-coded: bright yellow is closest to the camera, green is intermediate, dark blue/purple is farthest, and black is sky. Use both images to decide whether the statement is true or false. The statement is about the actual 3D scene (real sizes, distances and positions), not about the 2D picture. Statement: \{statement\} Answer with exactly one word: True or False.}

\paragraph{depth map with metric legend.} \texttt{You are shown two images. The first is a rendered 3D scene. The second is the exact ground-truth depth map of the same view with a numeric colour bar: the colour of every pixel encodes its distance from the camera in scene units (the same units as the objects' sizes), as read off the bar; the camera's horizontal field of view is \{hfov:.1f\} degrees. Use both images to decide whether the statement is true or false. The statement is about the actual 3D scene (real sizes, distances and positions), not about the 2D picture. Statement: \{statement\} Answer with exactly one word: True or False.}

\paragraph{camera depth (text).} \texttt{You are shown a rendered 3D scene and, below, the exact distance from the camera to each object, in scene units (the same units as the objects' sizes). Use the image and the distances to decide whether the statement is true or false. The statement is about the actual 3D scene (real sizes, distances and positions), not about the 2D picture. Camera distances: \{geometry\} Statement: \{statement\} Answer with exactly one word: True or False.}

\paragraph{top-down map.} \texttt{You are shown two images. The first is a rendered 3D scene. The second is an exact top-down map of the same scene: each object's footprint on the ground is drawn to scale in the object's colour and labelled, the grid lines are one scene unit apart, and the camera's position and viewing direction are marked. Heights are not shown on the map. Use both images to decide whether the statement is true or false. The statement is about the actual 3D scene (real sizes, distances and positions), not about the 2D picture. Statement: \{statement\} Answer with exactly one word: True or False.}

\paragraph{top-down map without labels.} \texttt{You are shown two images. The first is a rendered 3D scene. The second is an exact top-down map of the same scene: each object's footprint on the ground is drawn to scale in the object's colour (no text labels), the grid lines are one scene unit apart, and the camera's position and viewing direction are marked. Heights are not shown on the map. Identify the objects by their colours. Use both images to decide whether the statement is true or false. The statement is about the actual 3D scene (real sizes, distances and positions), not about the 2D picture. Statement: \{statement\} Answer with exactly one word: True or False.}

\paragraph{ground coordinates (text) with image.} \texttt{You are shown a rendered 3D scene and, below, the exact ground-truth geometry of every object in it. Coordinates are in a world frame with the ground plane at z = 0, in the same length units as the sizes; the camera position is given too. Use the image and the geometry together to decide whether the statement is true or false. The statement is about the actual 3D scene (real sizes, distances and positions), not about the 2D picture. Scene geometry: \{geometry\} Statement: \{statement\} Answer with exactly one word: True or False.}

\paragraph{ground coordinates only (no image).} \texttt{Below is the exact geometry of every object in a 3D scene. Coordinates are in a world frame with the ground plane at z = 0, in the same length units as the sizes; the camera position is given too. No image is provided. Decide whether the statement about this scene is true or false. Scene geometry: \{geometry\} Statement: \{statement\} Answer with exactly one word: True or False.}

\paragraph{visible-region centres only (no image).} \texttt{Below is the image position, in pixels, of the centre of each object's visible region in a picture of a 3D scene (image is \{w\}x\{h\}; x grows to the right, y grows downward; the camera looks down at objects resting on a flat ground). No image is provided. Decide whether the statement about the 3D scene is true or false. Visible-region centres: \{geometry\} Statement: \{statement\} Answer with exactly one word: True or False.}

\paragraph{ground-contact pixels only (no image).} \texttt{Below is the exact image position, in pixels, of the point where each object's footprint centre meets the ground, in a picture of a 3D scene (image is \{w\}x\{h\}; x grows to the right, y grows downward; the camera looks down at objects resting on a flat ground). No image is provided. Decide whether the statement about the 3D scene is true or false. Ground-contact points: \{geometry\} Statement: \{statement\} Answer with exactly one word: True or False.}

\paragraph{ground-contact pixels + homography (no image).} \texttt{Below is the exact image position, in pixels, of the point where each object's footprint centre meets the ground, in a picture of a 3D scene (image is \{w\}x\{h\}; x grows to the right, y grows downward), and the exact homography H that maps image pixels to ground-plane coordinates in scene units: for a pixel (u, v), compute (X, Y, W) = H (u, v, 1) and the ground point is (X/W, Y/W). No image is provided. Decide whether the statement about the 3D scene is true or false. Ground-contact points: \{geometry\} Statement: \{statement\} Answer with exactly one word: True or False.}

\paragraph{pairwise surface distances with image.} \texttt{You are shown a rendered 3D scene and, below, the exact minimum distance between the surfaces of every pair of objects in it, in scene units (the same units as the objects' sizes; 0 means the surfaces meet). Use the image and the distances to decide whether the statement is true or false. The statement is about the actual 3D scene (real sizes, distances and positions), not about the 2D picture. Surface distances: \{geometry\} Statement: \{statement\} Answer with exactly one word: True or False.}

\paragraph{drawn map fed back (no image, raw or depth-corrected layout).} \texttt{Below is a top-down layout of a 3D scene: each object's ground position (x, y) and footprint diameter, in scene units. x increases to the right as seen from the camera and y increases away from the camera. No image is provided. Decide whether the statement about this scene is true or false. Layout: \{geometry\} Statement: \{statement\} Answer with exactly one word: True or False.}

\paragraph{map reconstruction task (separate task, JSON output).} \texttt{Look at the image of a 3D scene. All objects rest on a flat ground plane. Estimate the layout of the scene as seen from directly above: give each object's ground position (x, y) in a top-down coordinate frame of your choosing, in units of the scene (take the \{ref\} as having footprint diameter \{refsize:.2f\} units to set the scale), together with its footprint diameter. x should increase to the right as seen from the camera and y should increase away from the camera. Objects: \{names\}. Answer with a JSON object only, of the form \{\{"$<$object name$>$": \{\{"x": $<$number$>$, "y": $<$number$>$, "diameter": $<$number$>$\}\}, ...\}\} with one entry per object and no other text.}

\paragraph{tool access, variant 1 (tools available, system prompt, answer tool offered from the first turn).} \texttt{You judge statements about 3D scenes from a single rendered image. You may use tools to inspect the image: zoom into regions and run Python over the pixel array to measure things (positions, extents, colours, apparent sizes). You have no access to ground truth. Reason about the actual 3D geometry (real sizes, distances and positions), not the 2D picture, and finish by calling `answer`. Use at most 10 tool calls.}

\paragraph{tool access, variant 2 (tools required first, system prompt, answer tool withheld for two turns).} \texttt{You judge statements about 3D scenes from a single rendered image. You may use tools to inspect the image: zoom into regions and run Python over the pixel array to measure things (positions, extents, colours, apparent sizes). You have no access to ground truth. Reason about the actual 3D geometry (real sizes, distances and positions), not the 2D picture. Before answering, measure the quantities the statement depends on (e.g. silhouette extents, ground-contact pixel rows, gaps, apparent sizes) with the tools, and use perspective reasoning to convert them to 3D. Finish by calling `answer`. Use at most 10 tool calls.}

\paragraph{tool access, user turn (both variants, the image prompt without its last line, followed by).} \texttt{Inspect the image with the tools as needed, then call `answer`.}

\paragraph{tool access, variant 2 nudge (sent if the model answers in prose during the first two turns).} \texttt{Before answering, inspect the image with `crop\_zoom` or measure with `python`. The `answer` tool becomes available after you have inspected.}

\FloatBarrier
\section{Additional tables}
\label{app:extra}

\begin{table}[h]
\centering\small
\caption{Difficulty tiers (generator seed 7). ``Scenes sampled'' counts scenes the balancer had to draw to fill all 20 cells.}
\label{tab:splits}
\begin{tabular}{lccccrrr}
\toprule
tier & objects & elevation ($^\circ$) & shadows & occlusion & items & scenes & scenes sampled \\
\midrule
easy & 4--5 & 45--60 & on & low & 600 & 343 & 1402 \\
medium & 5--7 & 25--45 & on & medium & 600 & 304 & 315 \\
hard & 6--8 & 8--20 & off & high & 600 & 302 & 313 \\
\bottomrule
\end{tabular}

\end{table}

\begin{table}[h]
\centering\small
\caption{Base-rate reliability of each 2-D cue over all candidate statements in 150 freshly sampled scenes per tier under random placement. Entries are agreement / balanced accuracy / prevalence of the true proposition (\%), with the number of candidates in parentheses. The last column gives the agreement of the lower-in-the-frame cue for \family{closer}.}
\label{tab:cue}
\resizebox{\textwidth}{!}{\begin{tabular}{lrrrrrr}
\toprule
tier & closer & taller & touching & between & fits & closer: lower in frame \\
\midrule
easy & 63 / 63 / 50 (532) & 87 / 87 / 50 (1754) & 96 / 98 / 3 (1008) & 100 / 99 / 3 (2741) & 89 / 91 / 63 (1722) & 100 (532) \\
medium & 58 / 58 / 50 (1534) & 91 / 91 / 50 (3062) & 89 / 94 / 2 (1824) & 99 / 97 / 3 (7830) & 82 / 84 / 59 (4172) & 100 (1534) \\
hard & 52 / 52 / 50 (2314) & 95 / 95 / 50 (3770) & 84 / 89 / 1 (2208) & 94 / 89 / 3 (10181) & 76 / 82 / 66 (5453) & 100 (2314) \\
\bottomrule
\end{tabular}
}
\end{table}

\begin{table}[h]
\centering\small
\caption{Passage relation and elevated tier, two small pilot splits, image condition, every model. A passage statement asks whether one object could be slid in a straight line to another without touching any other object. The elevated tier puts objects on pedestals and stacks and balances \family{closer} on height in the frame.}
\label{tab:smokes}
\begin{tabular}{l rl rl}
\toprule
 & \multicolumn{2}{c}{passage (64 items)} & \multicolumn{2}{c}{elevated tier (96 items)} \\
\cmidrule(lr){2-3}\cmidrule(lr){4-5}
model & acc & gap [95\% CI] & acc & gap [95\% CI] \\
\midrule
Claude Opus 5.5 & 89.1 & +15.6 [+1, +31] & 100.0 & +0.0 [+0, +0] \\
GPT-5.5 & 84.4 & +12.5 [$-$5, +30] & 96.9 & +6.2 [+0, +14] \\
Gemini 3.6 Flash & 81.2 & +6.2 [$-$12, +25] & 96.9 & +6.2 [+0, +14] \\
Pixtral Large & 57.8 & +3.1 [$-$22, +28] & 62.5 & +20.8 [+1, +40] \\
Qwen3-VL-235B & 64.1 & +3.1 [$-$20, +27] & 89.6 & +12.5 [+0, +25] \\
Llama 4 Maverick & 56.2 & $-$6.2 [$-$30, +17] & 71.9 & +14.6 [$-$4, +32] \\
Gemma 3 27B & 57.8 & +3.1 [$-$21, +28] & 56.2 & +16.7 [$-$2, +36] \\
Grok 4.7 & 75.0 & +12.5 [$-$9, +34] & 89.6 & +20.8 [+10, +34] \\
Nova 2 Pro & 62.5 & +0.0 [$-$22, +22] & 70.8 & +4.2 [$-$15, +23] \\
\bottomrule
\end{tabular}

\end{table}

\FloatBarrier
{\setlength{\tabcolsep}{3pt}\renewcommand{\small}{\scriptsize}{\small\begin{longtable}{llrrrrl}
\caption{All inputs on the medium tier with 95\% scene-bootstrap CIs. Second-view rows cover the 465 view-invariant items and use the second camera's own cue.}\label{tab:conditions}\\
\toprule
model & input & $n$ & acc & cue right & cue wrong & gap [95\% CI] \\
\midrule
\endfirsthead
\toprule
model & input & $n$ & acc & cue right & cue wrong & gap [95\% CI] \\
\midrule
\endhead
\bottomrule
\endfoot
Claude Opus 5.5 & image & 600 & 85.7 & 96.3 & 75.0 & +21.3 [+16, +27] \\
Claude Opus 5.5 & image + CoT & 600 & 86.5 & 95.0 & 78.0 & +17.0 [+12, +22] \\
Claude Opus 5.5 & image + tools (available) & 600 & 86.3 & 94.3 & 78.3 & +16.0 [+11, +21] \\
Claude Opus 5.5 & image + tools (required first) & 600 & 86.0 & 94.7 & 77.3 & +17.3 [+12, +23] \\
Claude Opus 5.5 & image + stated thresholds & 600 & 93.2 & 99.0 & 87.3 & +11.7 [+8, +16] \\
Claude Opus 5.5 & image + numeric definitions only & 600 & 89.0 & 96.0 & 82.0 & +14.0 [+9, +19] \\
Claude Opus 5.5 & image + reject bands only & 600 & 92.3 & 98.3 & 86.3 & +12.0 [+8, +16] \\
Claude Opus 5.5 & image + numeric definitions + bands & 600 & 94.2 & 98.7 & 89.7 & +9.0 [+5, +13] \\
Claude Opus 5.5 & image + verbal definitions & 600 & 88.0 & 95.3 & 80.7 & +14.7 [+10, +19] \\
Claude Opus 5.5 & text only & 600 & 50.3 & 48.3 & 52.3 & $-$4.0 [$-$11, +4] \\
Claude Opus 5.5 & image + exact pixel measurements (text) & 600 & 86.5 & 94.0 & 79.0 & +15.0 [+10, +20] \\
Claude Opus 5.5 & pixel measurements, no image & 600 & 86.3 & 92.3 & 80.3 & +12.0 [+7, +18] \\
Claude Opus 5.5 & image + pixel measurements + camera depth (text) & 600 & 89.2 & 97.0 & 81.3 & +15.7 [+11, +21] \\
Claude Opus 5.5 & image + depth map (no legend) & 600 & 86.7 & 93.3 & 80.0 & +13.3 [+8, +19] \\
Claude Opus 5.5 & image + depth map (metric legend) & 600 & 86.5 & 94.7 & 78.3 & +16.3 [+11, +22] \\
Claude Opus 5.5 & image + camera depth (text) & 600 & 88.7 & 97.7 & 79.7 & +18.0 [+14, +23] \\
Claude Opus 5.5 & image + top-down map & 600 & 93.0 & 95.0 & 91.0 & +4.0 [$-$0, +8] \\
Claude Opus 5.5 & image + top-down map, no labels & 600 & 92.5 & 95.7 & 89.3 & +6.3 [+2, +11] \\
Claude Opus 5.5 & image + ground coordinates (text) & 600 & 95.7 & 98.0 & 93.3 & +4.7 [+1, +8] \\
Claude Opus 5.5 & ground coordinates only (no image) & 600 & 95.5 & 98.7 & 92.3 & +6.3 [+3, +10] \\
Claude Opus 5.5 & ground coordinates only + verbal definitions & 600 & 97.5 & 99.0 & 96.0 & +3.0 [+1, +5] \\
Claude Opus 5.5 & ground coordinates only + numeric definitions & 600 & 99.0 & 99.0 & 99.0 & +0.0 [$-$2, +2] \\
Claude Opus 5.5 & second view (own-camera cue) & 465 & 86.5 & 86.9 & 83.8 & +3.1 [$-$6, +12] \\
\addlinespace
GPT-5.5 & image & 600 & 78.5 & 96.0 & 61.0 & +35.0 [+29, +41] \\
GPT-5.5 & image + CoT & 600 & 78.3 & 94.7 & 62.0 & +32.7 [+26, +39] \\
GPT-5.5 & image + stated thresholds & 600 & 80.8 & 97.3 & 64.3 & +33.0 [+27, +39] \\
GPT-5.5 & image + numeric definitions only & 600 & 78.7 & 94.7 & 62.7 & +32.0 [+25, +38] \\
GPT-5.5 & image + reject bands only & 600 & 80.5 & 97.7 & 63.3 & +34.3 [+28, +40] \\
GPT-5.5 & image + numeric definitions + bands & 600 & 77.7 & 96.7 & 58.7 & +38.0 [+32, +44] \\
GPT-5.5 & image + verbal definitions & 600 & 77.0 & 93.7 & 60.3 & +33.3 [+27, +39] \\
GPT-5.5 & text only & 600 & 50.3 & 50.7 & 50.0 & +0.7 [$-$7, +9] \\
GPT-5.5 & image + exact pixel measurements (text) & 600 & 79.7 & 95.0 & 64.3 & +30.7 [+25, +37] \\
GPT-5.5 & pixel measurements, no image & 600 & 81.5 & 91.0 & 72.0 & +19.0 [+13, +25] \\
GPT-5.5 & image + pixel measurements + camera depth (text) & 600 & 85.8 & 97.0 & 74.7 & +22.3 [+17, +27] \\
GPT-5.5 & image + depth map (no legend) & 600 & 78.7 & 93.7 & 63.7 & +30.0 [+24, +36] \\
GPT-5.5 & image + depth map (metric legend) & 600 & 81.0 & 94.3 & 67.7 & +26.7 [+21, +32] \\
GPT-5.5 & image + camera depth (text) & 600 & 88.8 & 96.7 & 81.0 & +15.7 [+11, +21] \\
GPT-5.5 & image + top-down map & 600 & 93.0 & 95.7 & 90.3 & +5.3 [+1, +9] \\
GPT-5.5 & image + top-down map, no labels & 600 & 92.7 & 96.3 & 89.0 & +7.3 [+3, +12] \\
GPT-5.5 & image + ground coordinates (text) & 600 & 93.8 & 96.0 & 91.7 & +4.3 [+1, +8] \\
GPT-5.5 & ground coordinates only (no image) & 600 & 92.5 & 95.7 & 89.3 & +6.3 [+2, +11] \\
GPT-5.5 & ground coordinates only + verbal definitions & 600 & 95.5 & 96.3 & 94.7 & +1.7 [$-$1, +5] \\
GPT-5.5 & ground coordinates only + numeric definitions & 600 & 100.0 & 100.0 & 100.0 & +0.0 [+0, +0] \\
GPT-5.5 & second view (own-camera cue) & 465 & 83.2 & 87.7 & 57.4 & +30.3 [+19, +42] \\
\addlinespace
Gemini 3.6 Flash & image & 600 & 81.7 & 94.0 & 69.3 & +24.7 [+19, +30] \\
Gemini 3.6 Flash & image + CoT & 600 & 80.7 & 93.3 & 68.0 & +25.3 [+19, +31] \\
Gemini 3.6 Flash & image + stated thresholds & 600 & 84.0 & 95.0 & 73.0 & +22.0 [+17, +27] \\
Gemini 3.6 Flash & image + numeric definitions only & 600 & 85.2 & 95.0 & 75.3 & +19.7 [+15, +25] \\
Gemini 3.6 Flash & image + reject bands only & 600 & 84.5 & 95.3 & 73.7 & +21.7 [+16, +27] \\
Gemini 3.6 Flash & image + numeric definitions + bands & 600 & 83.8 & 95.7 & 72.0 & +23.7 [+18, +30] \\
Gemini 3.6 Flash & image + verbal definitions & 600 & 81.5 & 93.7 & 69.3 & +24.3 [+19, +30] \\
Gemini 3.6 Flash & text only & 600 & 46.5 & 47.3 & 45.7 & +1.7 [$-$7, +9] \\
Gemini 3.6 Flash & image + exact pixel measurements (text) & 600 & 80.2 & 92.3 & 68.0 & +24.3 [+18, +30] \\
Gemini 3.6 Flash & pixel measurements, no image & 600 & 79.7 & 87.7 & 71.7 & +16.0 [+10, +23] \\
Gemini 3.6 Flash & image + pixel measurements + camera depth (text) & 600 & 82.2 & 93.3 & 71.0 & +22.3 [+17, +28] \\
Gemini 3.6 Flash & image + depth map (no legend) & 600 & 79.5 & 90.0 & 69.0 & +21.0 [+15, +27] \\
Gemini 3.6 Flash & image + depth map (metric legend) & 600 & 79.3 & 90.7 & 68.0 & +22.7 [+17, +29] \\
Gemini 3.6 Flash & image + camera depth (text) & 600 & 84.3 & 93.3 & 75.3 & +18.0 [+12, +23] \\
Gemini 3.6 Flash & image + top-down map & 600 & 92.0 & 94.3 & 89.7 & +4.7 [+0, +9] \\
Gemini 3.6 Flash & image + top-down map, no labels & 600 & 91.5 & 96.0 & 87.0 & +9.0 [+4, +13] \\
Gemini 3.6 Flash & image + ground coordinates (text) & 600 & 93.2 & 95.7 & 90.7 & +5.0 [+1, +9] \\
Gemini 3.6 Flash & ground coordinates only (no image) & 600 & 92.8 & 94.7 & 91.0 & +3.7 [$-$0, +8] \\
Gemini 3.6 Flash & ground coordinates only + verbal definitions & 600 & 95.0 & 95.3 & 94.7 & +0.7 [$-$3, +4] \\
Gemini 3.6 Flash & ground coordinates only + numeric definitions & 600 & 99.5 & 99.3 & 99.7 & $-$0.3 [$-$2, +1] \\
Gemini 3.6 Flash & second view (own-camera cue) & 465 & 82.6 & 83.9 & 75.0 & +8.9 [$-$2, +20] \\
\addlinespace
Pixtral Large & image & 600 & 60.5 & 67.3 & 53.7 & +13.7 [+6, +22] \\
Pixtral Large & image + CoT & 600 & 62.5 & 73.7 & 51.3 & +22.3 [+15, +30] \\
Pixtral Large & image + verbal definitions & 600 & 54.5 & 60.3 & 48.7 & +11.7 [+4, +20] \\
Pixtral Large & text only & 600 & 50.0 & 50.7 & 49.3 & +1.3 [$-$7, +9] \\
Pixtral Large & image + exact pixel measurements (text) & 600 & 56.0 & 67.7 & 44.3 & +23.3 [+16, +31] \\
Pixtral Large & image + depth map (no legend) & 600 & 60.3 & 64.7 & 56.0 & +8.7 [+1, +16] \\
Pixtral Large & image + camera depth (text) & 600 & 63.7 & 65.3 & 62.0 & +3.3 [$-$4, +11] \\
Pixtral Large & image + top-down map & 600 & 57.8 & 62.3 & 53.3 & +9.0 [+2, +16] \\
Pixtral Large & image + ground coordinates (text) & 600 & 58.8 & 66.0 & 51.7 & +14.3 [+6, +22] \\
Pixtral Large & ground coordinates only (no image) & 600 & 63.3 & 66.0 & 60.7 & +5.3 [$-$2, +12] \\
Pixtral Large & second view (own-camera cue) & 465 & 56.6 & 58.9 & 42.6 & +16.3 [+3, +29] \\
\addlinespace
\end{longtable}}
}

\begin{table}[h]
\centering\small
\caption{Gap (points) on items whose referenced objects were placed by a sampler construct versus the natural remainder, pooled over the three tiers (number of cue-wrong items in parentheses).}
\label{tab:natcon}
\resizebox{\textwidth}{!}{\begin{tabular}{lrrrrrrrr}
\toprule
 & \multicolumn{2}{c}{taller} & \multicolumn{2}{c}{touching} & \multicolumn{2}{c}{between} & \multicolumn{2}{c}{fits} \\
\cmidrule(lr){2-3}\cmidrule(lr){4-5}\cmidrule(lr){6-7}\cmidrule(lr){8-9}
model & natural & constructed & natural & constructed & natural & constructed & natural & constructed \\
\midrule
Claude Opus 5.5 & $-$3 (35) & +3 (145) & +0 (13) & +31 (167) & +57 (8) & +41 (172) & +23 (12) & +27 (168) \\
GPT-5.5 & +20 (35) & +21 (145) & +69 (13) & +38 (167) & +57 (8) & +43 (172) & +78 (12) & +62 (168) \\
Gemini 3.6 Flash & +11 (35) & +17 (145) & +0 (13) & +22 (167) & +51 (8) & +42 (172) & +23 (12) & +46 (168) \\
\bottomrule
\end{tabular}
}
\end{table}

\begin{table}[h]
\centering\small
\caption{Three runs of each frontier model on the medium image condition, with accuracy (\%), gap (points), the standard deviation of the gap and item-level agreement (\%) between runs.}\label{tab:repeats}
\begin{tabular}{lrllrr}
\toprule
model & repeats & accuracies & gaps & gap s.d. & item-level agreement \\
\midrule
Claude Opus 5.5 & 3 & 85.7/86.5/86.2 & +21.3/+19.7/+19.0 & 1.2 & 92.8 \\
GPT-5.5 & 3 & 78.5/77.7/78.8 & +35.0/+36.0/+34.3 & 0.8 & 89.5 \\
Gemini 3.6 Flash & 3 & 81.7/82.0/80.8 & +24.7/+22.7/+21.7 & 1.5 & 88.8 \\
\bottomrule
\end{tabular}

\end{table}

\begin{table}[h]
\centering\small
\caption{Accuracy (\%) and gap (points) at low and high reasoning effort on the medium image condition. The middle columns are the high-effort run with a 16{,}384-token budget, which gives the high-effort results in the text. The last two columns are a second high-effort run with a 4{,}096-token budget, in which truncated responses were sampled again. The lower block gives the \family{fits} gap and the accuracy on the cue-wrong \family{fits} items at low and high effort (16{,}384-token run), with the 95\% interval of the change in the gap.}\label{tab:effort}
\resizebox{\textwidth}{!}{\begin{tabular}{lrlrlrl}
\toprule
 & \multicolumn{2}{c}{low effort} & \multicolumn{2}{c}{high effort (16k budget, single run)} & \multicolumn{2}{c}{high effort (4k budget + retries)} \\
\cmidrule(lr){2-3}\cmidrule(lr){4-5}\cmidrule(lr){6-7}
model & acc & gap [95\% CI] & acc & gap [95\% CI] & acc & gap [95\% CI] \\
\midrule
Claude Opus 5.5 & 85.7 & +21.3 [+16, +27] & 88.0 & +16.7 [+12, +22] & 87.8 & +15.0 [+10, +20] \\
GPT-5.5 & 78.5 & +35.0 [+29, +41] & 83.7 & +25.3 [+20, +31] & 82.3 & +30.0 [+24, +36] \\
Gemini 3.6 Flash & 81.7 & +24.7 [+19, +30] & 83.0 & +17.3 [+12, +23] & 80.3 & +23.3 [+17, +29] \\
\bottomrule
\end{tabular}
}

\vspace{0.9em}
\begin{tabular}{l rrl rr}
\toprule
 & \multicolumn{3}{c}{\family{fits} gap (points)} & \multicolumn{2}{c}{cue-wrong \family{fits} accuracy (\%)} \\
\cmidrule(lr){2-4}\cmidrule(lr){5-6}
model & low effort & high effort & change, 95\% CI & low effort & high effort \\
\midrule
Claude Opus 5.5 & \n{r10.eff.Opus.fits.low} & \n{r10.eff.Opus.fits.high} & \n{r10.eff.Opus.fits.dci} & \n{r10.eff.Opus.fitsdisc.low} & \n{r10.eff.Opus.fitsdisc.high} \\
GPT-5.5 & \n{r10.eff.Gpt.fits.low} & \n{r10.eff.Gpt.fits.high} & \n{r10.eff.Gpt.fits.dci} & \n{r10.eff.Gpt.fitsdisc.low} & \n{r10.eff.Gpt.fitsdisc.high} \\
Gemini 3.6 Flash & \n{r10.eff.Gemini.fits.low} & \n{r10.eff.Gemini.fits.high} & \n{r10.eff.Gemini.fits.dci} & \n{r10.eff.Gemini.fitsdisc.low} & \n{r10.eff.Gemini.fitsdisc.high} \\
\bottomrule
\end{tabular}

\end{table}

\begin{table}[h]
\centering\small
\caption{Accuracy (\%) and gap (points) on the same medium scenes and statements with a plain and a checkerboard ground (image condition).}
\label{tab:checker}
\begin{tabular}{lrrrl}
\toprule
 & \multicolumn{2}{c}{plain ground} & \multicolumn{2}{c}{checkerboard ground} \\
\cmidrule(lr){2-3}\cmidrule(lr){4-5}
model & acc & gap & acc & gap [95\% CI] \\
\midrule
Claude Opus 5.5 & 85.7 & +21.3 & 86.8 & +18.3 [+13, +24] \\
GPT-5.5 & 78.5 & +35.0 & 79.2 & +31.0 [+25, +37] \\
Gemini 3.6 Flash & 81.7 & +24.7 & 81.8 & +23.7 [+18, +30] \\
\bottomrule
\end{tabular}

\end{table}

\begin{table}[h]
\centering\scriptsize
\caption{Alternative cues on the medium tier. Geometry-faithful cues are those that respect perspective. A base point is a ground-contact pixel, top y is the image row of the visible top, x between tests the middle object's image column and bbox is the visible box. The column ``agree.'' gives each member cue's agreement (\%) with the un-negated truth on the split's items (the balanced cue is 50 by construction). Model columns are Firth-logit coefficients of the model's answer on truth and on one composite per cue set (the mean of the members' $\pm1$ codes). VIF is the variance inflation factor in the Opus 5.5 fit. The relation \family{closer} is omitted because its fit is saturated.}
\label{tab:cuebank}
\resizebox{\textwidth}{!}{\begin{tabular}{llrrrrrrrrrrr}
\toprule
relation & regressor (members) & agree. & VIF & \rotatebox{60}{Claude Opus 5.5} & \rotatebox{60}{GPT-5.5} & \rotatebox{60}{Gemini 3.6 Flash} & \rotatebox{60}{Grok 4.7} & \rotatebox{60}{Qwen3-VL-235B} & \rotatebox{60}{Pixtral Large} & \rotatebox{60}{Nova 2 Pro} & \rotatebox{60}{Llama 4 Maverick} & \rotatebox{60}{Gemma 3 27B} \\
\midrule
taller & truth & -- & 1.4 & +4.0 & +2.6 & +4.1 & +1.8 & +1.0 & +0.6 & +0.8 & +0.5 & +0.1 \\
 & pixel cues (pixel height, area) & 50/48 & 1.1 & +0.9 & +1.3 & +1.9 & +1.3 & +1.0 & +0.3 & +0.9 & +0.4 & +0.5 \\
 & geometry-faithful (depth-corrected height, top y) & 69/71 & 1.5 & +0.3 & +1.1 & +0.6 & +1.0 & +1.2 & +0.6 & +0.9 & $-$0.5 & +0.4 \\
\addlinespace
touching & truth & -- & 2.2 & +2.2 & +0.7 & +0.9 & +0.6 & +0.7 & +0.2 & +0.9 & +0.3 & $-$0.8 \\
 & silhouette cues (mask distance, bbox overlap) & 50/50 & 1.3 & +1.3 & +1.4 & +0.5 & +1.5 & +1.0 & +0.6 & +1.0 & +1.1 & +0.2 \\
 & geometry-faithful (base points close) & 82 & 2.5 & $-$0.2 & +0.3 & +0.6 & +0.2 & +0.4 & +0.2 & +0.3 & +0.3 & +1.0 \\
\addlinespace
between & truth & -- & 3.9 & +3.3 & +2.0 & +2.2 & +1.7 & +0.6 & +0.2 & +0.4 & +0.1 & +0.5 \\
 & image-plane collinearity (centroid collinear, bbox centre collinear, x between) & 50/63/64 & 1.3 & +3.5 & +2.8 & +2.7 & +2.4 & +0.9 & +0.0 & +0.7 & +0.1 & +0.3 \\
 & geometry-faithful (base collinear) & 92 & 4.4 & +0.3 & $-$0.2 & $-$0.5 & $-$0.5 & $-$0.3 & +0.1 & $-$0.3 & +0.3 & $-$0.4 \\
\addlinespace
fits & truth & -- & 1.6 & +2.3 & +1.0 & +1.1 & +0.7 & +0.3 & +0.6 & $-$0.2 & +0.0 & +0.3 \\
 & pixel cues (pixel gap vs width, base gap vs width) & 50/56 & 1.8 & +1.5 & +1.9 & +1.1 & +1.0 & +0.4 & +0.8 & +0.0 & +0.3 & +0.2 \\
 & geometry-faithful (depth-corrected) & 72 & 2.5 & +1.9 & +0.8 & +0.5 & +0.5 & $-$0.4 & $-$0.4 & +0.5 & $-$0.2 & $-$0.2 \\
\addlinespace
\bottomrule
\end{tabular}
}
\end{table}

\FloatBarrier
\section{Checks of the rules, renderer, solvers and adjustment}
\label{app:audit}
\small

\paragraph{Formal predicates.} Objects are convex primitives resting on the ground plane $z=0$ with footprint centre $p_i \in \mathbb{R}^2$, size $s_i$ (cube side or diameter), height $h_i$ and, for cubes, yaw $\psi_i$. Let $d(\cdot,\cdot)$ be the exact minimum Euclidean distance between two solids (0 if they meet), $d_F$ the minimum distance between two footprints in the ground plane, and $z(\cdot)$ the depth of a point along the camera's viewing axis. With margins as in Table~\ref{tab:families}, the propositions are
\[\resizebox{\textwidth}{!}{$\begin{aligned}
\text{closer}(A,B) &: z(c_A) < z(c_B),\ c_i = (p_i, h_i/2), &\text{reject if } \max(z_A,z_B)/\min(z_A,z_B) < 1.15,\\
\text{taller}(A,B) &: h_A > h_B, &\text{reject if } \max(h)/\min(h) < 1.25,\\
\text{touching}(A,B) &: d(A,B) < 0.02, &\text{reject if } 0.02 \le d \le 0.5\min(s_A,h_A,s_B,h_B),\\
\text{between}(C;A,B) &: \max(\angle(p_C{-}p_A, p_B{-}p_A),\ \angle(p_C{-}p_B, p_A{-}p_B)) \le 10^\circ, &\text{reject if } 10^\circ < \cdot < 30^\circ,\\
\text{fits}(A;B,C) &: d_F(B,C) / s_A \ge 1.2 \text{ (true)},\ \le 0.8 \text{ (false)}, &\text{reject otherwise, if } d_F < 0.1,\\
&&\text{or if any other footprint is within 0.25 of the gap segment.}
\end{aligned}$}\]
The \family{between} rule is angular. A point $0.2$ units off the line but next to an endpoint is not between (a case in the test suite), and a point beyond an endpoint on the line is not between. The \family{fits} rule compares the minimum footprint gap with the narrowest width. It is a clearance test at the gap, not a general collision-free path. The 2-D cues apply the same rules to the visible masks of the main view, with pixel area (closer), visible height (taller), mask distance $\le 3$\,px (touching), the angular rule on visible-mask centroids at $20^\circ$ (between), and pixel gap $\ge$ pixel width (fits). ``Overlapping'' means a mask distance of 0 on the analytic instance mask, where each pixel belongs to the nearest surface.

\paragraph{Hand-checked cases.} The test suite covers spheres, cubes, rotated cubes, cylinders and cones at known distances. It covers unequal tangent spheres, which touch at horizontal centre separation $2\sqrt{r_1 r_2}$ while their footprint discs overlap, so footprint overlap is not contact. It also covers cones touching at their bases, the angular \family{between} cases above, a third object inside a \family{fits} gap (rejected), the round trip of the image-to-ground homography and exact recovery of every label from the coordinate, homography and surface-distance serialisations on a sample split.

\paragraph{Renderer consistency.} The 2-D statistics come from an analytic ray caster, while the models see Blender renders of tessellated meshes (96$\times$48 sphere, 128-sided cylinders and cones). We re-rendered all \n{data.medium.scenes} scenes of the first medium split (seed 7) with flat emission colours, no lights and a near-zero pixel filter, so the silhouettes of the actual meshes classify exactly. Per-object visible masks agree with the analytic masks to a median IoU of \n{render.iou} (1st percentile \n{render.ioup1}). Recomputing the balanced cues on the Blender masks flips \n{render.nflip} of 600 cue decisions, all at the 3-pixel touching rule, the pixel-width rule of \family{fits} or the pixel-height rule of \family{taller}. Relabelling those items changes the frontier models' medium gaps to \n{Opus.render.gap}, \n{Gpt.render.gap} and \n{Gemini.render.gap}, against \n{Opus.medium.image.gap}, \n{Gpt.medium.image.gap} and \n{Gemini.medium.image.gap}.

\paragraph{Solvers.} The solvers of Table~\ref{tab:solvers} regenerate each input's serialisation with the evaluation code. On 80 sampled items per input the regenerated text is identical to the prompt sent to the model. The coordinate serialisation (two decimals) recovers all labels, including contact under rounding. The pixel-plus-depth serialisation recovers \family{closer} and \family{between} exactly. With intrinsics from the field of view and a central principal point, each ground-contact pixel and its camera depth give the contact point in camera coordinates, and the angles of \family{between} are invariant to the camera pose. This serialisation lacks object sizes, so \family{fits}, \family{touching} and \family{taller} need size estimates from pixel extents.

\begin{table}[h]
\centering\small
\caption{Difficulty-adjusted gap (points, image condition), the average marginal effect of the cue being right in a Firth logit with difficulty and construct covariates and interactions. CIs resample scenes 2{,}000 times. The tertile columns give the raw and conditional adjusted gap on the items of each rank-based within-relation tertile of the margin, T1 the smallest (\family{touching} excluded).}
\label{tab:matched}
\resizebox{\textwidth}{!}{\begin{tabular}{llrlll}
\toprule
model & tier & raw gap & adjusted gap [95\% CI] & raw gap by margin tertile (T1/T2/T3) & adjusted gap by margin tertile \\
\midrule
Claude Opus 5.5 & easy & +19.7 & +10.3 [+5, +15] & +29 / +4 / $-$4 & +19 / +2 / $-$2 \\
Claude Opus 5.5 & medium & +21.3 & +14.6 [+11, +18] & +33 / +10 / +0 & +28 / +10 / +0 \\
Claude Opus 5.5 & hard & +17.7 & +12.3 [+7, +17] & +18 / +12 / +12 & +16 / +12 / +3 \\
GPT-5.5 & easy & +37.0 & +28.6 [+21, +35] & +40 / +34 / +3 & +29 / +27 / +14 \\
GPT-5.5 & medium & +35.0 & +27.5 [+22, +33] & +42 / +22 / +9 & +31 / +27 / +14 \\
GPT-5.5 & hard & +28.7 & +23.0 [+17, +29] & +37 / +27 / +22 & +28 / +27 / +17 \\
Gemini 3.6 Flash & easy & +24.3 & +12.7 [+8, +18] & +36 / +12 / +2 & +23 / +8 / $-$0 \\
Gemini 3.6 Flash & medium & +24.7 & +16.7 [+12, +22] & +39 / +19 / +3 & +29 / +19 / +4 \\
Gemini 3.6 Flash & hard & +23.7 & +16.4 [+12, +21] & +34 / +22 / +8 & +29 / +19 / +5 \\
\bottomrule
\end{tabular}
}
\end{table}

\begin{table}[h]
\centering\small
\caption{Tool-access intervention (Opus 5.5, medium tier). $\Delta$gap is the paired change against the low-effort image condition (2{,}000 scene-bootstrap draws). The high-effort row is the 16{,}384-token run. ``\% used'' is the share of episodes with at least one tool call. The by-episode columns split accuracy / gap by whether tools were used, a self-selected split.}
\label{tab:agentic}
\resizebox{\textwidth}{!}{\begin{tabular}{l rlr rrrr rr rrrrr}
\toprule
 & & & & \multicolumn{4}{c}{tool use} & \multicolumn{2}{c}{acc / gap by episode} & \multicolumn{5}{c}{gap by relation} \\
\cmidrule(lr){5-8}\cmidrule(lr){9-10}\cmidrule(lr){11-15}
input & acc & gap [95\% CI] & $\Delta$gap vs.\ low [CI] & calls/item & \% used & \% python & \% zoom & tools used & not used & closer & taller & touching & between & fits \\
\midrule
image, no tools (low effort) & 85.7 & +21.3 [+16, +27] & -- & -- & -- & -- & -- & -- & -- & +0 & +3 & +27 & +45 & +32 \\
image, no tools (high effort, 16k) & 88.0 & +16.7 [+12, +22] & $-$4.7 [$-$9, $-$1] & -- & -- & -- & -- & -- & -- & +0 & +2 & +30 & +42 & +10 \\
tools available & 86.3 & +16.0 [+11, +21] & $-$5.3 [$-$9, $-$2] & 0.5 & 50 & 22 & 30 & 77 / +15 & 96 / +11 & +0 & +0 & +22 & +40 & +18 \\
tools required first & 86.0 & +17.3 [+12, +23] & $-$4.0 [$-$8, $-$1] & 0.6 & 55 & 25 & 32 & 80 / +18 & 93 / +13 & +0 & +2 & +22 & +42 & +22 \\
\bottomrule
\end{tabular}
}
\end{table}

\begin{table}[h]
\centering\small
\caption{Second-view accuracy and \sgap{} on the same view-invariant medium items under two groupings, by the first camera's cue (the balanced split, 231 cue-right / 234 cue-wrong) and by the second camera's own cue (397 / 68, not balanced). $\Delta$ is the paired accuracy change from view 1 to view 2. CIs resample scenes (2{,}000).}
\label{tab:twoviewgap}
\resizebox{\textwidth}{!}{\begin{tabular}{l rrr rr rr}
\toprule
 & \multicolumn{3}{c}{accuracy} & \multicolumn{2}{c}{grouped by view-1 cue (231 / 234)} & \multicolumn{2}{c}{grouped by view-2 cue (397 / 68)} \\
\cmidrule(lr){2-4}\cmidrule(lr){5-6}\cmidrule(lr){7-8}
model & view 1 & view 2 & $\Delta$ [95\% CI] & gap, view 1 & gap, view 2 & gap, view 1 & gap, view 2 \\
\midrule
Claude Opus 5.5 & 81.9 & 86.5 & +4.5 [+1, +8] & +26.4 [+20, +33] & +11.4 [+6, +17] & $-$7.4 [$-$16, +1] & +3.1 [$-$6, +12] \\
GPT-5.5 & 72.7 & 83.2 & +10.5 [+6, +15] & +44.0 [+37, +51] & +11.8 [+5, +18] & +2.5 [$-$9, +14] & +30.3 [+19, +42] \\
Gemini 3.6 Flash & 76.8 & 82.6 & +5.8 [+2, +10] & +30.7 [+23, +38] & +12.2 [+5, +19] & $-$8.3 [$-$17, +2] & +8.9 [$-$2, +20] \\
\bottomrule
\end{tabular}
}
\end{table}

\begin{table}[h]
\centering\small
\caption{The model's drawn map fed back as text (no image, numeric definitions), \family{fits} and \family{between} items of the medium tier. ``Depth-corrected'' stretches the map's depth axis by the inverse of the median compression estimated on the other half of the scenes. $\Delta$gap is the paired change from the drawn map to the depth-corrected map (scene bootstrap). The table pools the two relations, and Section~\ref{sec:res-maps} gives the change for each. The last column is the share of items on which the model's answer equals a solver reading the same layout.}
\label{tab:ownmap}
\resizebox{\textwidth}{!}{\begin{tabular}{l r l l l l r}
\toprule
model & items & image + numeric definitions (acc / gap) & drawn map & drawn map, depth-corrected & $\Delta$gap [95\% CI] & answers = solver (\%) \\
\midrule
Claude Opus 5.5 & 240 & 87.1 / +20.8 & 77.1 / +35.8 & 79.2 / +28.3 & $-$7.5 [$-$12, $-$3] & 100 \\
GPT-5.5 & 240 & 72.1 / +52.5 & 67.9 / +50.8 & 70.4 / +44.2 & $-$6.7 [$-$15, +2] & 100 \\
Gemini 3.6 Flash & 240 & 77.9 / +37.5 & 67.5 / +30.0 & 73.3 / +25.0 & $-$5.0 [$-$14, +4] & 100 \\
\bottomrule
\end{tabular}
}
\end{table}

\begin{table}[h]
\centering\small
\caption{Common support for the difficulty adjustment (medium, image). The column ``\% off'' gives the share of items with a propensity of being cue-right outside $[0.05, 0.95]$ under a Firth propensity model on the difficulty covariates and proposition truth. The gaps are raw, raw on items with propensity in $[0.1, 0.9]$, and inverse-propensity weighted (weights trimmed at $[0.05, 0.95]$). The adjusted gaps come from a within-relation Firth outcome model without the interaction of cue status with margin, with construct membership as a binary flag or as construct type and scene-bootstrap CIs.}
\label{tab:overlap}
\resizebox{\textwidth}{!}{\begin{tabular}{ll rr rrr rr}
\toprule
 & & \multicolumn{2}{c}{common support} & \multicolumn{3}{c}{gap (points)} & \multicolumn{2}{c}{adjusted gap, within relation} \\
\cmidrule(lr){3-4}\cmidrule(lr){5-7}\cmidrule(lr){8-9}
relation & model & \% off & $n$ supported (cue right/wrong) & raw & supported & IPW & binary construct [CI] & construct type [CI] \\
\midrule
taller & Claude Opus 5.5 & 9 & 102 (43/59) & +3 & +3 & +3 & +3 [$-$6, +13] & +5 [$-$2, +14] \\
taller & GPT-5.5 & 9 & 102 (43/59) & +17 & +16 & +13 & +10 [$-$1, +27] & +15 [+1, +33] \\
taller & Gemini 3.6 Flash & 9 & 102 (43/59) & +17 & +17 & +15 & +14 [+4, +29] & +15 [+4, +33] \\
\addlinespace
touching & Claude Opus 5.5 & 20 & 87 (42/45) & +27 & +29 & +31 & +32 [+15, +44] & +33 [+18, +48] \\
touching & GPT-5.5 & 20 & 87 (42/45) & +50 & +46 & +51 & +39 [+20, +57] & +39 [+21, +59] \\
touching & Gemini 3.6 Flash & 20 & 87 (42/45) & +18 & +20 & +16 & +19 [+3, +34] & +19 [+2, +40] \\
\addlinespace
between & Claude Opus 5.5 & 4 & 103 (51/52) & +45 & +42 & +34 & +17 [$-$4, +25] & +13 [$-$8, +35] \\
between & GPT-5.5 & 4 & 103 (51/52) & +43 & +40 & +31 & +8 [$-$8, +25] & +7 [$-$15, +23] \\
between & Gemini 3.6 Flash & 4 & 103 (51/52) & +43 & +42 & +33 & +11 [$-$4, +26] & +10 [$-$7, +33] \\
\addlinespace
fits & Claude Opus 5.5 & 26 & 70 (31/39) & +32 & +38 & +26 & +28 [+17, +40] & +29 [+18, +46] \\
fits & GPT-5.5 & 26 & 70 (31/39) & +65 & +66 & +59 & +63 [+51, +76] & +58 [+42, +76] \\
fits & Gemini 3.6 Flash & 26 & 70 (31/39) & +45 & +46 & +43 & +41 [+26, +58] & +47 [+32, +61] \\
\bottomrule
\end{tabular}
}
\end{table}

\begin{table}[h]
\centering\small
\caption{Per-relation gap (points, image condition), raw / adjusted under the shared-slope model [95\% CI] / adjusted under the within-relation model (Appendix~\ref{app:details}). This model keeps the interaction of cue status with margin, which Appendix Table~\ref{tab:overlap} drops, so the two tables differ by a few points.}
\label{tab:familyadj}
\resizebox{\textwidth}{!}{\begin{tabular}{lllllll}
\toprule
model & tier & closer: raw / shared [CI] / within & taller: raw / shared [CI] / within & touching: raw / shared [CI] / within & between: raw / shared [CI] / within & fits: raw / shared [CI] / within \\
\midrule
Claude Opus 5.5 & easy & $-$2 / $-$2 [$-$7, +0] / $-$3 & +5 / $-$0 [$-$9, +7] / +0 & +35 / +27 [+12, +40] / +14 & +42 / +20 [+8, +32] / +12 & +18 / +7 [$-$8, +18] / +7 \\
Claude Opus 5.5 & medium & +0 / +0 [$-$1, +3] / $-$0 & +3 / +1 [$-$4, +5] / +3 & +27 / +23 [+10, +35] / +27 & +45 / +25 [+16, +34] / +23 & +32 / +24 [+14, +34] / +27 \\
Claude Opus 5.5 & hard & +0 / $-$0 [$-$1, +0] / +0 & $-$3 / $-$6 [$-$14, $-$0] / $-$5 & +25 / +19 [+9, +31] / +24 & +37 / +22 [+9, +35] / +22 & +30 / +26 [+14, +40] / +38 \\
GPT-5.5 & easy & +0 / $-$0 [$-$1, +0] / +0 & +33 / +25 [+16, +35] / +24 & +50 / +49 [+34, +64] / +49 & +47 / +35 [+19, +49] / +21 & +55 / +34 [+13, +51] / +41 \\
GPT-5.5 & medium & +0 / $-$0 [$-$1, +0] / $-$0 & +17 / +11 [+1, +21] / +11 & +50 / +41 [+25, +56] / +38 & +43 / +29 [+12, +44] / +10 & +65 / +56 [+42, +69] / +61 \\
GPT-5.5 & hard & +0 / $-$0 [$-$6, +4] / +2 & +13 / +10 [+3, +17] / +15 & +22 / +19 [+6, +34] / +20 & +40 / +30 [+15, +45] / +25 & +68 / +57 [+40, +72] / +68 \\
Gemini 3.6 Flash & easy & $-$2 / $-$2 [$-$7, +0] / $-$3 & +23 / +16 [+8, +24] / +17 & +28 / +22 [+6, +36] / +2 & +43 / +20 [+8, +32] / +14 & +28 / +9 [$-$6, +22] / +4 \\
Gemini 3.6 Flash & medium & +0 / $-$0 [$-$1, +0] / $-$0 & +17 / +11 [+5, +18] / +14 & +18 / +15 [+1, +29] / +18 & +43 / +28 [+15, +41] / +17 & +45 / +29 [+15, +45] / +40 \\
Gemini 3.6 Flash & hard & +0 / $-$0 [$-$1, +0] / +0 & +8 / +5 [+1, +11] / +22 & +13 / +11 [+0, +23] / +12 & +38 / +22 [+6, +35] / +26 & +58 / +44 [+32, +56] / +48 \\
\bottomrule
\end{tabular}
}
\end{table}

\begin{table}[h]
\centering\small
\caption{Accuracy (\%) / gap (points) on the first medium split (seed 7) and on a second medium split generated from seed 11 with the same settings.}
\label{tab:seed}
\begin{tabular}{l ll ll ll}
\toprule
 & \multicolumn{2}{c}{image} & \multicolumn{2}{c}{ground coordinates only} & \multicolumn{2}{c}{image + top-down map} \\
\cmidrule(lr){2-3}\cmidrule(lr){4-5}\cmidrule(lr){6-7}
model & seed 7 & seed 11 & seed 7 & seed 11 & seed 7 & seed 11 \\
\midrule
Claude Opus 5.5 & 85.7 / +21.3 & 83.8 / +19.7 & 95.5 / +6.3 & 95.7 / +2.7 & 93.0 / +4.0 & 93.7 / +4.0 \\
GPT-5.5 & 78.5 / +35.0 & 76.8 / +35.0 & 92.5 / +6.3 & 91.8 / +2.3 & 93.0 / +5.3 & 93.5 / +4.3 \\
Gemini 3.6 Flash & 81.7 / +24.7 & 79.3 / +22.7 & 92.8 / +3.7 & 92.2 / +3.0 & 92.0 / +4.7 & 92.0 / +6.0 \\
\bottomrule
\end{tabular}

\end{table}

\begin{table}[h]
\centering\small
\caption{Item-level correspondence between the direct image answer and the answer re-derived from the model's drawn map (medium tier, \family{taller} excluded because heights are not mapped). The Mantel--Haenszel odds ratio pools the 2$\times$2 tables within relation $\times$ cue status strata, and $\phi$ is the correlation between the two correctness indicators. ``\% in band'' is the share of re-derived relations that fall inside a reject band, which count as wrong. The map gap is given on parsed scenes and with unparsed scenes counted as wrong.}
\label{tab:mapitems}
\resizebox{\textwidth}{!}{\begin{tabular}{l r r rrrr rrr r rr}
\toprule
 & & & \multicolumn{4}{c}{direct / map-derived correct} & & & & & \multicolumn{2}{c}{map gap} \\
\cmidrule(lr){4-7}\cmidrule(lr){12-13}
model & scenes parsed & items & both & direct only & map only & neither & agree \% & $\phi$ & MH odds ratio & \% in band & parsed & parse-incl. \\
\midrule
Claude Opus 5.5 & 304/304 & 480 & 305 & 91 & 30 & 54 & 75 & 0.34 & 2.4 & 23 & +20 & +20 \\
GPT-5.5 & 304/304 & 480 & 245 & 118 & 27 & 90 & 70 & 0.38 & 2.1 & 31 & +38 & +38 \\
Gemini 3.6 Flash & 304/304 & 480 & 260 & 120 & 21 & 79 & 71 & 0.39 & 3.9 & 28 & +30 & +30 \\
Grok 4.7 & 224/304 & 353 & 188 & 75 & 31 & 59 & 70 & 0.33 & 2.3 & 25 & +18 & +16 \\
\bottomrule
\end{tabular}
}
\end{table}

\begin{table}[h]
\centering\small
\caption{\family{between} accuracy (\%) by deviation of the third object from the line, medium tier, with item counts. The 30--60$^\circ$ bands hold all 30 cue-wrong false items and 4 cue-right ones. The $\ge60^\circ$ band holds the other 26 cue-right false items.}
\label{tab:betweenbands}
\resizebox{0.85\textwidth}{!}{\begin{tabular}{llrrrr}
\toprule
model & input & $\le10^\circ$ (true) & 30--40$^\circ$ & 40--60$^\circ$ & $\ge60^\circ$ \\
\midrule
Claude Opus 5.5 & image & 100 (60) & 18 (17) & 24 (17) & 100 (26) \\
Claude Opus 5.5 & verbal definitions & 100 (60) & 53 (17) & 47 (17) & 100 (26) \\
Claude Opus 5.5 & stated thresholds & 98 (60) & 71 (17) & 82 (17) & 100 (26) \\
Claude Opus 5.5 & coordinates only & 100 (60) & 29 (17) & 59 (17) & 100 (26) \\
Claude Opus 5.5 & coordinates only + verbal definitions & 100 (60) & 76 (17) & 94 (17) & 100 (26) \\
Claude Opus 5.5 & coordinates + image & 100 (60) & 35 (17) & 88 (17) & 100 (26) \\
\addlinespace
GPT-5.5 & image & 98 (60) & 12 (17) & 12 (17) & 96 (26) \\
GPT-5.5 & verbal definitions & 75 (60) & 41 (17) & 41 (17) & 100 (26) \\
GPT-5.5 & stated thresholds & 88 (60) & 35 (17) & 47 (17) & 100 (26) \\
GPT-5.5 & coordinates only & 100 (60) & 29 (17) & 65 (17) & 100 (26) \\
GPT-5.5 & coordinates only + verbal definitions & 100 (60) & 100 (17) & 94 (17) & 100 (26) \\
GPT-5.5 & coordinates + image & 100 (60) & 47 (17) & 82 (17) & 100 (26) \\
\addlinespace
Gemini 3.6 Flash & image & 97 (60) & 18 (17) & 18 (17) & 100 (26) \\
Gemini 3.6 Flash & verbal definitions & 90 (60) & 41 (17) & 47 (17) & 100 (26) \\
Gemini 3.6 Flash & stated thresholds & 92 (60) & 35 (17) & 47 (17) & 100 (26) \\
Gemini 3.6 Flash & coordinates only & 98 (60) & 41 (17) & 71 (17) & 100 (26) \\
Gemini 3.6 Flash & coordinates only + verbal definitions & 97 (60) & 100 (17) & 100 (17) & 100 (26) \\
Gemini 3.6 Flash & coordinates + image & 100 (60) & 41 (17) & 76 (17) & 100 (26) \\
\addlinespace
\bottomrule
\end{tabular}
}
\end{table}

\begin{table}[h]
\centering\small
\caption{Two-view consistency on the view-invariant relations (2-D cue on every tier, models on the medium tier). Agreement is the share (\%) of items answered identically on both cameras.}
\label{tab:twoview}
\resizebox{\textwidth}{!}{\begin{tabular}{llrrrrrr}
\toprule
model & tier & $n$ & agreement & both correct & cue flips & agreement on flips & agreement on non-flips \\
\midrule
2-D cue & easy & 476 & 62.6 & 48.3 & 178 & 0.0 & 100.0 \\
2-D cue & medium & 465 & 56.1 & 45.6 & 204 & 0.0 & 100.0 \\
2-D cue & hard & 416 & 52.4 & 44.2 & 198 & 0.0 & 100.0 \\
Claude Opus 5.5 & medium & 465 & 86.9 & 77.6 & 204 & 79.9 & 92.3 \\
GPT-5.5 & medium & 465 & 74.8 & 65.4 & 204 & 57.4 & 88.5 \\
Gemini 3.6 Flash & medium & 465 & 80.0 & 69.7 & 204 & 71.1 & 87.0 \\
\bottomrule
\end{tabular}
}
\end{table}

\begin{table}[!htbp]
\centering\small
\caption{Models, the model identifiers returned by the providers and the reasoning effort. All calls used provider-default sampling, PNG images and a 4{,}096-token completion budget, except the 16{,}384-token high-effort and map runs.}
\label{tab:manifest}
\begin{tabular}{lll}
\toprule
model & model identifier & reasoning effort \\
\midrule
Claude Opus 5.5 & \texttt{claude-opus-5-5} & low, high \\
Gemini 3.6 Flash & \texttt{gemini-3.6-flash} & low, high \\
Grok 4.7 & \texttt{global.xai.grok-4.7} & provider default \\
Gemma 3 27B & \texttt{google.gemma-3-27b-it} & provider default \\
Qwen3-VL-235B & \texttt{qwen.qwen3-vl-235b-a22b} & provider default \\
Nova 2 Pro & \texttt{us.amazon.nova-2-pro-preview-20251202-v1:0} & provider default \\
Llama 4 Maverick & \texttt{us.meta.llama4-maverick-17b-instruct-v1:0} & provider default \\
Pixtral Large & \texttt{us.mistral.pixtral-large-2502-v1:0} & provider default \\
GPT-5.5 & \texttt{gpt-5.5-2026-04-24} & low, high \\
\bottomrule
\end{tabular}

\end{table}

\section{Filled prompt examples}
\label{app:filled}
\scriptsize
\paragraph{Item \texttt{medium-0008} (between, cue-wrong, statement true).} Statement: ``The orange cylinder is between the green cube and the red sphere.''
\par\noindent\textbf{pixel measurements} (inserted at \texttt{\{geometry\}}):\\\texttt{- green cube: bbox x 538-644, y 336-444 (width 107 px, height 109 px); centroid (590, 389); ground-contact point (585, 418); visible pixels 8631; hidden 0\%\newline - red sphere: bbox x 276-340, y 366-441 (width 65 px, height 76 px); centroid (307, 401); ground-contact point (316, 434); visible pixels 3485; hidden 23\%\newline - orange cylinder: bbox x 348-438, y 273-449 (width 91 px, height 177 px); centroid (396, 353); ground-contact point (393, 424); visible pixels 13038; hidden 10\%\newline - yellow cone: bbox x 302-402, y 356-526 (width 101 px, height 171 px); centroid (351, 463); ground-contact point (352, 492); visible pixels 9883; hidden 0\%\newline - blue cylinder: bbox x 429-546, y 441-612 (width 118 px, height 172 px); centroid (485, 525); ground-contact point (481, 572); visible pixels 17088; hidden 0\%\newline Silhouette distances: green cube / red sphere: 197 px; green cube / orange cylinder: 104 px; green cube / yellow cone: 155 px; green cube / blue cylinder: 37 px; red sphere / orange cylinder: 16 px; red sphere / yellow cone: 1 px; red sphere / blue cylinder: 114 px; orange cylinder / yellow cone: 1 px; orange cylinder / blue cylinder: 23 px; yellow cone / blue cylinder: 28 px}
\par\noindent\textbf{pixel measurements + camera depth} (inserted at \texttt{\{geometry\}}):\\\texttt{- green cube: bbox x 538-644, y 336-444 (width 107 px, height 109 px); centroid (590, 389); ground-contact point (585, 418); visible pixels 8631; hidden 0\%; distance from camera: centre 9.91, ground-contact point 10.12\newline - red sphere: bbox x 276-340, y 366-441 (width 65 px, height 76 px); centroid (307, 401); ground-contact point (316, 434); visible pixels 3485; hidden 23\%; distance from camera: centre 9.65, ground-contact point 9.88\newline - orange cylinder: bbox x 348-438, y 273-449 (width 91 px, height 177 px); centroid (396, 353); ground-contact point (393, 424); visible pixels 13038; hidden 10\%; distance from camera: centre 9.58, ground-contact point 10.03\newline - yellow cone: bbox x 302-402, y 356-526 (width 101 px, height 171 px); centroid (351, 463); ground-contact point (352, 492); visible pixels 9883; hidden 0\%; distance from camera: centre 8.65, ground-contact point 9.10\newline - blue cylinder: bbox x 429-546, y 441-612 (width 118 px, height 172 px); centroid (485, 525); ground-contact point (481, 572); visible pixels 17088; hidden 0\%; distance from camera: centre 7.88, ground-contact point 8.20\newline Silhouette distances: green cube / red sphere: 197 px; green cube / orange cylinder: 104 px; green cube / yellow cone: 155 px; green cube / blue cylinder: 37 px; red sphere / orange cylinder: 16 px; red sphere / yellow cone: 1 px; red sphere / blue cylinder: 114 px; orange cylinder / yellow cone: 1 px; orange cylinder / blue cylinder: 23 px; yellow cone / blue cylinder: 28 px}
\par\noindent\textbf{camera depth (text)} (inserted at \texttt{\{geometry\}}):\\\texttt{- green cube: centre at 9.91 units from the camera, ground-contact point at 10.12\newline - red sphere: centre at 9.65 units from the camera, ground-contact point at 9.88\newline - orange cylinder: centre at 9.58 units from the camera, ground-contact point at 10.03\newline - yellow cone: centre at 8.65 units from the camera, ground-contact point at 9.10\newline - blue cylinder: centre at 7.88 units from the camera, ground-contact point at 8.20}
\par\noindent\textbf{ground coordinates} (inserted at \texttt{\{geometry\}}):\\\texttt{- green cube: ground position (x=0.01, y=-0.63), side 0.76, rotated 74 deg about the vertical, height 0.76\newline - red sphere: ground position (x=1.57, y=1.89), diameter 0.80, height 0.80\newline - orange cylinder: ground position (x=1.21, y=1.12), diameter 0.91, height 1.61\newline - yellow cone: ground position (x=0.57, y=2.13), diameter 1.01, height 1.61\newline - blue cylinder: ground position (x=-1.02, y=1.86), diameter 0.94, height 1.12\newline - camera: position (x=-5.10, y=6.09, z=6.02), looking at the scene centre}
\par\noindent\textbf{visible-region centres} (inserted at \texttt{\{geometry\}}):\\\texttt{- green cube: (589.7, 389.2)\newline - red sphere: (306.6, 401.0)\newline - orange cylinder: (395.9, 352.8)\newline - yellow cone: (351.2, 462.9)\newline - blue cylinder: (485.2, 525.1)}
\par\noindent\textbf{ground-contact pixels} (inserted at \texttt{\{geometry\}}):\\\texttt{- green cube: (584.9, 417.7)\newline - red sphere: (316.0, 434.1)\newline - orange cylinder: (392.7, 423.8)\newline - yellow cone: (352.5, 491.7)\newline - blue cylinder: (481.2, 572.5)}
\par\noindent\textbf{ground-contact pixels + homography} (inserted at \texttt{\{geometry\}}):\\\texttt{- green cube: (584.9, 417.7)\newline - red sphere: (316.0, 434.1)\newline - orange cylinder: (392.7, 423.8)\newline - yellow cone: (352.5, 491.7)\newline - blue cylinder: (481.2, 572.5)\newline Homography H (rows):\newline   [-0.0184154, -0.0347219, 25.2953]\newline   [-0.0239956, 0.0356756, -2.59666]\newline   [2.04533e-19, 0.0041513, 1]}
\paragraph{Item \texttt{medium-0007} (touching, cue-wrong, statement true).} Statement: ``There is a gap between the pink cylinder and the cyan cylinder.''
\par\noindent\textbf{pixel measurements} (inserted at \texttt{\{geometry\}}):\\\texttt{- red cylinder: bbox x 493-587, y 230-312 (width 95 px, height 83 px); centroid (539, 271); ground-contact point (538, 286); visible pixels 6559; hidden 0\%\newline - pink cylinder: bbox x 287-393, y 172-375 (width 107 px, height 204 px); centroid (336, 260); ground-contact point (346, 345); visible pixels 15999; hidden 18\%\newline - purple sphere: bbox x 435-484, y 277-371 (width 50 px, height 95 px); centroid (461, 324); ground-contact point (436, 361); visible pixels 2917; hidden 59\%\newline - cyan cylinder: bbox x 355-446, y 268-455 (width 92 px, height 188 px); centroid (399, 357); ground-contact point (399, 427); visible pixels 14932; hidden 3\%\newline - blue cone: bbox x 367-436, y 428-529 (width 70 px, height 102 px); centroid (401, 490); ground-contact point (402, 504); visible pixels 4279; hidden 0\%\newline Silhouette distances: red cylinder / pink cylinder: 100 px; red cylinder / purple sphere: 19 px; red cylinder / cyan cylinder: 48 px; red cylinder / blue cone: 160 px; pink cylinder / purple sphere: 41 px; pink cylinder / cyan cylinder: 1 px; pink cylinder / blue cone: 71 px; purple sphere / cyan cylinder: 1 px; purple sphere / blue cone: 70 px; cyan cylinder / blue cone: 1 px}
\par\noindent\textbf{pixel measurements + camera depth} (inserted at \texttt{\{geometry\}}):\\\texttt{- red cylinder: bbox x 493-587, y 230-312 (width 95 px, height 83 px); centroid (539, 271); ground-contact point (538, 286); visible pixels 6559; hidden 0\%; distance from camera: centre 10.79, ground-contact point 10.92\newline - pink cylinder: bbox x 287-393, y 172-375 (width 107 px, height 204 px); centroid (336, 260); ground-contact point (346, 345); visible pixels 15999; hidden 18\%; distance from camera: centre 9.58, ground-contact point 10.14\newline - purple sphere: bbox x 435-484, y 277-371 (width 50 px, height 95 px); centroid (461, 324); ground-contact point (436, 361); visible pixels 2917; hidden 59\%; distance from camera: centre 9.64, ground-contact point 9.95\newline - cyan cylinder: bbox x 355-446, y 268-455 (width 92 px, height 188 px); centroid (399, 357); ground-contact point (399, 427); visible pixels 14932; hidden 3\%; distance from camera: centre 8.73, ground-contact point 9.23\newline - blue cone: bbox x 367-436, y 428-529 (width 70 px, height 102 px); centroid (401, 490); ground-contact point (402, 504); visible pixels 4279; hidden 0\%; distance from camera: centre 8.22, ground-contact point 8.52\newline Silhouette distances: red cylinder / pink cylinder: 100 px; red cylinder / purple sphere: 19 px; red cylinder / cyan cylinder: 48 px; red cylinder / blue cone: 160 px; pink cylinder / purple sphere: 41 px; pink cylinder / cyan cylinder: 1 px; pink cylinder / blue cone: 71 px; purple sphere / cyan cylinder: 1 px; purple sphere / blue cone: 70 px; cyan cylinder / blue cone: 1 px}
\par\noindent\textbf{camera depth (text)} (inserted at \texttt{\{geometry\}}):\\\texttt{- red cylinder: centre at 10.79 units from the camera, ground-contact point at 10.92\newline - pink cylinder: centre at 9.58 units from the camera, ground-contact point at 10.14\newline - purple sphere: centre at 9.64 units from the camera, ground-contact point at 9.95\newline - cyan cylinder: centre at 8.73 units from the camera, ground-contact point at 9.23\newline - blue cone: centre at 8.22 units from the camera, ground-contact point at 8.52}
\par\noindent\textbf{ground coordinates} (inserted at \texttt{\{geometry\}}):\\\texttt{- red cylinder: ground position (x=1.41, y=0.89), diameter 0.98, height 0.40\newline - pink cylinder: ground position (x=-0.73, y=1.76), diameter 0.97, height 1.70\newline - purple sphere: ground position (x=-0.33, y=0.91), diameter 0.92, height 0.92\newline - cyan cylinder: ground position (x=-1.29, y=0.57), diameter 0.76, height 1.51\newline - blue cone: ground position (x=-1.99, y=-0.06), diameter 0.59, height 0.90\newline - camera: position (x=-6.41, y=-3.75, z=6.37), looking at the scene centre}
\par\noindent\textbf{pairwise surface distances} (inserted at \texttt{\{geometry\}}):\\\texttt{- red cylinder / pink cylinder: 1.341\newline - red cylinder / purple sphere: 0.794\newline - red cylinder / cyan cylinder: 1.857\newline - red cylinder / blue cone: 2.748\newline - pink cylinder / purple sphere: 0.000\newline - pink cylinder / cyan cylinder: 0.457\newline - pink cylinder / blue cone: 1.435\newline - purple sphere / cyan cylinder: 0.188\newline - purple sphere / blue cone: 1.235\newline - cyan cylinder / blue cone: 0.265}

\section{Generation details}
\label{app:generation}
\small
Occlusion level bounds the scene's maximum per-object occlusion, at most 0.45 for low, within $[0.10,0.60]$ for medium and within $[0.20,0.65]$ for high. Referenced objects must be at least 50\% visible. The disc of object positions is shrunk to what the lens can frame ($\min(6,\,0.72\,d\tan(\mathrm{hfov}/2))$, with $d$ the camera distance and hfov the horizontal field of view). The sampler uses four constructs. They are an exact-tangent pair (bisection on the convex surface distance), an occluder over a contact line or an object's base (prisms preferred, with the height chosen so that the sight line is blocked), a collinear triple with short endpoints and a tall middle object or tall endpoints and a short offset object, and a near gap seen side-on with a far object slightly wider than the gap. The \family{between} thresholds are $10^\circ$ (true) and $30^\circ$ (false) with the 2-D test at $20^\circ$. The full rules and the log of hand-inspected items will accompany the code release.

\FloatBarrier
\section{Parse rates and token usage}
\label{app:cost}
Appendix Table~\ref{tab:parse} gives the parse and truncation rates, and Appendix Table~\ref{tab:cost} the calls and tokens.
\begin{table}[h]
\centering\small
\caption{Unparseable, API-error and budget-truncated response rates (\%) on the image condition, all tiers pooled.}
\label{tab:parse}
\begin{tabular}{lrrrr}
\toprule
model & responses & unparseable (\%) & API error (\%) & truncated (\%) \\
\midrule
Claude Opus 5.5 & 1800 & 0.0 & 0.0 & 0.0 \\
Claude Opus 5.5 (high effort) & 600 & 0.0 & 0.0 & 0.0 \\
GPT-5.5 & 1800 & 0.0 & 0.0 & 0.0 \\
GPT-5.5 (high effort) & 600 & 0.3 & 0.0 & 0.3 \\
Gemini 3.6 Flash & 1800 & 0.0 & 0.0 & 0.0 \\
Gemini 3.6 Flash (high effort) & 600 & 0.0 & 0.0 & 0.7 \\
Pixtral Large & 1800 & 0.0 & 0.0 & 0.0 \\
Qwen3-VL-235B & 1800 & 0.0 & 0.0 & 0.0 \\
Llama 4 Maverick & 1800 & 0.0 & 0.0 & 0.0 \\
Gemma 3 27B & 1800 & 0.0 & 0.0 & 0.0 \\
Grok 4.7 & 1800 & 2.2 & 0.0 & 2.2 \\
Nova 2 Pro & 1800 & 0.0 & 0.0 & 0.0 \\
\bottomrule
\end{tabular}

\end{table}
\begin{table}[h]
\centering\small
\caption{Calls and tokens over every input on the rendered tiers. The total leaves out the CV-Bench calls and the calls to the two probed models, which ran locally.}
\label{tab:cost}
\begin{tabular}{lrrr}
\toprule
model & calls & input tokens (M) & output tokens (M) \\
\midrule
Claude Opus 5.5 & 21,491 & 26.8 & 5.66 \\
Claude Opus 5.5 (high effort) & 1,200 & 1.1 & 0.32 \\
GPT-5.5 & 20,191 & 17.5 & 11.17 \\
GPT-5.5 (high effort) & 1,200 & 0.9 & 3.58 \\
Gemini 3.6 Flash & 20,191 & 25.6 & 8.81 \\
Gemini 3.6 Flash (high effort) & 1,200 & 1.4 & 2.45 \\
Pixtral Large & 8,431 & 21.5 & 0.44 \\
Qwen3-VL-235B & 8,431 & 6.6 & 0.33 \\
Llama 4 Maverick & 8,431 & 26.4 & 2.56 \\
Gemma 3 27B & 8,431 & 4.0 & 0.24 \\
Grok 4.7 & 8,431 & 6.5 & 13.05 \\
Nova 2 Pro & 8,431 & 8.3 & 0.33 \\
\midrule
total & 116,059 & & \\
\bottomrule
\end{tabular}

\end{table}

\section{Proposed human study protocol (not run)}
\label{app:human}
\small
The code release will include a rating interface that shows an item with no labels, records a True / False / cannot-tell answer and a confidence, and a scorer that computes the same metrics as for models, including the adjusted gap, plus Cohen's $\kappa$ between raters. The planned design is at least ten naive raters on the 600 medium items, single-view presentation, confidence and response time recorded, majority vote, and the human raw and adjusted gaps reported beside the models, with the 49 items that are cue-wrong in both views as the difficulty control cell. Viewing should be monocular, with a chin rest at the distance at which the image subtends its rendered horizontal field of view, since perceived depth in pictures changes with viewing distance \citep{cooper2012perceptual}. A second, free-viewing condition and a map-drawing version of the task would place human and model compression on the constancy index of Appendix~\ref{app:constancy}. This study was not run, and no other study in the paper replaces it.

\FloatBarrier
\section{Error examples}
\label{app:examples}
\begin{figure}[h]\centering\includegraphics[width=\textwidth]{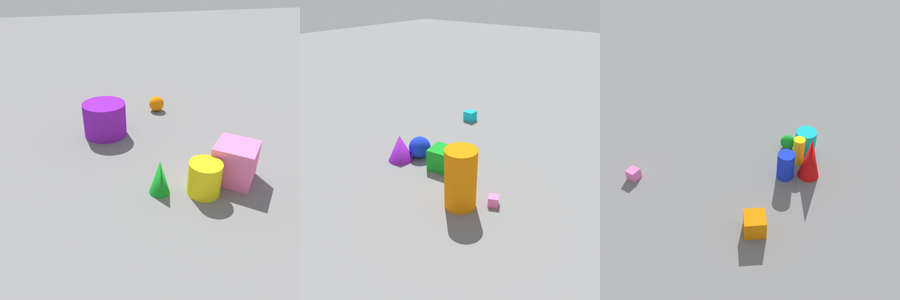}
\caption{Cue-wrong items answered wrongly by Claude Opus 5.5 with chain of thought (medium tier). From left to right, \family{fits}, \family{between} and \family{touching}.}\label{fig:examples}\end{figure}

\small
We show three cue-wrong items that Opus 5.5 gets wrong with chain of thought, the median-length failure in each of three relations. The rationales are illustrative outputs, not a faithful trace of the computation. On a \family{fits} item the model estimates ``the cube's width (about 100 px at that depth)'' and ``the gap (about 60--70 px)'' and attempts a perspective correction, which it gets wrong. On a \family{between} item it reasons on the ground plane and then accepts a ``roughly midway'' offset, a looser test than our $10^\circ$ rule. On a \family{touching} item it concludes from a ``margin [of] only a few pixels''. Appendix Figure~\ref{fig:examples} shows the three items.

\paragraph{\family{fits}, medium-0130.} \emph{The pink cube would fit through the gap between the purple cylinder and the orange sphere.} Truth: true, 2-D cue: false, model: false. Rationale excerpt: ``[...] If it were moved back to the depth of the cylinder and sphere, perspective would shrink it to roughly 100 px. 4. **Comparison:** The cube's width (about 100 px at that depth) is clearly larger than the gap (about 60–70 px). Even rotated to its narrowest orientation, a cube's minimum width is its side length, so it cannot pass through.''

\paragraph{\family{between}, medium-0185.} \emph{The green cube lies in between the purple cone and the orange cylinder.} Truth: false, 2-D cue: true, model: true. Rationale excerpt: ``[...] So on the actual ground plane the offset is somewhat larger than it looks, but still modest relative to the distance between the cone and the cylinder. The objects form a rough chain on the ground: cone, then sphere, then cube, then cylinder. The cube is not exactly on the line, but it sits roughly midway between the cone and the cylinder, so I judge the statement true.''

\paragraph{\family{touching}, medium-0580.} \emph{The green sphere and the cyan cylinder are in contact with each other.} Truth: true, 2-D cue: false, model: false. Rationale excerpt: ``[...] So they are also offset in depth, which increases the true ground-plane distance further. **Conclusion:** These measurements suggest a small gap between the sphere and the cyan cylinder, with the yellow cylinder near both of them. The margin is only a few pixels, so this is a close call, but they most likely do not touch.''

\section{Datasheet}
\label{app:datasheet}
\small
\paragraph{Motivation.} \bench{} was created to measure whether vision-language models answer 3-D spatial statements from scene geometry or from image-plane regularities. It was built by the author at Scale AI. No external funding is involved.

\paragraph{Composition.} Three splits (\texttt{easy}, \texttt{medium}, \texttt{hard}) of 600 true/false statements each from seed 7, a second 600-item medium split from seed 11, 1{,}252 rendered scenes of 4--8 primitives across these four splits, plus a 600-item checkerboard-ground re-render of the medium scenes and a 50-item sample inspected by hand. Each item carries the statement, its truth, the 2-D cue's prediction, whether the cue is right or wrong, relation, margin, polarity, paraphrase variant, referenced object indices, the rendered image ($768\times768$ PNG), a 16-bit depth map, a colourised depth visualisation, and for view-invariant relations a second-camera render with its own cue prediction. Each scene carries the full scene graph (object shape, colour, position, size, height, yaw), camera and light parameters, per-object analytic 2-D statistics (visible mask area, occlusion, centroid, ground-contact pixel, pixel area, depth) and the list of sampler constructs used. Every model response used in this paper will be released as a JSON record (prompt, raw response, parsed answer, token usage, latency). No people, personal data or offensive content appear, and all images are synthetic.

\paragraph{Collection process.} Scenes are drawn by a seeded rejection sampler (Appendix~\ref{app:generation}). Statements are enumerated exhaustively per scene, labelled from the scene graph and selected by an exact balancer. Rendering uses Blender 5.0.1 Cycles, and masks and depth come from an analytic ray caster over the same primitives. Model responses were collected in October 2026 through an OpenAI-compatible proxy with the parameters in Section~\ref{sec:protocol}. No human annotation was used. A rating interface for a human study will be released. No ratings were collected.

\paragraph{Preprocessing and labelling.} None beyond the generator. Truth and cue labels are computed, not annotated. The reject bands in Table~\ref{tab:families} discard near-threshold candidates.

\paragraph{Uses.} Intended for evaluating 3-D spatial understanding of VLMs and for studying cue reliance. The generator can produce arbitrarily many fresh items from a new seed, so training on the benchmark splits defeats the purpose and is discouraged. Not intended as a measure of performance on natural photographs.

\paragraph{Distribution and maintenance.} Code (generator, evaluation harness, analysis, rating tool) will be released under the MIT licence, and rendered data, labels and model responses under CC BY 4.0. Code, data and model responses will be hosted in a public repository. A DOI and Croissant metadata will accompany the data release. The author maintains the benchmark, and issues and new-model submissions will be accepted through that repository. Version 1.0 will hold the four splits evaluated here, the easy, medium and hard splits from seed 7 and the second medium split from seed 11. Any change to generation rules or margins will be a new version with its own results table.

\paragraph{Ethical considerations.} The data are synthetic and carry no privacy or consent risk. The main ethical consideration is interpretive. The shortcut gap is a one-sided instrument (a zero gap shows no cue-right advantage for the named cue, not that the cue is never used), and we ask users not to read a small gap at chance-level accuracy as 3-D competence. The gap measures alignment of answers with a cue, and reading it as reliance needs the input controls described in the paper.

\end{document}